%% file: main.tex
\documentclass[10pt,twocolumn,letterpaper]{article}

\usepackage[pagenumbers]{iccv} 
\usepackage{booktabs}    
\usepackage{multirow}    
\usepackage{threeparttable}  
\usepackage{amssymb}   
\usepackage{wasysym}  
\usepackage{xcolor}  
\usepackage{url}
\usepackage{pifont}     

\definecolor{iccvblue}{rgb}{0.21,0.49,0.74}
\usepackage[pagebackref,breaklinks,colorlinks,allcolors=iccvblue]{hyperref}

\title{GEOBench-VLM: Benchmarking Vision-Language Models for Geospatial Tasks}

\author{
    \fontsize{10}{12}\selectfont Muhammad Sohail Danish$^{*1}$ 
    \quad
    \fontsize{10}{10}\selectfont Muhammad Akhtar Munir$^{*1}$ 
    \quad
    \fontsize{10}{10}\selectfont Syed Roshaan Ali Shah$^2$ 
    \quad
    \fontsize{10}{10}\selectfont Kartik Kuckreja$^1$ \\
    \fontsize{10}{10}\selectfont Fahad Shahbaz Khan$^{1, 3}$
    \quad
    \fontsize{10}{10}\selectfont Paolo Fraccaro$^4$ 
    \quad
    \fontsize{10}{10}\selectfont Alexandre Lacoste$^5$ 
    \quad
    \fontsize{10}{10}\selectfont Salman Khan$^{1, 6}$
    \\
    \footnotesize{$^1$Mohamed bin Zayed University of Artificial Intelligence},  
    \footnotesize{$^2$University College London}, 
    \footnotesize{$^3$Link\"oping University, Sweden} \\
    \footnotesize{$^4$IBM Research Europe, UK}, 
    \footnotesize{$^5$ServiceNow Research},
    \footnotesize{$^6$Australian National University}
}

\begin{document}
\input{sec/main_fig}
\maketitle

\input{sec/0_abstract}

\input{sec/1_intro}
\input{sec/2_benchtask}

\input{sec/3_evalres}

\input{sec/4_analysis}

\input{sec/5_conclusion}

{
    \small
    \bibliographystyle{ieeenat_fullname}
    \bibliography{main}
}
\newpage
\clearpage
\input{sec/X_suppl}

\end{document}

%% file: sec/main_fig.tex
\twocolumn[{%
\renewcommand\twocolumn[1][]{#1}%
\maketitle%
\vspace{-0.3in}%
\begin{center}
    \centering

    \includegraphics[width=0.9\textwidth]{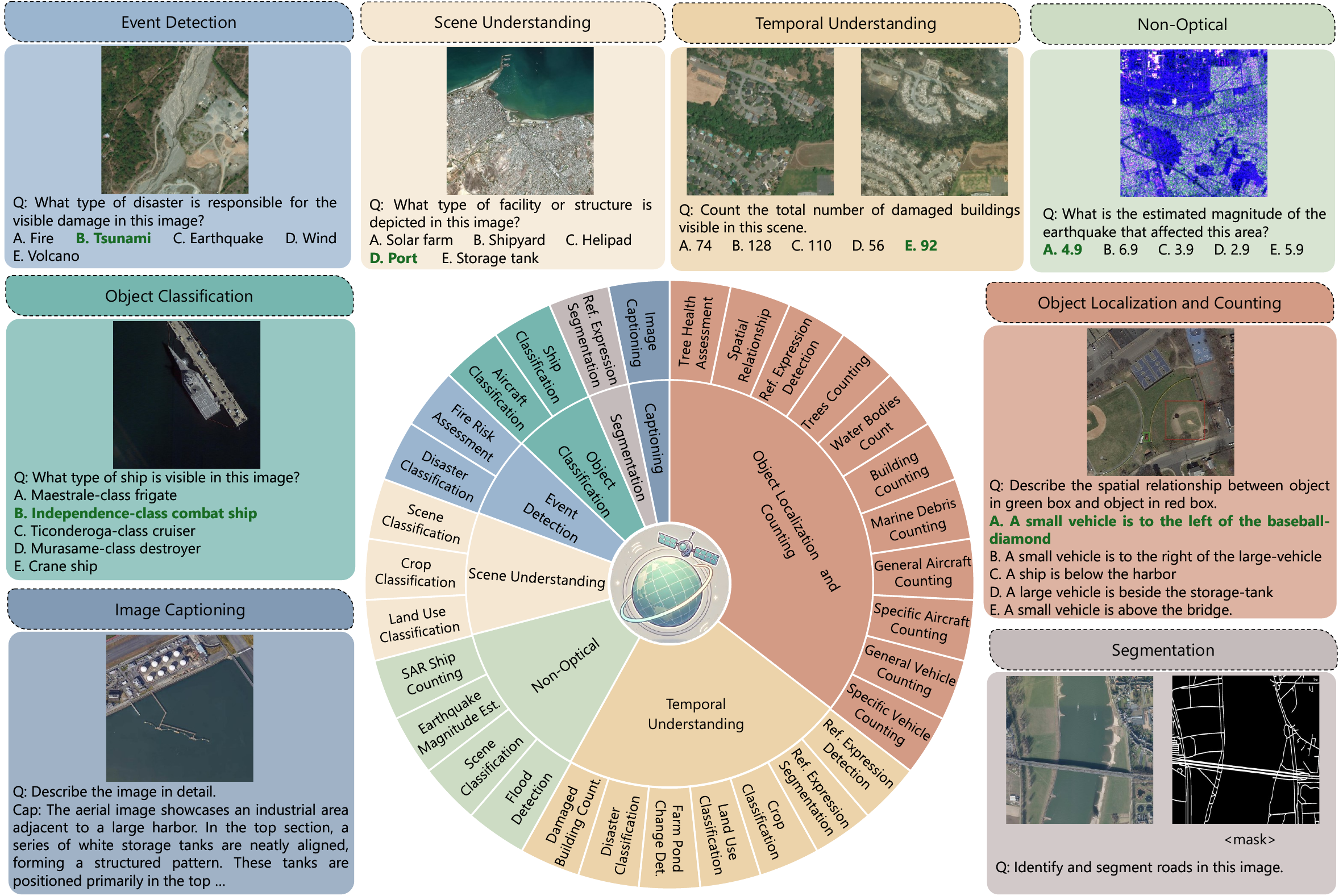}
    \captionof{figure}{GEOBench-VLM comprehensively covers 31 fine-grained tasks categorized into 8 broad categories: scene and object classification, object detection, segmentation, captioning, event detection, non-optical and temporal understanding tasks.}
    \label{fig:teaser}
\end{center}
}]

%% file: sec/0_abstract.tex
\begingroup
\renewcommand\thefootnote{}
\footnotetext{*Equally contributing first authors.}
\endgroup
\begin{abstract}
\label{sec:abs}
While numerous recent benchmarks focus on evaluating generic Vision-Language Models (VLMs), they do not effectively address the specific challenges of geospatial applications.
Generic VLM benchmarks are not designed to handle the complexities of geospatial data, an essential component for applications such as environmental monitoring, urban planning, and disaster management.
Key challenges in the geospatial domain include temporal change detection, large-scale object counting, tiny object detection, and understanding relationships between entities in remote sensing imagery.
To bridge this gap, we present GEOBench-VLM, a comprehensive benchmark specifically designed to evaluate VLMs on geospatial tasks, including scene understanding, object counting, localization, fine-grained categorization, segmentation, and temporal analysis. 
Our benchmark features over 10,000 manually verified instructions and spanning diverse visual conditions, object types, and scales.
We evaluate several state-of-the-art VLMs to assess performance on geospatial-specific challenges. 
The results indicate that although existing VLMs demonstrate potential, they face challenges when dealing with geospatial-specific tasks, highlighting the room for further improvements.  
Notably, the best-performing LLaVa-OneVision achieves only 41.7\% accuracy on MCQs, slightly more than GPT-4o, which is approximately
double the random guess performance.
Our benchmark is publicly available at \href{https://github.com/The-AI-Alliance/GEO-Bench-VLM}{https://github.com/The-AI-Alliance/GEO-Bench-VLM}.
\end{abstract}

%% file: sec/1_intro.tex
\vspace{-0.2in}
\section{Introduction}
\label{sec:intro}
Deep learning has revolutionized computer vision \cite{russwurm2020meta, jean2016combining, xie2021segformer, zhu2020deformable, zhangdino} and natural language processing \cite{bubeck2023sparks, touvron2023llama1, touvron2023llama, achiam2023gpt}, enabling significant advancements in integrating visual and textual data. 
VLMs combine image and language understanding to handle tasks that involve visual and textual information \cite{li2024llavaone, chen2024internvl, bai2023qwen, rasheed2024glamm}.
These models have demonstrated strong performance in tasks like image captioning \cite{yang2024exploring}, visual question answering \cite{chen2024spatialvlm}, and object recognition \cite{xiao2024florence}, benefiting from training on large-scale visual-text datasets.

VLMs have the potential to enhance geospatial analysis in domains that rely on satellite and aerial imagery, such as urban planning, environmental assessment, and disaster management \cite{yuan2024rrsis, wang2024earthvqa, kuckreja2024geochat, soni2024earthdial,liu2024rotated, li2024new}. 
By automating labor-intensive tasks such as land cover mapping, damage assessment, and object detection, VLMs can reduce manual effort while improving the efficiency and accuracy of geospatial analysis.
However, despite their potential, VLMs remain limited in geospatial applications due to specific challenges inherent to geospatial data.
This is due to the unique challenges posed by geospatial data, e.g., objects in satellite imagery can vary in scale, making it difficult for models trained on natural imagery datasets.
Moreover, satellite images are captured under diverse resolutions and lighting conditions, affecting visibility and image quality.
Many geospatial applications also require temporal analysis to detect changes over time, such as monitoring urban development or environmental degradation, introducing an additional layer of complexity. 
The lack of geospatial-specific benchmarks hinders the objective assessment of VLMs, making it difficult to identify performance gaps and areas for improvement \cite{li2024seed, yue2024mmmu, liu2025mmbench}.

Existing benchmarks, such as SEED-Bench \cite{li2024seed} and MMMU \cite{yue2024mmmu}, have contributed to evaluating VLMs in general vision-language tasks such as visual question answering and scene understanding.
However, these benchmarks do not adequately cover geospatial tasks requiring spatial reasoning and temporal analysis.
On the other hand, VLEO benchmarks geospatial tasks, evaluating VLM performance in earth observation applications \cite{zhang2024good}, but it lacks components for extended temporal analysis, non-optical data, and segmentation tasks.
Another limitation is that VLEO primarily evaluates generic VLMs rather than models tailored for geospatial tasks, not considering specialized approaches.
Although VLMs can improve the interpretation of data, the lack of a thorough evaluation highlights the need to assess both geospatial-specific and generic models on comprehensive geospatial benchmarks to effectively address geospatial tasks.

To bridge this gap, we introduce GEOBench-VLM, a comprehensive benchmark suite designed to evaluate VLMs \emph{(generic and geospatial-specific)} on geospatial tasks (Fig.~\ref{fig:teaser}). 
GEOBench-VLM integrates diverse datasets with varying visual conditions, resolutions, and object scales supplemented with both automated and manually verified annotations.
It employs multiple-choice questions (MCQs) similar to \cite{li2024seed}, to ensure objective, scalable, and automated evaluation, reducing biases associated with open-ended responses. 
This structured approach also prevents hallucinations and addresses problems in understanding model responses for fair assessment.
GEOBench-VLM covers key task categories, including \emph{scene understanding, object counting, visual grounding, image captioning, temporal understanding, non-optical, referring segmentation, and relational reasoning} (Fig.~\ref{fig:teaser}). 
These tasks are essential for a wide range of applications, including but not limited to urban planning, monitoring deforestation, assessing environmental changes, and managing natural disaster response.

We evaluate several state-of-the-art VLMs using GEOBench-VLM to assess their performance in geospatial tasks. 
Our findings indicate that while generic VLMs, including commercial ones like GPT-4o, have potential, they struggle with geospatial tasks.
We also include geospatial-specific VLMs in our evaluation to provide a comparative analysis of different models.
Performance varies across geospatial tasks, with no single model excelling in all areas.
LLaVA-OneVision \cite{li2024llavaone} leads in object localization and counting; GPT-4o excels in object classification; Qwen2-VL \cite{wang2024qwen2} shows strength in event detection and interpreting non-optical imagery.
Our comparative analysis highlights the benefits of analyzing models for geospatial tasks and identifies areas needing improvement.
We can summarize our contributions as follows:

\begin{enumerate}
    \item We introduce GEOBench-VLM, a benchmark suite designed specifically for evaluating VLMs on geospatial tasks, addressing geospatial data challenges. It covers \textit{\textbf{8}} broad categories and \textit{\textbf{31}} sub-tasks with over \textit{\textbf{10,000}} manually verified instructions.
    \item We provide a detailed evaluation of \textit{\textbf{13}} state-of-the-art VLMs, including generic (open and closed-source) and geospatial-specific VLMs, highlighting their capabilities and limitations in geospatial analysis.
    \item We analyze performance across a range of tasks, including scene classification, counting, change detection, relationship prediction, referring expression detection, segmentation, image captioning, disaster detection, and temporal analysis, among others, providing key insights that can help in improving VLMs for geospatial applications.
\end{enumerate}

%% file: sec/2_benchtask.tex
\begin{table*}[ht]
\centering
\setlength{\tabcolsep}{2pt} 
\renewcommand{\arraystretch}{1.0} 
\newcommand{\cmark}{\ding{51}} 
\newcommand{\xmark}{\ding{55}} 

\resizebox{\textwidth}{!}{%
\begin{tabular}{l c c c c c c c c}
\toprule
\textbf{Benchmark} & \textbf{Domain} & \textbf{Modalities} & \textbf{Data Sources} & \textbf{Answer Type} & \textbf{Annotation Type} & \textbf{Human Verify} & \textbf{Year} & \textbf{RS Category} \\ \midrule
CulturalVQA\cite{nayak2024benchmarking} & General & O & Curated & FF & M & \xmark & 2024 & N/A \\ 
EXAMS-V\cite{yin2021broaden} & General & O & Academic Exams & MCQ & M & - & 2024 & N/A \\ 
M4U\cite{wang2024m4u} & General & O & Academic Exams & MCQ & M & \cmark & 2024 & N/A \\ 
MMMU\cite{yue2024mmmu} & General & O & Academic Exams & FF, MCQ & M & \cmark & 2024 & N/A \\ 
MME\cite{zhang2024mme} & General & O & Various Open-Source & Yes/No & M & \cmark & 2024 & N/A \\ 
MMBench\cite{liu2025mmbench} & General & O & Various Open-Source & MCQ & A+M & \xmark & 2024 & N/A \\ 
MMStar\cite{chen2024we} & General & O & Existing Benchmarks & MCQ & M & \cmark & 2024 & N/A \\ 
LAMM\cite{yin2024lamm} & General & O, PC & Various Open-Source & FF & A+M & \xmark & 2023 & N/A \\ 
SEED-Bench\cite{li2023seed} & General & O, V & Various Open-Source & MCQ & A+M & \cmark & 2023 & N/A \\ 
SEED-Bench2\cite{li2024seed} & General & O, MI, V & Various Open-Source & MCQ & A+M & \cmark & 2024 & N/A \\ 
SEED-Bench-H\cite{li2024seed} & General & O, MI, V & Public Data / Curated & MCQ & M & \cmark & 2024 & N/A \\ 
\hline \hline
RSIEval\cite{hu2023rsgpt} & RS & O, PAN & DOTA\cite{xia2018dota} & FF & M & \cmark & 2023 & 6 \\ 
LHRS-Bench\cite{muhtar2024lhrs} & RS & O & GE + OSM & SC & M & \cmark & 2024 & 4 \\ 
EarthGPT\cite{zhang2024earthgpt} & RS & O, IR, SAR & DRSD & FF, BBox & A+M & - & 2024 & 5 \\ 
SkyEyeGPT\cite{zhan2024skyeyegpt} & RS & O, V & DRSD & FF, MCQ & A+M & \cmark & 2024 & 6 \\ 
Fit-RSRC\cite{luo2024skysensegpt} & RS & O & DRSD & FF, MCQ & A+M & \cmark & 2024 & 1 \\ 
Fit-RSFG\cite{luo2024skysensegpt} & RS & O & DRSD & FF, MCQ & A+M & \xmark & 2024 & 5 \\ 
VRSBench\cite{li2024vrsbench} & RS & O & DOTA\cite{xia2018dota}, DIOR\cite{li2020object} & FF, BBox & A+M & \cmark & 2024 & 3 \\ 
VLEO-Bench\cite{zhang2024good} & RS & O, BT & DRSD & FF, BBox, MCQ & A+M & \cmark & 2024 & 6 \\ 
EarthVQA\cite{wang2024earthvqa} & RS & O & DRSD & FF & A+M & \xmark & 2023 & 5 \\ 
RemoteCount\cite{liu2024remoteclip} & RS & O & DOTA\cite{xia2018dota} & SC & A+M & \cmark & 2024 & 1 \\ 
FineGrip\cite{zhao2024panoptic} & RS & O & MAR20\cite{wenqi2024mar20} & FF, Seg & A+M & \cmark & 2024 & 5 \\ 
GeoChat-Bench\cite{kuckreja2024geochat} & RS & O & DRSD & FF, BBox & A+M & \cmark & 2023 & 6 \\ 
\hline
GEOBench-VLM (\small{Ours}) & RS & O, MS, SAR, BT, MT & DRSD & MCQ, BBox, Seg & A+M & \cmark & 2025 & 8 \\ 
\bottomrule
\end{tabular}
}
\vspace{-0.1in}
\caption{Overview of Generic and Geospatial-specific Datasets \& Benchmarks, detailing modalities (O=Optical, PAN=Panchromatic, MS=Multi-spectral, IR=Infrared, SAR=Synthetic Aperture Radar, V=Video, MI=Multi-image, BT=Bi-Temporal, MT=Multi-temporal), data sources (DRSD=Diverse RS Datasets, OSM=OpenStreetMap, GE=Google Earth, 
answer types (MCQ=Multiple Choice, SC=Single Choice, FF=Free-Form, BBox=Bounding Box, Seg=Segmentation Mask), and annotation types (A=Automatic, M=Manual).} 
\vspace{-0.1in}
\label{tab:benchmarks}
\end{table*}

\section{Benchmark Overview}
\label{sec:overviewdata}

VLM benchmarks cover tasks from general multimodal understanding to specialized geospatial applications. A detailed comparison is in Table~\ref{tab:benchmarks}.

\noindent\textbf{Generic VLMs Benchmarks:}
Several benchmarks evaluate multimodal models across visual tasks, each with strengths and limitations. 
MMMU \cite{yue2024mmmu} evaluates models in multiple dimensions, incorporating diverse visual formats, and tests perceptual capabilities, but lacks geospatial tasks.
SEED-Bench \cite{li2024seed} focuses on spatial and temporal understanding with diverse datasets for complex multimodal contexts. 
SEED-Bench-2 \cite{li2024seed2} specializes in text-rich visual scenarios like charts and maps but interprets structured, generalized maps rather than complex geospatial data.
Other benchmarks like MMBench \cite{liu2025mmbench}, MM-Vet \cite{yu2023mm}, and MMSTAR \cite{chen2024we} evaluate spatial reasoning but do not address geospatial applications in earth observation.

\noindent\textbf{Geospatial-specific:}
Few methods, along with benchmarks, including EarthVQA \cite{wang2024earthvqa}, LHRS \cite{muhtar2024lhrs}, RS-LLaVA \cite{bazi2024rs}, EarthDial \cite{soni2024earthdial}, and GeoChat \cite{kuckreja2024geochat}, evaluate VLMs in remote sensing. 
RS-LLaVA  \cite{bazi2024rs} focuses on image captioning and VQA but lacks change detection and temporal analysis.
LHRS \cite{muhtar2024lhrs} focuses on high-resolution remote sensing tasks with an RS-specific image-text dataset but lacks multi-temporal diversity.
EarthVQA \cite{wang2024earthvqa} supports complex relational reasoning with a multi-modal dataset, though its emphasis on non-temporal imagery restricts broader applicability.
GeoChat \cite{kuckreja2024geochat} also evaluates geospatial tasks like region captioning and spatially grounded responses; however, it lacks diverse temporal datasets and segmentation.
VLEO \cite{zhang2024good} is an earth observation benchmark; however, it is limited to non-remote sensing specialized methods and lacks the diversity of tasks.
Specifically, VLEO does not include multi-temporal datasets, segmentation tasks, or non-optical imagery, restricting its effectiveness in evaluating geospatial VLM capabilities.

A comprehensive geospatial benchmark is essential for evaluating VLMs. GEOBench-VLM addresses this gap by integrating multi-temporal analysis, segmentation, and non-optical data, among other critical tasks.

\begin{figure*}[t]
    \centering
    \includegraphics[width=1.0\linewidth]{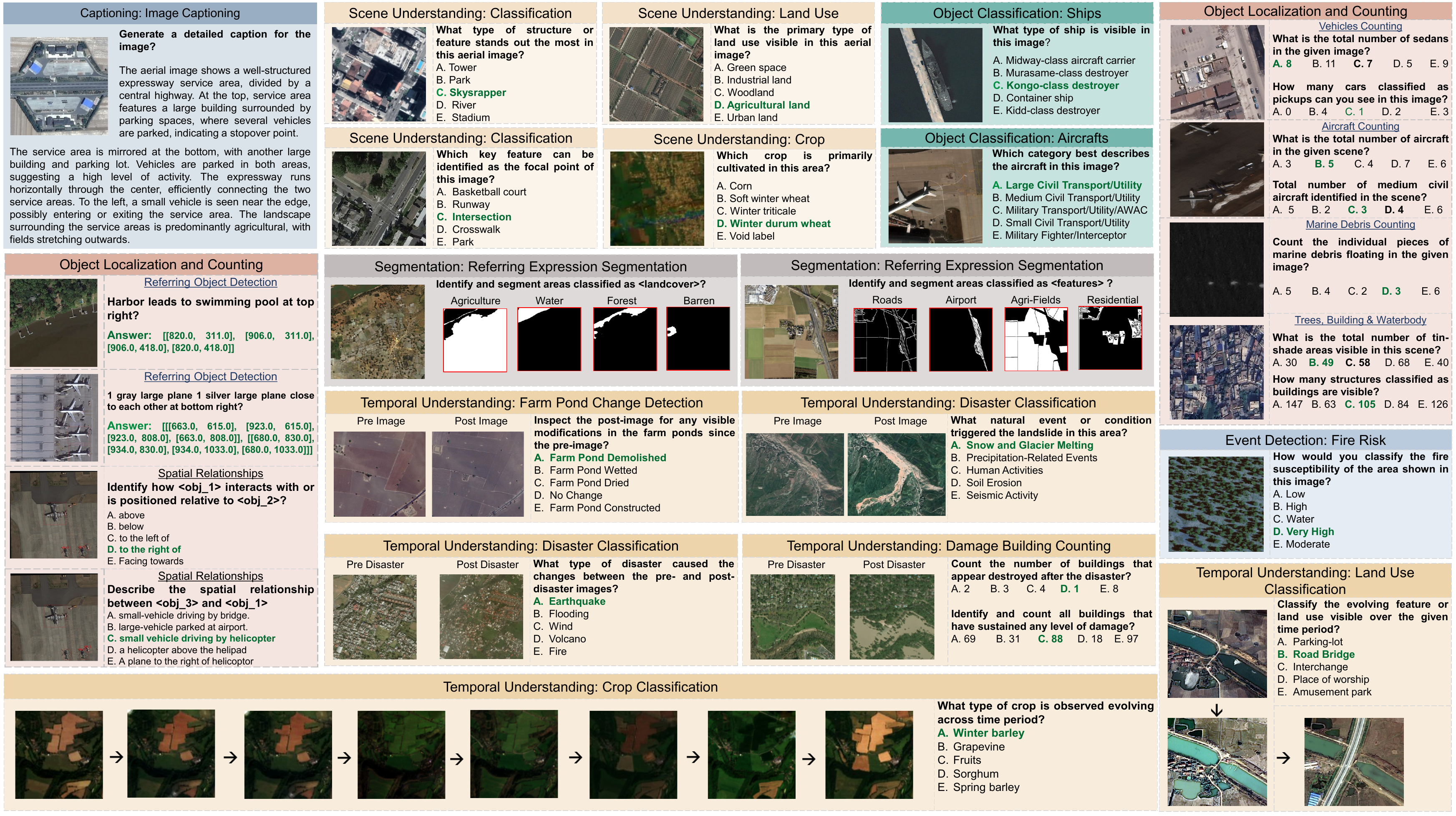}
    \captionsetup{justification=justified}
    \vspace{-0.25in}
    \caption{Comprehensive benchmark for VLMs in numerous geospatial tasks. This benchmark evaluates VLMs across eight core task categories, assessing their ability to interpret complex spatial data, classify scenes, identify and localize objects, detect events, generate captions, segment regions, analyze temporal changes, and process non-optical data. Tasks span from classifying landscapes and objects (e.g., land use, crop types, ships, aircraft) to counting, detecting hazards, and assessing disaster impact, testing VLMs on spatial reasoning.}
    \vspace{-0.1in}
    \label{fig:tasks_examples}
\end{figure*}

\section{GEOBench-VLM}
Our proposed benchmark is curated using a human-verified data pipeline. It covers 8 broad and 31 fine-grained geospatial task categories outlined below. 

\noindent\textbf{Categories:}
The diverse geospatial tasks in GEOBench-VLM are carefully designed to evaluate the capabilities of VLMs.  
The 8 categories offer a comprehensive framework for assessing spatial reasoning, object classification, segmentation, temporal understanding, and more. 
These tasks are crucial for real-world applications like disaster response, urban planning, and environmental monitoring. See Fig.~\ref{fig:tasks_examples} for task illustrations.
 \textbf{1) Scene Understanding:} This category includes tasks like scene classification and land use classification, where models distinguish environments (e.g., airports, forests, bridges) and land types (e.g., agricultural, industrial, residential) respectively. Crop classification requires models to identify crops using visual and environmental cues.
\textbf{2) Object Classification:} This category focuses on identifying specific objects, including ship type classification (e.g., aircraft carrier, cargo ship) and aircraft type classification (from civil transport to military bombers), assessing the model’s capability in fine-grained categories.
\textbf{3) Object Localization and Counting:} This includes referring expression detection, where models predict bounding boxes based on text queries and spatial relationship prediction. Counting tasks cover general and specific objects, examples include vehicle and aircraft counting, damaged and healthy tree counting, building and water body counting, and marine debris assessment.
\textbf{4) Event Detection:} This focuses on fire risk detection and assesses fire hazards across forests, while disaster-type classification involves inferring potential disaster causes from post-event imagery.
\textbf{5) Caption Generation:} This task evaluates image captioning, assessing a model’s ability to generate descriptions that capture both overall scene context and specific object details.
 \textbf{6) Semantic Segmentation:} The task includes referring expression segmentation, where models generate binary masks for specified objects or regions (e.g., urban vs. non-urban areas).
\textbf{7) Temporal Understanding:} This category includes tasks such as change detection and damage assessment to identify differences over time, such as post-disaster building conditions. Long temporal analysis includes tasks like crop classification.
\textbf{8) Non-Optical:} This category focuses on non-optical imagery analysis, including ship detection, flood detection, and earthquake magnitude estimation, crucial for scenarios where optical imagery is insufficient.

\begin{figure*}[t]
    \centering
    \includegraphics[width=1.0\linewidth]{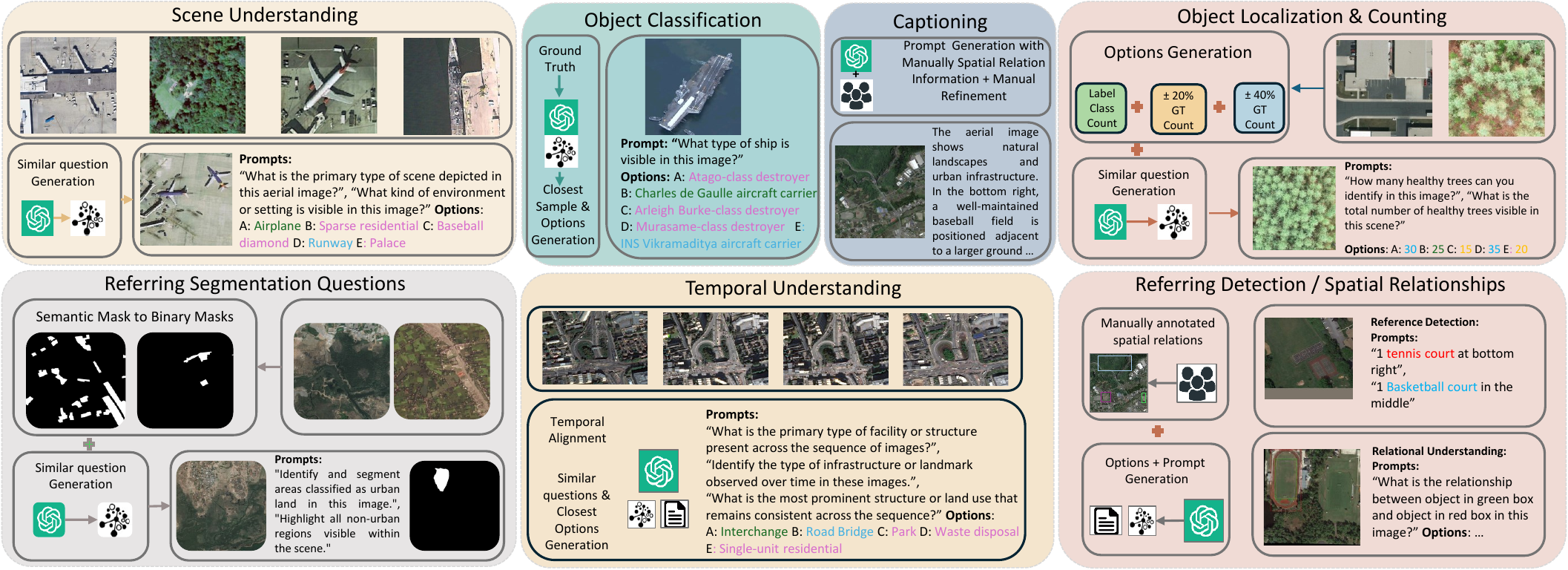}
    \captionsetup{justification=justified}
    \vspace{-0.2in}
    \caption{Data pipeline for the GEOBench-VLM: Our pipeline integrates diverse datasets, automated tools, and manual annotation. Tasks such as scene understanding, object classification, and non-optical analysis are based on classification datasets, while GPT-4o generates unique MCQs with five options: one correct answer, one semantically similar ``closest" option, and three plausible alternatives. Spatial relationship tasks rely on manually annotated object pair relationships, ensuring consistency through cross-verification. Caption generation leverages GPT-4o, combining image, object details, and spatial interactions with manual refinement for high precision.}
    \vspace{-0.1in}
    \label{fig:dataset_process}
\end{figure*}
\noindent\textbf{Dataset pipeline:}
GEOBench-VLM integrates open datasets, and manual annotation aided with automated tools. For diversity, each task samples images from multiple datasets\footnote{For the complete list of datasets, see the supplementary section.}. 
By combining multi-source datasets with structured question design, GEOBench-VLM enables scalable and high-quality evaluation of VLMs across geospatial tasks.
Fig.~\ref{fig:dataset_process} shows the overall pipeline.

For scene understanding tasks, including scene classification, land use classification, and crop type classification, we use classification datasets \cite{airound, resis, MtSCCD, FireRisk, pastis, PatternNet}.
We used GPT-4o to generate unique questions for each task, with five answer options: one correct answer, one semantically similar ``closest" option (verified manually), and three plausible alternatives. 
For counting tasks, we converted object detection data \cite{cowc, RarePlanes_Dataset, deforestation-satellite-imagery-335n4_dataset} into questions about the number of specific objects in an image, providing the correct count and alternatives with controlled deviations ($\pm 20\%$ and $\pm 40\%$) to maintain plausibility. For referring expression segmentation, we use segmentation datasets to create binary masks and prompts for identifying specific objects or regions.
\noindent For spatial relationship tasks, we manually annotated relationships between object pairs using object locations from detection datasets \cite{xia2018dota, dior, fair1m}, with cross-verification for consistency. Following our standard approach, five MCQs were designed to assess these relationships.
For referring expression tasks, where existing datasets lack complex queries, our benchmark provides diverse questions. We collect basic object attributes (names, locations, colors) and annotated spatial relationships to generate queries that incorporate object details and interactions. Each query is reviewed for clarity and complexity.
For caption generation, we use GPT-4o with image data, object attributes, and spatial relationships to create descriptions combining scene context and object details. Captions are manually refined for clarity, removing irrelevant or repetitive details.

\begin{figure*}[t]
    \centering
    \includegraphics[width=1.0\linewidth]{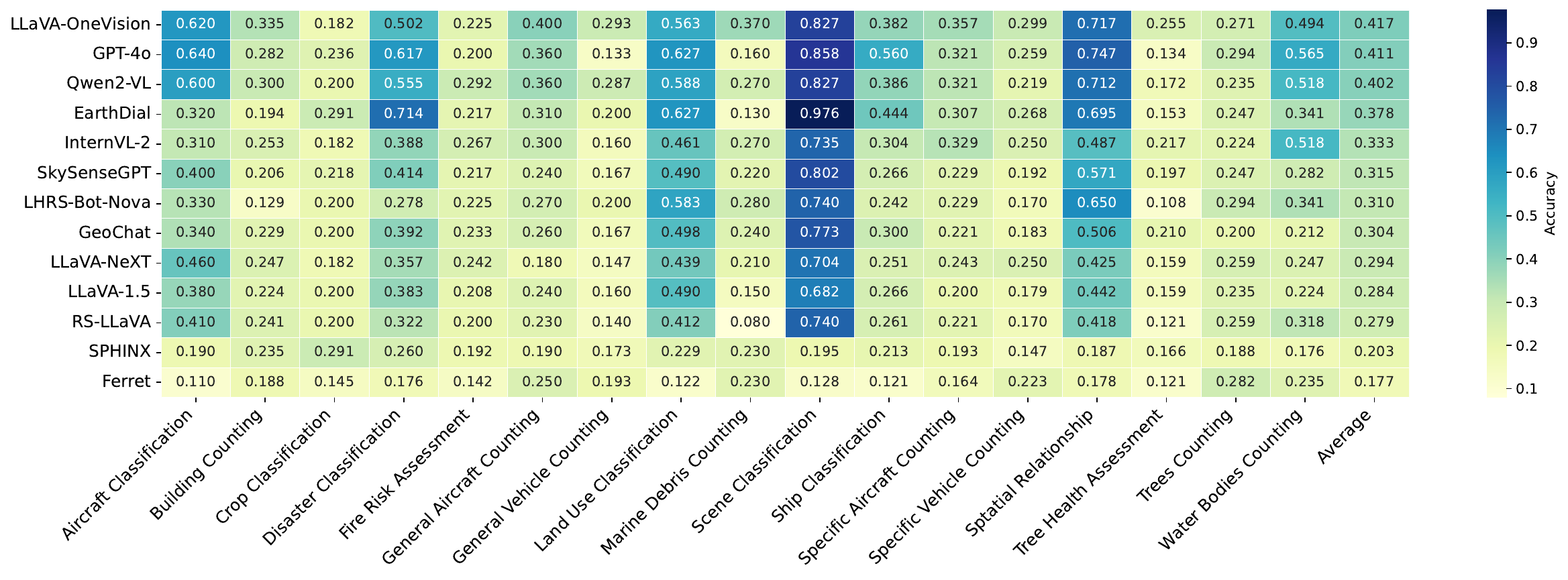}
    \captionsetup{justification=justified}
    \vspace{-0.2in}
    \caption{
    Performance summary of VLMs. LLaVA-OneVision achieves the average accuracy (41.7\%), slightly outperforming GPT-4o, which is relatively better in building counting, and general aircraft counting. EarthDial demonstrates strong results in scene classification. The overall results highlight VLMs' varying strengths across geospatial tasks, with even the best models achieving accuracy only slightly above double the random guess.}
    \vspace{-0.1in}
    \label{fig:results_heatmap}
\end{figure*}

\section{VLMs Benchmarking}
\subsection{Selection of VLMs}
\noindent\textbf{Model Variability and Capabilities:}
We selected both generic and geospatial-specific VLMs, prioritizing recently developed models with advanced capabilities across various tasks.
Generic VLMs include LLaVA-1.5 \cite{li2024llava}, LLaVA-NeXT \cite{liu2024llavanext}, LLaVA-OneVision \cite{li2024llavaone}, Sphinx \cite{lin2023sphinx}, Ferret \cite{you2023ferret}, InternVL2 \cite{chen2024internvl}, and Qwen2-VL \cite{wang2024qwen2}. 
Also, we added GPT-4o, a closed-source, commercially available model, to compare its adaptability to remote sensing tasks with open-source models.
These VLMs demonstrate robust performance in tasks such as scene understanding, and fine-grained visual classification.
For geospatial-specific VLMs, we selected GeoChat \cite{kuckreja2024geochat}, RS-LLaVA \cite{bazi2024rs}, SkySenseGPT \cite{luo2024skysensegpt}, EarthDial \cite{soni2024earthdial}, and LHRS-Bot-Nova \cite{muhtar2024lhrs}, models tailored for satellite and aerial image interpretation.

\noindent\textbf{Domain Specific Model Suitability:}
We consider VLMs based on their relevance to geospatial tasks. 
Despite being trained on satellite and aerial data, domain-specific models \cite{kuckreja2024geochat, bazi2024rs, muhtar2024lhrs, luo2024skysensegpt, soni2024earthdial} may struggle with counting and spatial relationships due to dataset or architecture constraints. 
Meanwhile, we evaluate whether generic VLMs can effectively handle scene understanding, object detection, and visual reasoning in remote sensing. 
This selection ensures a focused study of VLM performance across diverse geospatial applications.

\noindent\textbf{Open vs Closed Models:}
Open-source models like LLaVA \cite{li2024llava, liu2024llavanext, li2024llavaone}, Qwen2-VL \cite{wang2024qwen2}, GeoChat \cite{kuckreja2024geochat}, and others provide transparency, aiding in understanding their strengths and limitations. 
Closed models like GPT-4o, despite their lack of transparency, demonstrate strong generalization and performance on complex tasks, benefiting from proprietary datasets and advanced architectures. 
Evaluating both ensures a comprehensive assessment of VLM capabilities.

\subsection{Benchmarking Approaches}
\noindent\textbf{Task Setup \& Complexity Levels:}
Our benchmark evaluates VLMs across varied geospatial task complexities, spanning basic (e.g., scene classification), intermediate (e.g., image captioning, object counting), and advanced tasks (e.g., risk assessment, spatial relations, change detection) that require spatial and temporal reasoning. The dataset includes annotations, temporal sequences, and segmentation data for a structured evaluation.

\noindent\textbf{Metrics:}
Our evaluation framework employs task-specific metrics to assess model performance.
For MCQ-based tasks, we report accuracy as the evaluation metric. Precision evaluates referring expression detection, mIOU measures segmentation performance, and BERT score estimates image caption quality.

%% file: sec/3_evalres.tex
\section{Evaluations and Results}
\label{sec:evalres}

In this section, we evaluate and discuss the results for each task in GEOBench-VLM as follows.

\begin{table}[t]
\centering
\setlength{\tabcolsep}{2pt} 
\renewcommand{\arraystretch}{1.0} 
\fontsize{8}{10}\selectfont
\resizebox{\columnwidth}{!}{

\begin{tabular}{l|c|c|c|c|c}
\hline 
\textbf{Model} & \textbf{Event Det} & \textbf{Object Cls} & \textbf{Counting} & \textbf{Scene Und} & \textbf{Image Cap} \\
\hline\hline

GPT-4o \cite{openai2024gpt4o} & 0.4726 & \textbf{0.5863} & 0.3965 & 0.7114 & 0.6418 \\ 
EarthDial \cite{soni2024earthdial} & \textbf{0.5418} & 0.4039 & 0.3626 & \textbf{0.7705} & 0.5378 \\ 
Qwen2-VL \cite{wang2024qwen2} & 0.4640 & 0.4560 & 0.4019 & 0.6761 & 0.5895 \\ 
LLaVA-OneVision \cite{li2024llavaone} & 0.4063 & 0.4593 & \textbf{0.4377} & 0.6636 & 0.6317 \\ 
SkySenseGPT \cite{luo2024skysensegpt} & 0.3458 & 0.3094 & 0.3119 & 0.6205 & 0.6416 \\ 
InternVL-2 \cite{chen2024internvl} & 0.3458 & 0.3062 & 0.3280 & 0.5727 & 0.5968 \\ 
GeoChat \cite{kuckreja2024geochat} & 0.3372 & 0.3127 & 0.2922 & 0.6091 & 0.4395 \\ 
LHRS-Bot-Nova \cite{muhtar2024lhrs} & 0.2594 & 0.2704 & 0.3286 & 0.6330 & 0.6275 \\ 
LLaVA-NeXT \cite{liu2024llavanext} & 0.3170 & 0.3192 & 0.2737 & 0.5477 & 0.6293 \\ 
LLaVA-1.5 \cite{li2024llava} & 0.3228 & 0.3029 & 0.2618 & 0.5625 & 0.6346 \\ 
RS-LLaVA \cite{bazi2024rs} & 0.2795 & 0.3094 & 0.2534 & 0.5534 & 0.5604 \\ 
SPHINX \cite{lin2023sphinx} & 0.2363 & 0.2052 & 0.1860 & 0.2170 & \textbf{0.6451} \\ 
Ferret\cite{you2023ferret} & 0.1643 & 0.1173 & 0.1956 & 0.1261 & 0.5615 \\
\hline
\end{tabular}}
\vspace{-0.1in}
\caption{VLMs accuracies across geospatial tasks. Evaluation includes event detection, object classification, counting, scene understanding, and image captioning. 
EarthDial performs better in event detection and scene understanding, demonstrating geospatial reasoning capabilities. GPT-4o performs best in object classification, while LLaVA-OneVision leads in counting. As evaluated by BERT score \cite{bert-score}, Sphinx excels in image captioning.
}
\vspace{-0.1in}
\label{tab:accuracy_scores}
\end{table}

\noindent\textbf{Scene Understanding:}
Scene understanding in remote sensing encompasses agriculture, urban planning, and environmental monitoring. Models struggle with crop classification, probably due to low-resolution imagery. GPT-4o and EarthDial \cite{soni2024earthdial} excel in land use, and scene classification (Fig.~\ref{fig:results_heatmap}). Ferret \cite{you2023ferret} underperforms, reflecting limited geospatial focus.

\noindent\textbf{Object Classification:}
Object type classification in remote sensing, such as identifying ship and aircraft types, is vital for maritime and airspace monitoring. 
GPT-4o achieves superior performance in object classification, probably benefiting from extensive training data and advanced model design that supports fine-grained recognition, while Ferret \cite{you2023ferret} performs lower in this task as it prioritizes region-based spatial localization over explicit object type classification. For detailed results, we refer to Fig.~\ref{fig:results_heatmap}.

\noindent\textbf{Object Localization \& Counting:}
In remote sensing, this category involves locating and counting objects like water bodies, buildings, and trees. The task is challenging due to diverse scenes, varying object scales, and the need for high spatial resolution. Counting is framed as a classification task with multiple answer choices.

LLaVA-OneVision \cite{li2024llavaone}  performs relatively better in tasks such as \emph{building/vehicle counting, specific vehicle/aircraft type counting, marine debris counting, and tree health assessment}, demonstrating the ability to capture fine details.
Despite its remote sensing focus, RS-LLaVA \cite{bazi2024rs} underperforms in \emph{marine debris counting task}.
LHRS-Bot-Nova \cite{muhtar2024lhrs} performs better in the \emph{tree counting} task among geospatial models.
In addition to Fig.~\ref{fig:results_heatmap} results, we present the precision for the referring expression detection task in Table~\ref{tab:ref_detection}. 
These results demonstrate that Sphinx achieves the relatively highest precision scores at \emph{IoU: 0.25 \& 0.50} compared to other models within this category while GPT-4o performs the worst.
Among geospatial models, EarthDial \cite{soni2024earthdial} performs the best and is the second-best overall for this task.
A lower IoU value captures the model's approximate localization ability and its effectiveness in interpreting task-specific instructions, while an IoU of 0.5 reflects the standard setting for this task.

\begin{table}[t]
\centering
\setlength{\tabcolsep}{2pt} 
\renewcommand{\arraystretch}{1.0} 
\fontsize{8}{10}\selectfont
\resizebox{\columnwidth}{!}{
\begin{tabular}{l|c|c|c|c|c}
\hline
\textbf{Model} & \textbf{Crop} & \textbf{Damaged } & \textbf{Disaster } & \textbf{Farm } & \textbf{Land } \\
 & \textbf{Cls} & \textbf{Bldg Cnt} & \textbf{Cls} & \textbf{Pond CD} & \textbf{Use Cls} \\
\hline\hline

EarthDial\cite{soni2024earthdial} & \textbf{0.2182} & 0.4362 & 0.5727 & \textbf{0.2105} & \textbf{0.6623} \\ 
GPT-4o\cite{openai2024gpt4o} & 0.1818 & \textbf{0.5667} & \textbf{0.6300} & 0.1711 & 0.6525 \\ 
LLaVA-OneVision\cite{li2024llavaone} & 0.1455 & 0.4810 & 0.4537 & 0.1842 & 0.5869 \\ 
Qwen2-VL\cite{wang2024qwen2} & 0.1091 & 0.5000 & 0.5991 & 0.1974 & 0.5967 \\

\hline
\end{tabular}}
\vspace{-0.1in}
\caption{VLM performance on temporal geospatial tasks. Evaluation spans crop type classification, disaster type classification, farm pond change detection (CD), land use classification, and damaged building counting. 
EarthDial performs best in land use classification, while GPT-4o achieves better performance in disaster classification and damaged building counting. 
Qwen2-VL stands second in disaster classification.
}
\vspace{-0.1in}
\label{tab:temporal_accuracy_scores}
\end{table}

\noindent\textbf{Event Detection:}
Event detection involves identifying disasters crucial for environmental monitoring and risk management. Tasks such as fire risk assessment and disaster classification play a key role in mitigation and response efforts.
Qwen2-VL \cite{wang2024qwen2} show potential in fire risk assessment, while EarthDial \cite{soni2024earthdial} performs better in disaster classification. We also report overall category results in Tab.~\ref{tab:accuracy_scores}.

\noindent\textbf{Caption Generation:}
In this category, we evaluate image captioning using BERTScore\cite{bert-score}, which measures the semantic similarity between generated captions and reference descriptions. 
Sphinx \cite{lin2023sphinx} achieves the highest BERTScore\cite{bert-score}, followed closely by SkySenseGPT \cite{luo2024skysensegpt} and GPT-4o, indicating their strong ability to generate semantically rich and contextually relevant captions (Tab.~\ref{tab:accuracy_scores}).
GeoChat \cite{kuckreja2024geochat}, have the lowest scores, suggesting limitations in capturing fine-grained visual details.
Our captions include both fine-grained and high-level scene descriptions. Since Sphinx is trained with detailed visual grounding, it outperforms other models, demonstrating its effectiveness in generating precise and contextually relevant captions.

\noindent\textbf{Semantic Segmentation:}
Including referring segmentation in our benchmark enhances the evaluation of query-driven segmentation in remote sensing, crucial for land-use mapping and urban analysis. While no remote sensing-specific models currently support this task, non-specialized models like GlaMM \cite{rasheed2024glamm} achieve a baseline mIoU of 0.1411. This integration helps assess adaptability of models to spatial queries in remote sensing.

\noindent\textbf{Temporal Understanding:}
GEOBench-VLM also incorporates temporal tasks to analyze changes over time, an essential aspect of remote sensing.
We evaluate GPT-4o, Qwen2-VL \cite{wang2024qwen2}, EarthDial \cite{soni2024earthdial}, and LLaVA-OneVision \cite{li2024llavaone} on five tasks: crop classification, damaged building counting, disaster type classification, farm pond change detection, and land use classification (Table~\ref{tab:temporal_accuracy_scores}).
Overall GPT-4o emerges as the top performer in disaster classification and damaged building counting, while Qwen2-VL trails GPT-4o.
EarthDial performs best in land use classification.
While temporal information has the potential to enhance performance, it is not yet fully exploited by existing models. 
The low accuracy in change detection and crop classification suggests that current VLMs struggle to leverage temporal dependencies effectively. 
Future improvements should focus on incorporating long-term temporal reasoning to better capture changes in geospatial data.

\noindent\textbf{Non-Optical:}
In the non-optical domain, we evaluate land use classification and earthquake impact estimation. Table~\ref{tab:non_optical} shows Qwen2-VL \cite{wang2024qwen2} scores highest in earthquake magnitude estimation, while GPT-4o ranks lowest. For land use classification, GPT-4o performs best, with GeoChat \cite{kuckreja2024geochat} and RS-LLaVA \cite{bazi2024rs} performing comparably but struggling with earthquake magnitude estimation.

\begin{table}[t]
    \centering
    \begin{minipage}[t]{0.49\linewidth}
        \centering
        \setlength{\tabcolsep}{2pt} 
        \renewcommand{\arraystretch}{1.0} 
        \fontsize{7}{8}\selectfont
        \begin{tabular}{l|c|c}
        \hline
        \textbf{Model} & \textbf{Prec@0.5} & \textbf{Prec@0.25} \\
        \hline\hline
        Sphinx \cite{lin2023sphinx} & \textbf{0.3408} & \textbf{0.5289} \\
        EarthDial \cite{soni2024earthdial}  & 0.2429 & 0.4139 \\
        GeoChat \cite{kuckreja2024geochat} & 0.1151 & 0.2100 \\
        Ferret \cite{you2023ferret} & 0.0943 & 0.2003 \\
        Qwen2-VL \cite{wang2024qwen2} & 0.1518 & 0.2524 \\
        GPT-4o \cite{openai2024gpt4o} & 0.0087 & 0.0386 \\
        LHRS-Nova \cite{muhtar2024lhrs} & 0.0930 & 0.2423 \\
        SkySenseGPT \cite{luo2024skysensegpt} & 0.1082 & 0.3224 \\
        \hline
        \end{tabular}
        \caption{Referring expression detection. We report Precision on 0.5 IoU and 0.25 IoU.}
        \label{tab:ref_detection}
    \end{minipage}
    \hfill
    \begin{minipage}[t]{0.49\linewidth}
        \vspace{-3.3em}
        \centering
        \setlength{\tabcolsep}{2pt} 
        \renewcommand{\arraystretch}{1.2} 
        \fontsize{7}{8}\selectfont
        \begin{tabular}{l|c|c}
        \hline
        \textbf{Model} & \textbf{EQME} & \textbf{LU Cls} \\
        \hline\hline
        RS-LLaVA \cite{bazi2024rs} & 0.0863 & 0.3123 \\
        GeoChat \cite{kuckreja2024geochat} & 0.1403 & 0.3156 \\
        Qwen2-VL \cite{wang2024qwen2} & \textbf{0.2734} & 0.2525 \\
        GPT-4o \cite{openai2024gpt4o} & 0.0827 & \textbf{0.3256} \\
        \hline
        \end{tabular}
        \vspace{0.1in}
        \caption{Performance comparison of different models on Non-Optical tasks. EQME: EarthQuake Magnitude Estimation}
\label{tab:non_optical}
    \end{minipage}
\end{table}

%% file: sec/4_analysis.tex
\begin{figure}[t]
    \centering
    \includegraphics[width=\linewidth]{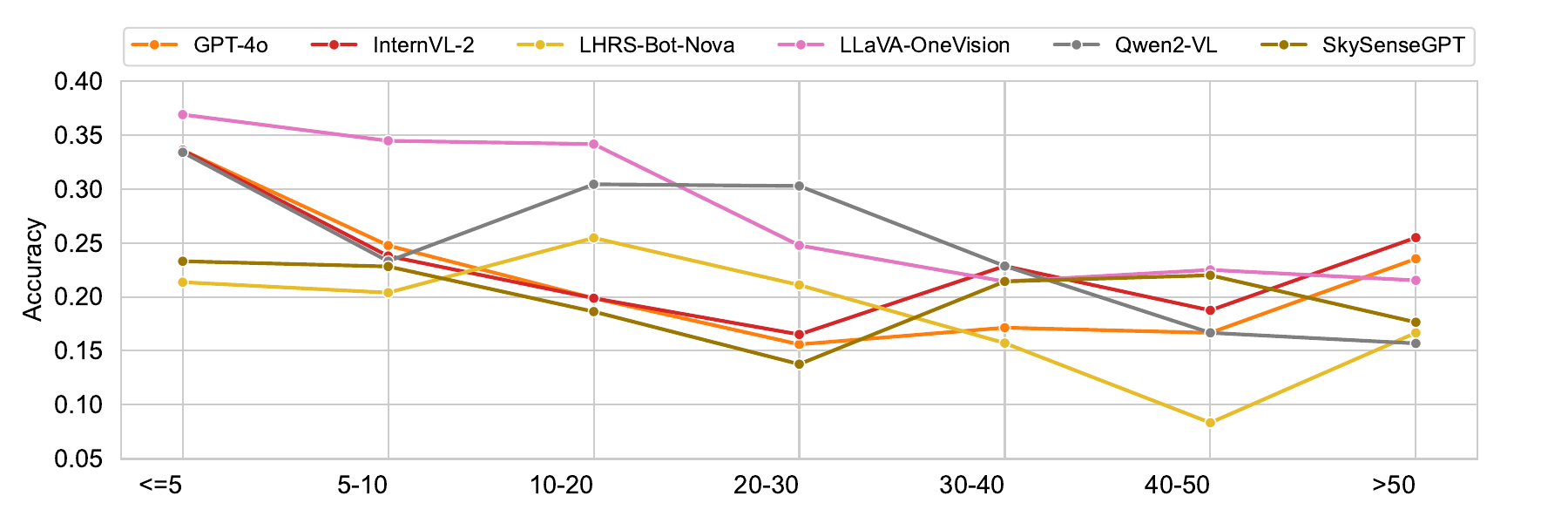}
    \captionsetup{justification=justified}
    \vspace{-0.2in}
    \caption{Object Density vs. Counting Accuracy. VLMs are evaluated on how well they maintain counting accuracy as the number of objects increases. LLaVA-OneVision shows better performance in less dense ranges, whereas Qwen2-VL, LHRS-Bot-Nova, and SkySenseGPT exhibit more significant performance drops at higher object densities.}
    \vspace{-0.1in}
    \label{fig:counting_bining_chart}
\end{figure}

\begin{figure}[t]
    \centering
    \begin{minipage}{0.49\linewidth}
        \centering        \includegraphics[width=\linewidth]{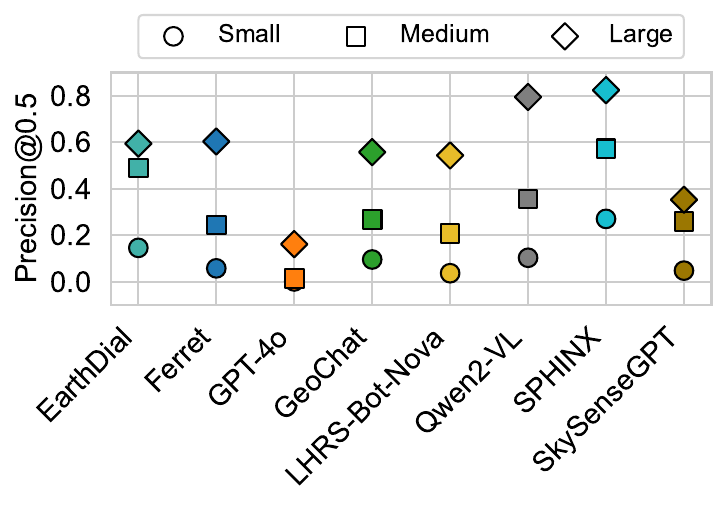}
        \caption{Performance comparison of different models on Referring Expression Detection across various object sizes.}
        \vspace{-0.1in}
    \label{fig:size_wise_precision}
    \end{minipage}
    \hfill
    \begin{minipage}{0.48\linewidth}
        \centering
    \includegraphics[width=\linewidth]{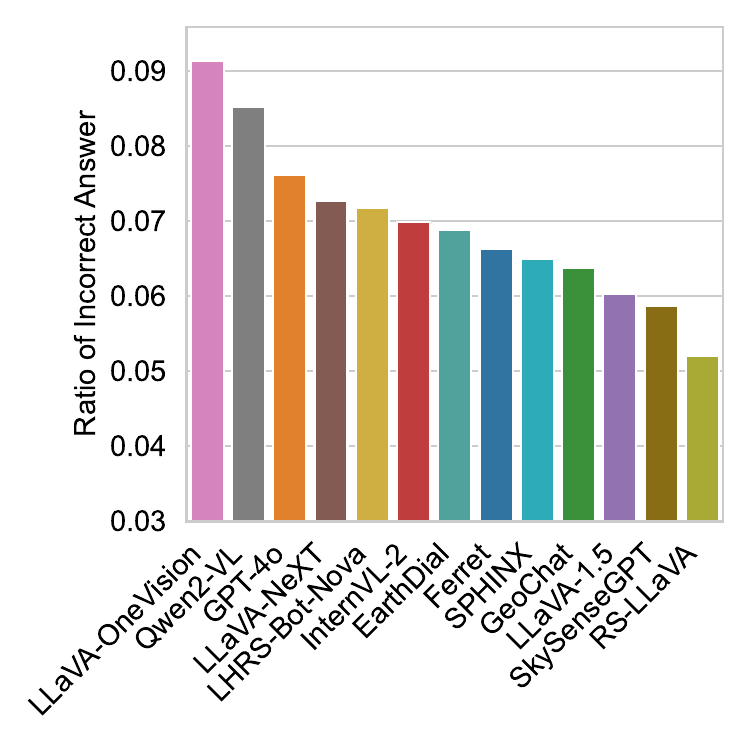}
    \vspace{-.8cm}
        \caption{Percentage of incorrect answers falling within the 20\%
          error range}
          \vspace{-0.1in}
          \label{fig:counting_bining_chart_20}
    \end{minipage}
\end{figure}

\begin{figure}[t]
    \centering
    \includegraphics[width=1.0\linewidth]{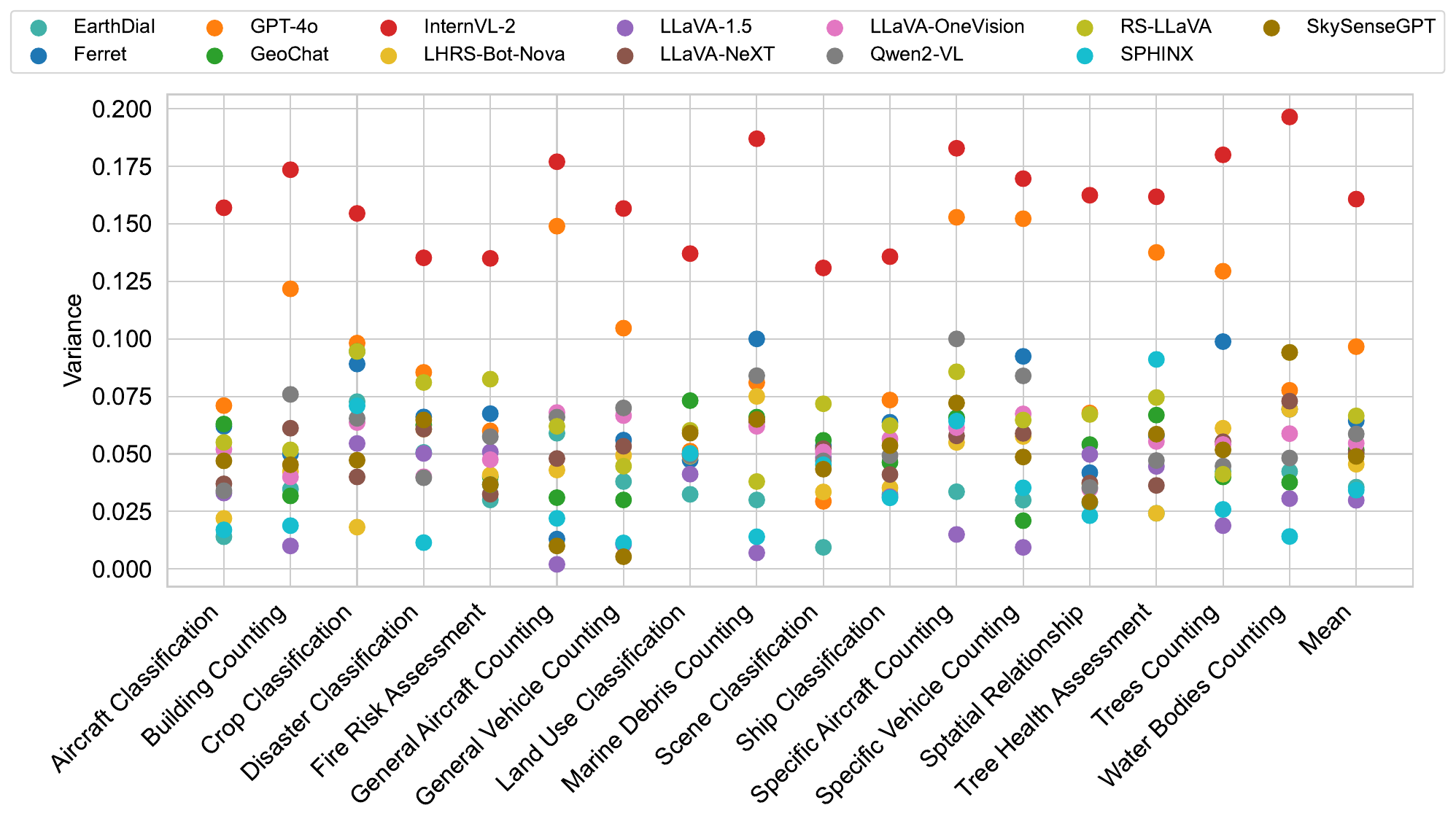}
    \captionsetup{justification=justified}
    \vspace{-0.2in}
    \caption{Accuracy Variance across different prompts.}
    \label{fig:variance_plot}
\end{figure}

\begin{figure}[t]
    \centering
    \includegraphics[width=1.0\linewidth]{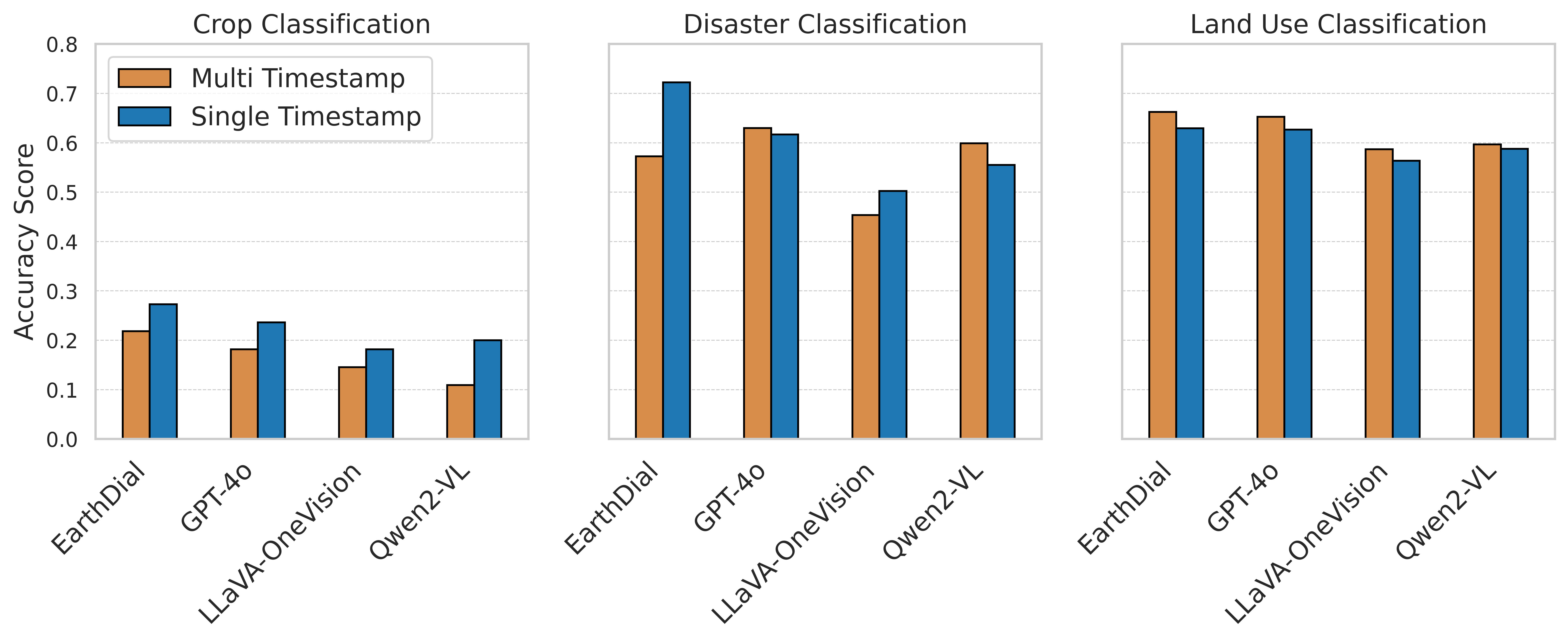}
    \captionsetup{justification=justified}
    \vspace{-0.2in}
    \caption{Single vs Multi-temporal.}
    \vspace{-0.15in}
    \label{fig:single_vs_temporal}

\end{figure}

\section{Analysis}
\label{sec:analysis}
\noindent\textbf{Counting Accuracy by Object Count Range:}
The object count ranges ($\le$10 to $>$50) illustrate how well each model maintains counting accuracy from sparse to highly dense scenes (Fig.~\ref{fig:counting_bining_chart}). 
LLaVA-OneVision \cite{li2024llavaone} leads from low to moderate densities ($\le$5 up to 20). InternVL2 \cite{chen2024internvl} and GPT-4o perform best in the highest density ($>$50) category. Qwen2-VL \cite{wang2024qwen2} remains competitive in mid-density scenes but drops at higher object counts, and SkySenseGPT \cite{luo2024skysensegpt} usually exhibits the accuracy decline as density grows. These results emphasize the need for robust spatial alignment and fine-grained detection when counting objects in diverse density conditions.

\noindent\textbf{Incorrect Answer Percentage by Option Range Distribution:}
Fig.~\ref{fig:counting_bining_chart_20} illustrates the ratio of incorrect responses when answer choices deviate from the ground truth within 20\%. 
All models, including LLaVA-OneVision, Qwen2-VL, GPT-4o, and LHRS-Bot-Nova, exhibit that these models are sensitive to shifts in the distribution of possible options. 
Notably, no model demonstrates stable performance as the gap between the correct and alternative answers increases, suggesting weak numerical reasoning capabilities in counting tasks. 
This inconsistency highlights a critical limitation in VLMs when precise object count is required, making them prone to errors in complex geospatial scenes.

\noindent\textbf{Prompt-Based Performance Variance:}
We find that models exhibit different sensitivity levels to prompt variations, affecting task accuracy.
In Fig.~\ref{fig:variance_plot}, it is shown, GPT-4o and InternVL2 \cite{chen2024internvl} exhibiting high variance across multiple tasks, suggesting a stronger sensitivity to prompt changes. Models such as Qwen2-VL \cite{wang2024qwen2} and RS-LLaVA \cite{bazi2024rs} appear in the mid-range, while LLaVA-NeXT \cite{liu2024llavanext}, EarthDial \cite{soni2024earthdial} and SkySenseGPT \cite{luo2024skysensegpt} generally show lower variance, indicating stability under varied prompt structures.

\noindent\textbf{Single vs. Multi-Temporal Data:}
For the single vs. multi-temporal data analysis (Fig.~\ref{fig:single_vs_temporal}), we evaluate GPT-4o, LLaVA-OneVision \cite{li2024llavaone}, EarthDial \cite{soni2024earthdial}, and Qwen2-VL \cite{wang2024qwen2} across three tasks: crop, disaster, and land use classifications.
In crop classification, multi-temporal data result in slightly lower performance across all models, suggesting that the temporal information introduces variability that the models struggle to effectively capture.
In disaster classification, GPT-4o, EarthDial, and Qwen2-VL benefit from multi-temporal data, whereas LLaVA-OneVision exhibits a slight decline.
In land use classification, all models show improved performance with multi-temporal data, likely due to the stability of land use patterns over time, enabling better generalization for this task.

\noindent\textbf{Impact of Object Size on Detection Performance:}
In referring expression detection across different object sizes (Fig.~\ref{fig:size_wise_precision}), we evaluate multiple models on small, medium, and large objects. 
Sphinx \cite{lin2023sphinx} and Qwen2-VL \cite{wang2024qwen2} demonstrate relatively high performance, particularly excelling in detecting large objects. GPT-4o, however, performs the worst, struggling significantly with all sized objects, suggesting limitations in fine-grained localization.
Ferret \cite{you2023ferret}, EarthDial \cite{soni2024earthdial}, LHRS-Bot-Nova \cite{muhtar2024lhrs} and GeoChat \cite{kuckreja2024geochat} exhibit moderate performance on large objects.
For medium objects, EarthDial \cite{soni2024earthdial} and Sphinx \cite{lin2023sphinx} perform overall better.
These results highlight the challenges in detection, especially in medium and small objects.

%% file: sec/5_conclusion.tex
\section{Conclusion}
\label{sec:conclusion}
While generic VLMs have seen much progress in developing comprehensive benchmarks covering various reasoning aspects, the earth-observation domain lacks large-scale manually verified benchmarks covering diverse capabilities. 
Unlike existing benchmarks that lack the unique requirements of geospatial data, GEOBench-VLM is designed specifically for 31 remote-sensing tasks like flood detection, disaster monitoring, crop classification, marine debris counting, and temporal change segmentation. 
GEOBench-VLM covers a range of computer vision tasks, ranging from referring expression segmentation to counting, detection, recognition, and temporal analysis. 
Our benchmark, featuring over 10,000 manually verified instructions across diverse conditions, highlights the limitations of current VLMs in geospatial contexts, even for state-of-the-art models. 
Through detailed experiments with the 13 best performing VLMs, we show the current gaps that need to be addressed to unlock several remote-sensing applications.

%% file: sec/X_suppl.tex
\renewcommand{\thefigure}{A\arabic{figure}}
\setcounter{figure}{0}  
\renewcommand{\thetable}{A\arabic{table}}
\setcounter{table}{0}

\newcounter{secnumber}
\setcounter{secnumber}{1} 
\renewcommand{\thesection}{S\arabic{secnumber}}

\twocolumn[{
\begin{center}
    {\large \textbf{GEOBench-VLM: Benchmarking Vision-Language Models for Geospatial Tasks}} \\[1em]
    {\LARGE Supplementary Material}
\end{center}
}]

\noindent This supplementary material includes the dataset table (\ref{sec:datasettab}) and quantitative results to illustrate multiple cases for assessing model responses (\ref{sec:failure}). 
It also provides results on multispectral images (\ref{sec:multispec}) 
along with a comparison between bi-temporal and multi-temporal approaches (\ref{sec:bimulti}). 
Additionally, a geographical analysis is included (\ref{sec:geoanalysis}) to observe the span of data coverage across locations, followed by a detailed description of the word cloud (\ref{sec:wordcloud}).

\begin{table*}[t]
\centering
\setlength{\tabcolsep}{3.2pt} 
\renewcommand{\arraystretch}{1.3} 
\fontsize{9}{10}\selectfont
\begin{tabular}{p{0.21\textwidth}p{0.21\textwidth}p{0.21\textwidth}p{0.24\textwidth}p{0.05\textwidth}}
\hline
\textbf{Name} & \multicolumn{1}{c}{\textbf{Task}} & \multicolumn{1}{c}{\textbf{Annotation Type}} & \textbf{Sensor (Res)} & \textbf{Year} \\
\hline
AiRound\cite{airound} & \multirow[c]{6}{=}{\centering Scene Understanding, Object Classification} & \multirow[c]{6}{=}{\centering Class} & RGB, Sentinel-2 (10m) & 2020 \\
RESICS45\cite{resis} &  &  & RGB & 2017 \\
PatternNet\cite{PatternNet} &  &  & RGB & 2018 \\
MtSCCD\cite{MtSCCD} &  &  & RGB (1m) & 2024 \\
FireRisk\cite{FireRisk} &  &  & RGB (1m) & 2023 \\
FGSCR\cite{fgscr} &  & & RGB & 2021 \\
\hline
FAIR1M\cite{fair1m} & \multirow[c]{3}{=}{\centering Spatial Relation Classification, Referring Expression Detection, Captioning} & \multirow[c]{3}{=}{\centering Bounding Box} & RGB (0.3–0.8m) & 2021 \\
DIOR\cite{dior} &  &  & RGB & 2020 \\
DOTA\cite{xia2018dota} &  &  & RGB (0.1–1m) & 2021 \\
\hline
Forest Damage\cite{forest_damage} & \multirow[c]{5}{=}{\centering Counting} & \multirow[c]{5}{=}{\centering Bounding Box} & RGB & 2021 \\
Deforestation\cite{deforestation-satellite-imagery-335n4_dataset} &  &  & RGB & 2024 \\
COWC\cite{cowc} &  &  & RGB (15 cm) & 2016 \\
NASA Marine Debris\cite{nasa_marine_debris} &  &  & RGB (3m) & 2024 \\
The RarePlanes Dataset\cite{RarePlanes_Dataset} &  &  & RGB (0.3m) & 2020 \\
\hline
fMoW\cite{fmow} & \multirow[c]{5}{=}{\centering Temporal Understanding} & \centering Class & RGB (1m) & 2018 \\
xBD\cite{xbd} &  & \centering Bounding Box, Instance Mask, Class & RGB (0.8m) & 2019 \\
PASTIS\cite{pastis} &  & \centering Semantic Mask & MSI (10m) & 2021 \\
FPCD\cite{FPCD} &  & \centering Semantic Mask & RGB (1m) & 2022 \\
GVLM\cite{gvlm_cd} &  & \centering Class & RGB (0.6m) & 2023 \\
\hline
DeepGlobe Land Cover\cite{DeepGlobe} & \multirow[c]{2}{=}{\centering Referring Expression Segmentation} & \multirow[c]{2}{=}{\centering Semantic Mask} & RGB (0.5m) & 2018 \\
GeoNRW\cite{GeoNRW} &  &  & RGB (1m) & 2021 \\
\hline
So2Sat\cite{so2Sat} & \multirow[c]{2}{=}{\centering Non-Optical} & \centering Class & SAR, MSI (10m) & 2020 \\
QuakeSet\cite{QuakeSet} &  & \centering Number & SAR (10m) & 2024 \\
\hline
\end{tabular}
\caption{Comprehensive overview of geospatial datasets utilized for evaluating Vision-Language Models (VLMs) across diverse tasks, including Scene Understanding, Spatial Relation, Object Classification, Spatial Relation Classification, Referring Expression Detection, Captioning, Temporal Understanding, Referring Expression Segmentation, and Non-Optical tasks. The datasets are categorized by annotation types (e.g., class, bounding box, semantic mask) and sensor types (e.g., RGB imagery, Multispectral Imaging (MSI), Synthetic Aperture Radar (SAR)), highlighting their versatility for a wide range of geospatial applications.}
\label{tab:datasets}
\end{table*}

\section{Datasets}
\stepcounter{secnumber}
\label{sec:datasettab}

The datasets we use in our evaluation cover a wide range of geospatial tasks, showing the variety and depth of challenges in geospatial analysis. As shown in Table~\ref{tab:datasets}, these datasets include tasks like scene understanding, spatial relation, instance counting, temporal understanding, referring expression segmentation, and working with non-optical data. This diversity enables us to create versatile question-answer pairs tailored to each specific task. The inclusion of datasets from recent years ensures that our evaluation tackles recent challenges and uses up-to-date information.

These datasets also offer a rich variety of annotation types, sensor data, and spatial resolutions, reflecting the diverse nature of geospatial data. The annotation types range from class labels and bounding boxes to semantic and instance masks, giving different levels of detail for model evaluation. The sensor data includes RGB images, Multispectral Imaging (MSI), and Synthetic Aperture Radar (SAR), with resolutions from fine to coarse scales. This heterogeneity allows us to test models under different imaging conditions and resolutions, fostering robustness and generalizability. For example, datasets like FAIR1M \cite{fair1m}, DIOR \cite{dior}, and DOTA \cite{xia2018dota} provide high-resolution RGB images with bounding box annotations, which are critical for tasks like object detection and understanding spatial relationships in complex scenes. Temporal understanding datasets such as fMoW \cite{fmow}, xBD\cite{xbd}, PASTIS\cite{pastis}, FPCD\cite{FPCD}, and GVLM\cite{gvlm_cd} are crucial for tracking changes over time, helping in tasks such as disaster assessment and monitoring urban development. Non-optical datasets like So2Sat \cite{so2Sat} and QuakeSet \cite{QuakeSet} introduce SAR data, expanding our analysis to situations where optical imagery isn't available due to weather or lighting conditions. Scene Understanding datasets like AiRound \cite{airound} and RESICS45 \cite{resis} offer class annotations that help categorize large-scale scenes, essential for land use and land cover classification.



\section{Qualitative Results}
\stepcounter{secnumber}
\label{sec:failure}

\begin{figure*}[t]
    \centering
    \makebox[\textwidth][c]{
        \includegraphics[width=.85\textwidth]{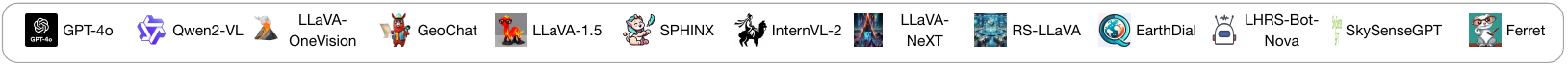}
    }
    \vspace{6pt} 
    \includegraphics[width=0.22\textwidth]{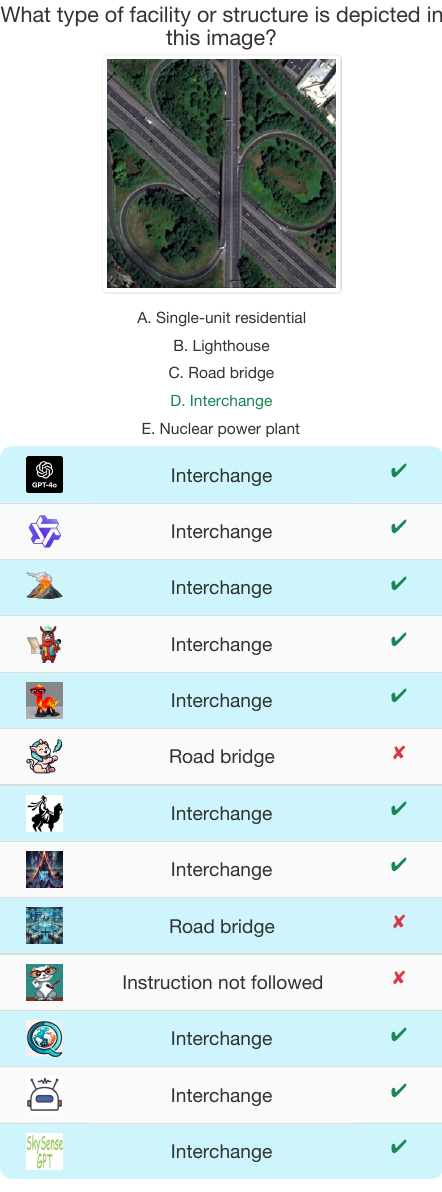}
    \includegraphics[width=0.22\textwidth]{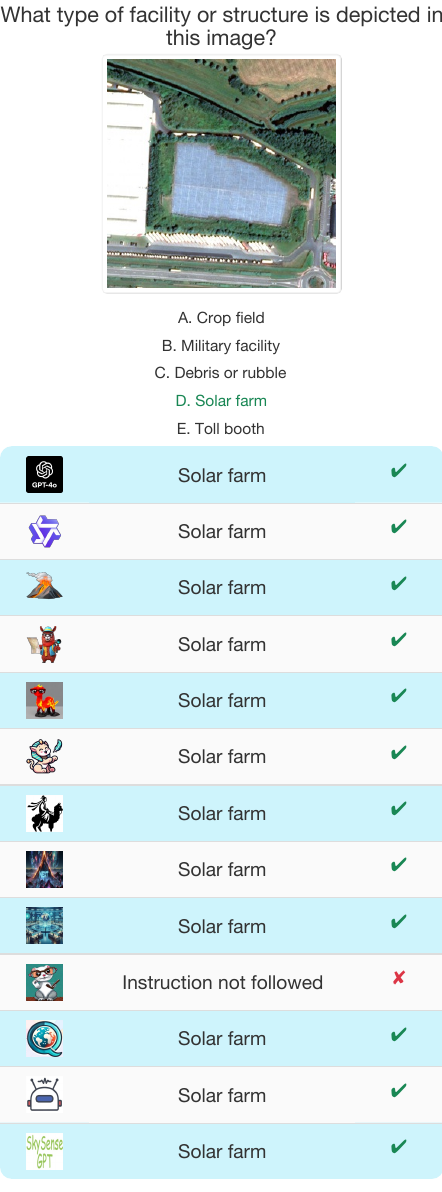}
    \includegraphics[width=0.22\textwidth]{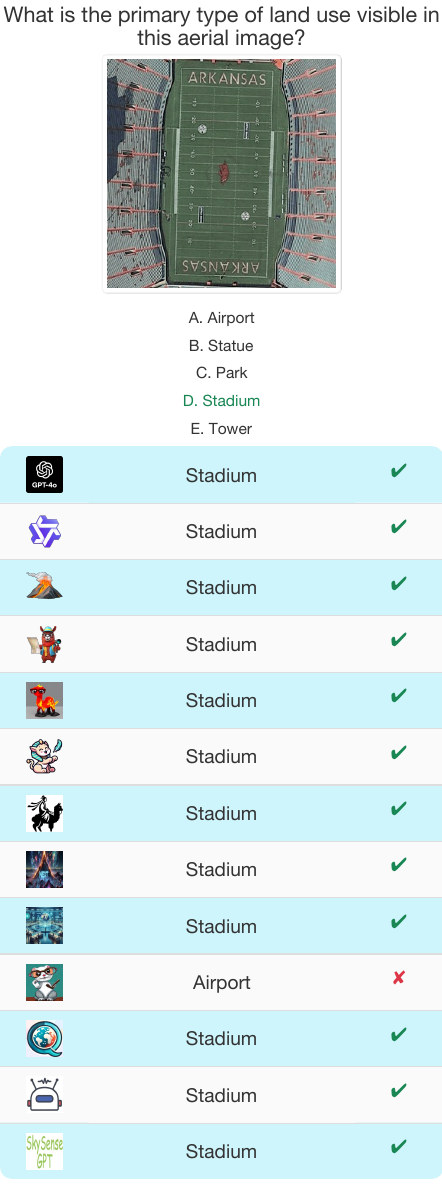}
    \includegraphics[width=0.22\textwidth]{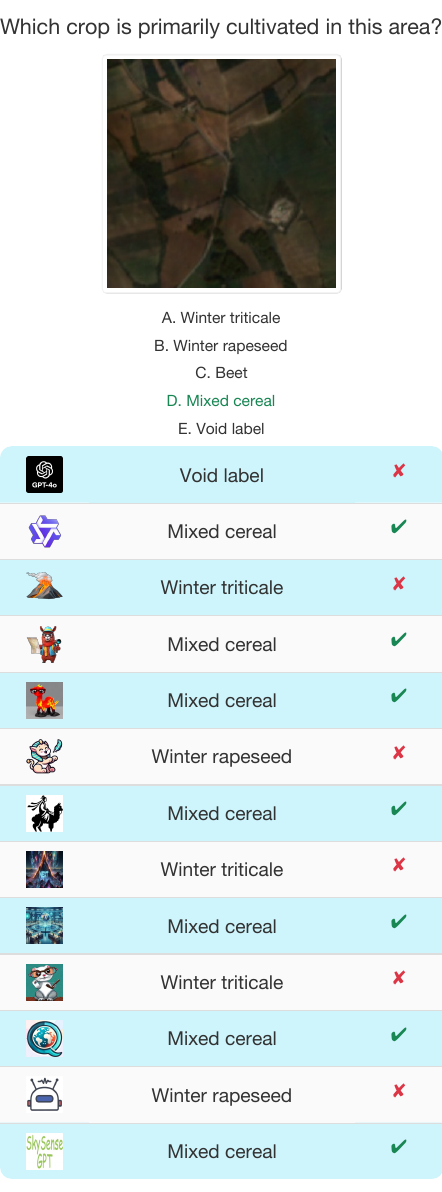}

    \includegraphics[width=0.22\textwidth]{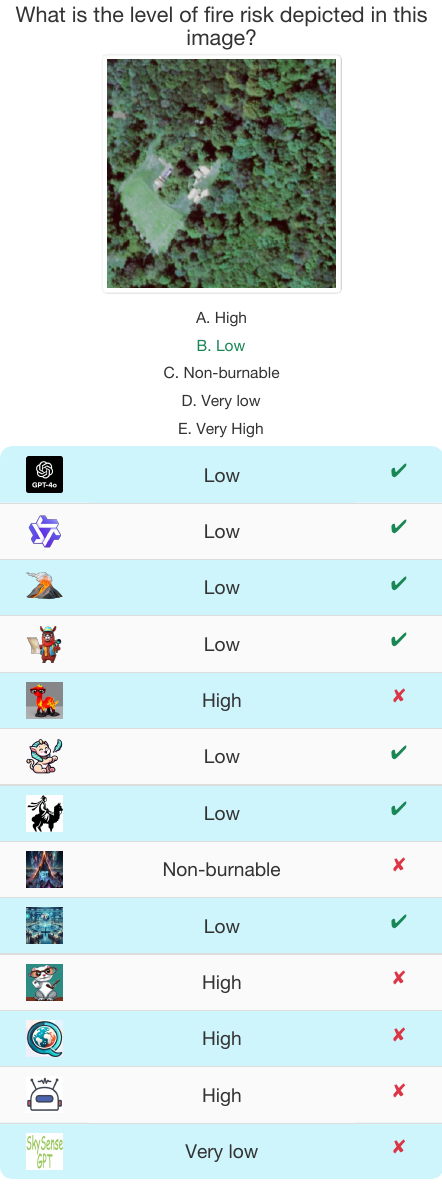}
    \includegraphics[width=0.22\textwidth]{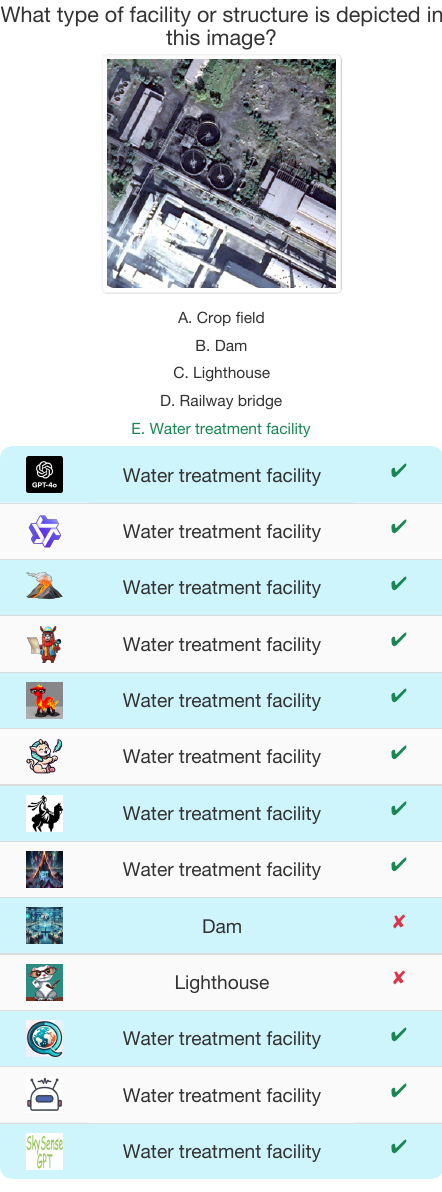}
    \includegraphics[width=0.22\textwidth]{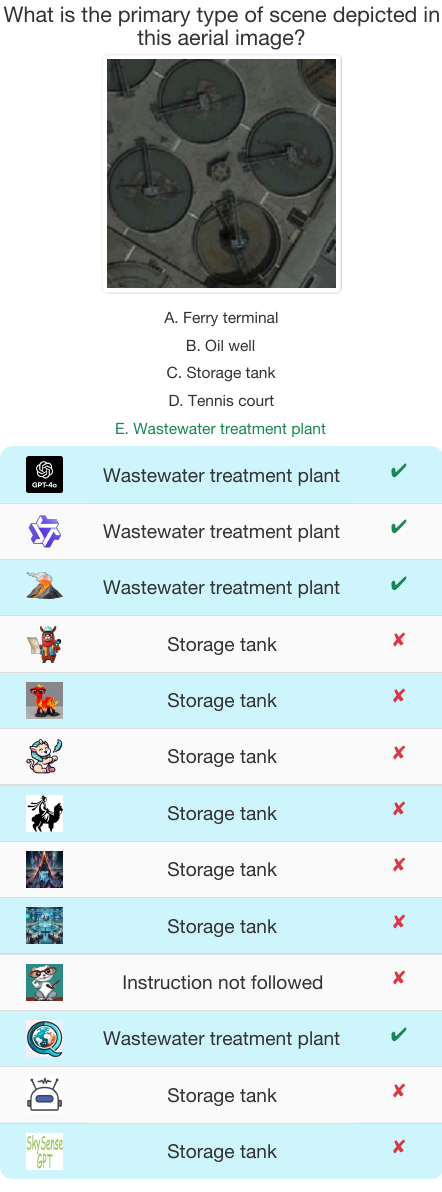}
    \includegraphics[width=0.22\textwidth]{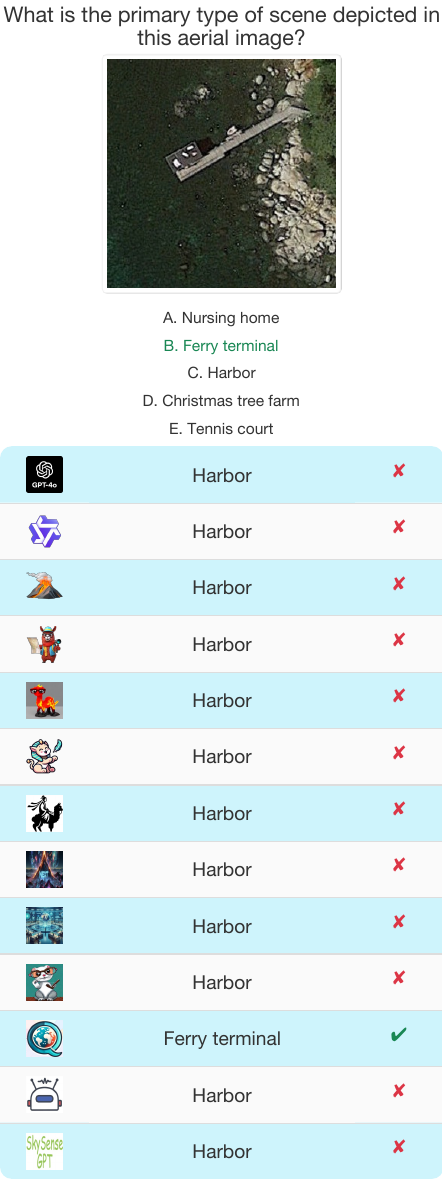}

    \caption{Scene Understanding: This illustrates model performance on geospatial scene understanding tasks, highlighting successes in clear contexts and challenges in ambiguous scenes. The results emphasize the importance of contextual reasoning and addressing overlapping visual cues for accurate classification.}
    \label{fig:scene}
\end{figure*}

\begin{figure*}[t]
    \centering
    \makebox[\textwidth][c]{
        \includegraphics[width=.85\textwidth]{images/supplementary/models_legend.pdf}
    }
    \includegraphics[width=0.22\textwidth]{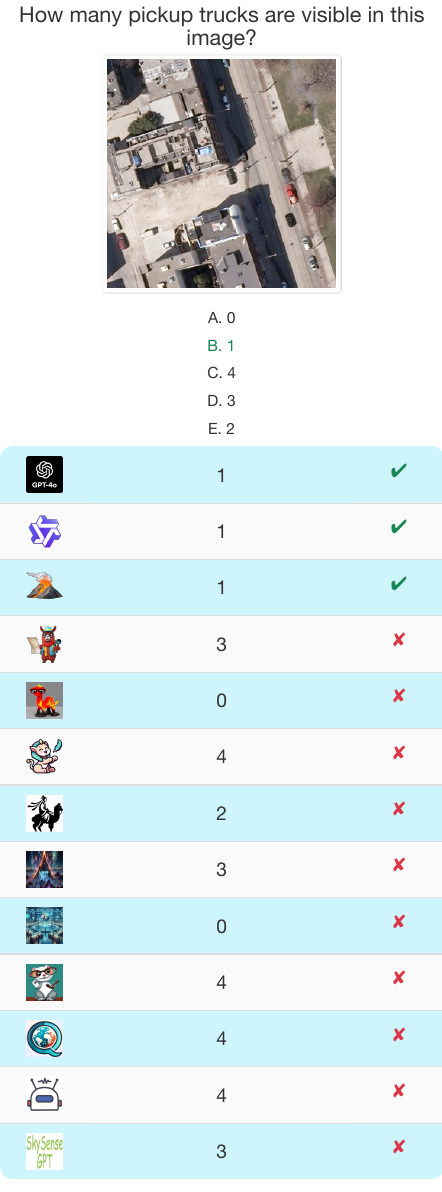}
    \includegraphics[width=0.22\textwidth]{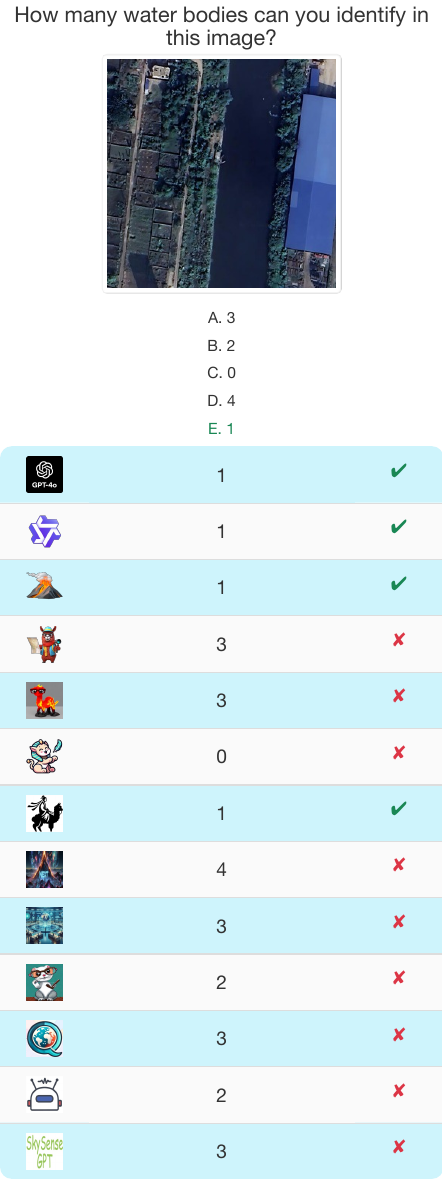}
    \includegraphics[width=0.22\textwidth]{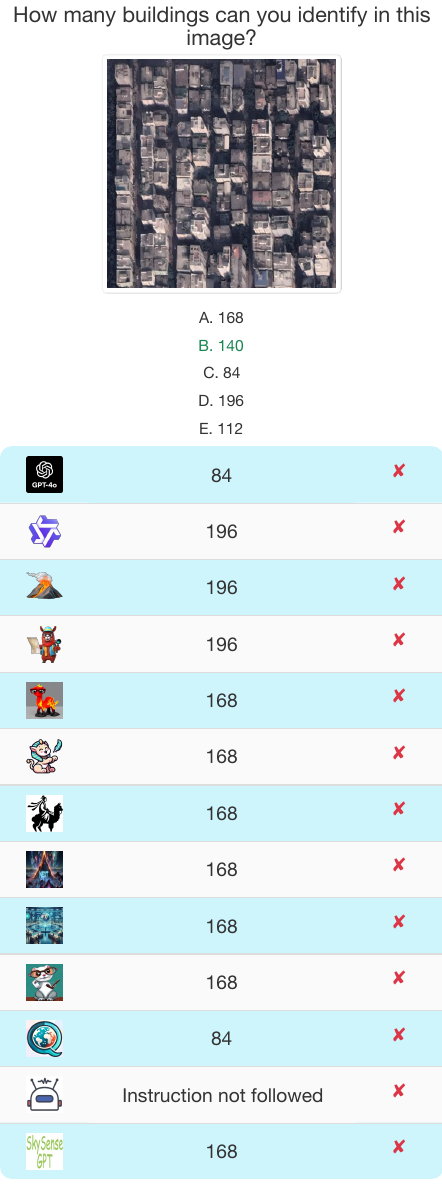}
    \includegraphics[width=0.22\textwidth]{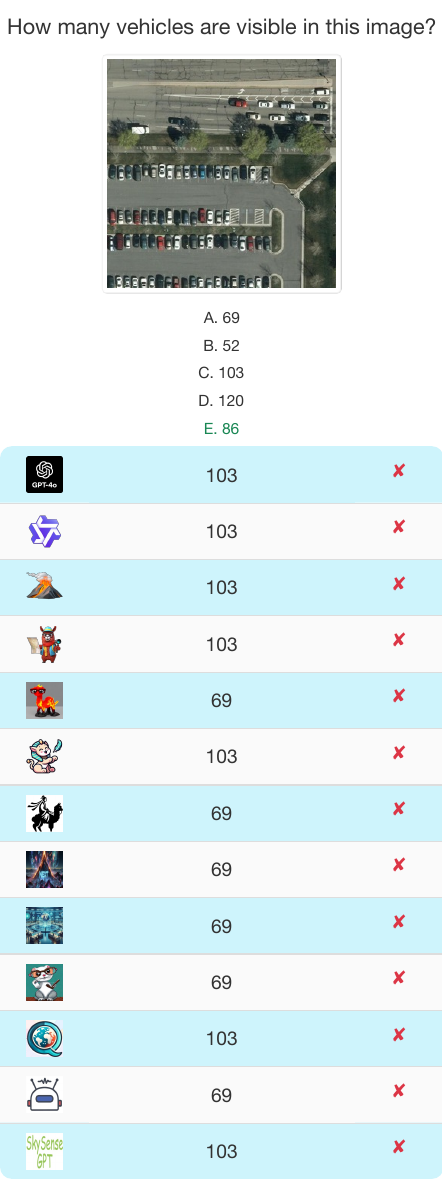}

    \includegraphics[width=0.22\textwidth]{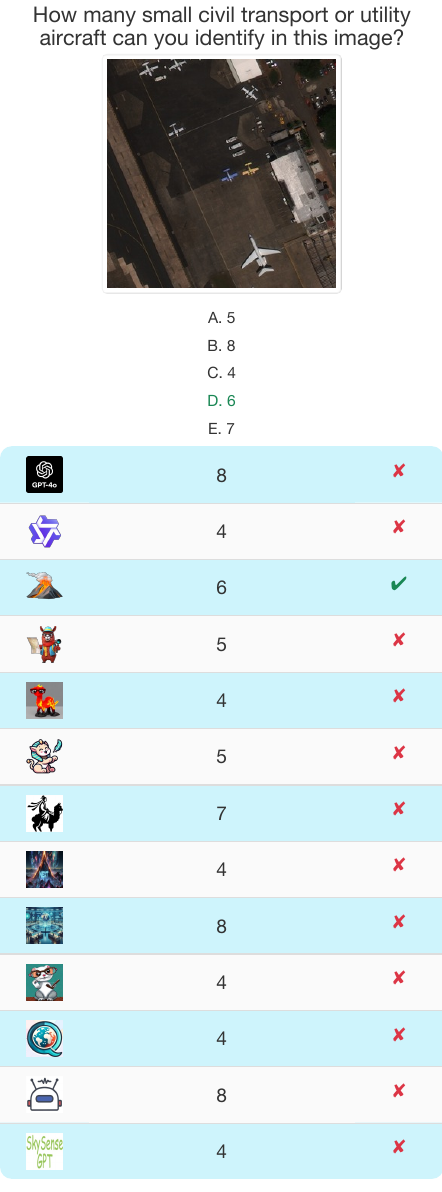}
    \includegraphics[width=0.22\textwidth]{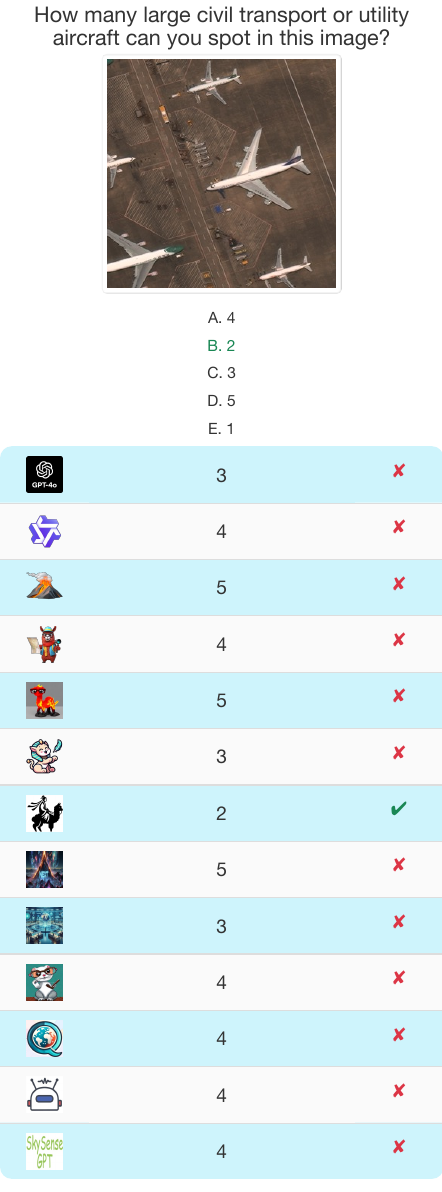}
    \includegraphics[width=0.22\textwidth]{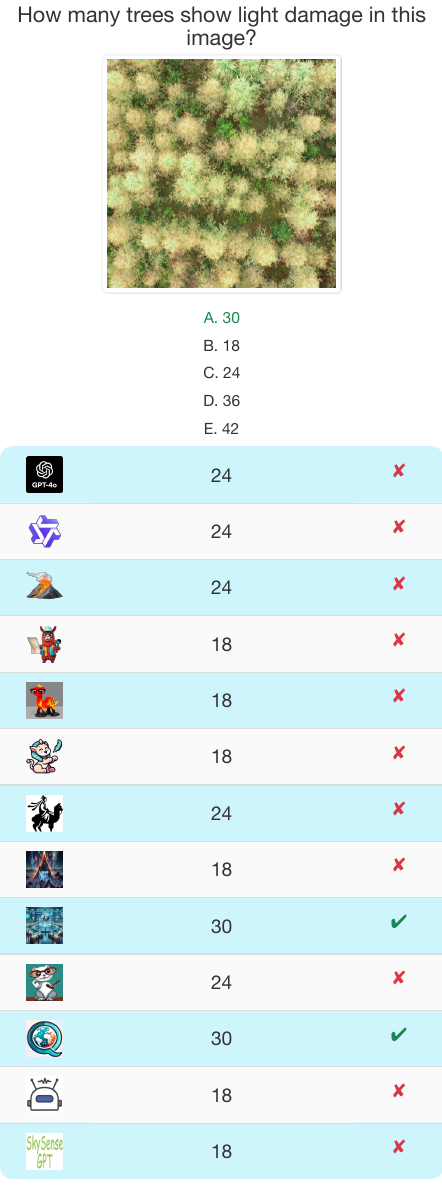}
    \includegraphics[width=0.22\textwidth]{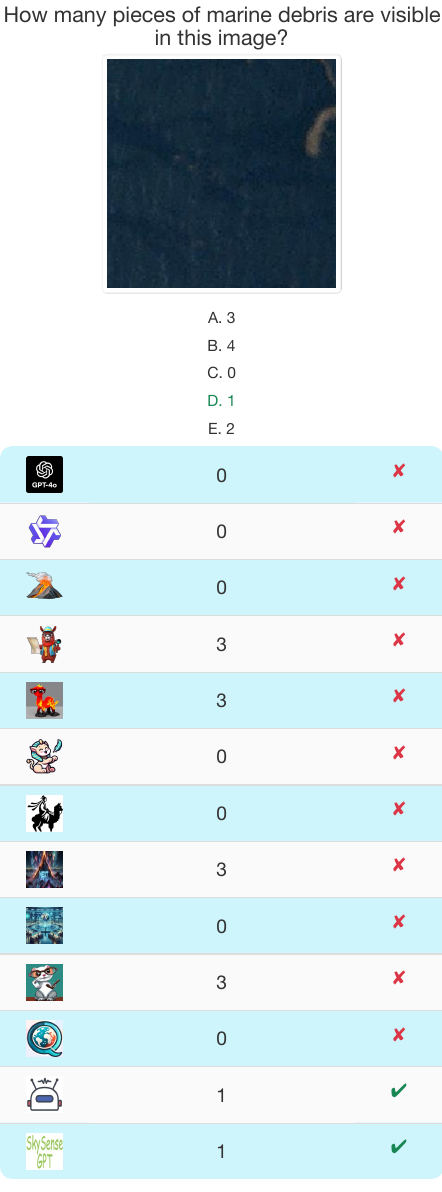}

    \caption{Counting: The figure showcases model performance on counting tasks.
    }
    \label{fig:counting}
\end{figure*}

The images in Fig.~\ref{fig:scene} show patterns in how the models performed on geospatial tasks relevant to scene understanding.
Models perform well in identifying scenes with distinct features, such as ``interchange", where most models succeeded except Ferret \cite{you2023ferret}, RS-LLaVA \cite{bazi2024rs}, and Sphinx \cite{lin2023sphinx}, respectively. 
In the third image, all models except Ferret correctly identified the ``stadium", demonstrating notable contextual understanding.
For the fourth image, only a few models correctly identified ``mixed cereal" crops, with failures attributed to the ambiguous nature of crop patterns. 
The first image in the second row shows dense greenery, indicating a moist environment, with the fire risk correctly classified as ``low".
The second image in the second row benefits from clear context, aiding classification as a water treatment facility. In contrast, the third image in the same row lacks context, making it prone to misclassification. This comparison highlights the importance of contextual information for accurate scene classification.
In the last image of Fig.~\ref{fig:scene}, ambiguous scenes such as the ``ferry terminal" where all models except EarthDial \cite{soni2024earthdial} failed, the misclassification is likely due to overlapping visual cues. The visual similarity between a harbor and a ferry terminal makes it challenging for models to differentiate between these categories. 
For the counting tasks in Fig.\ref{fig:counting}, 
almost all models struggled, with wrongly estimating due to difficulty in differentiating objects in complex environments.

\begin{figure*}[t]
    \centering

        \makebox[\textwidth][c]{
        \includegraphics[width=.85\textwidth]{images/supplementary/models_legend.pdf}
    }
    \includegraphics[width=0.22\textwidth]{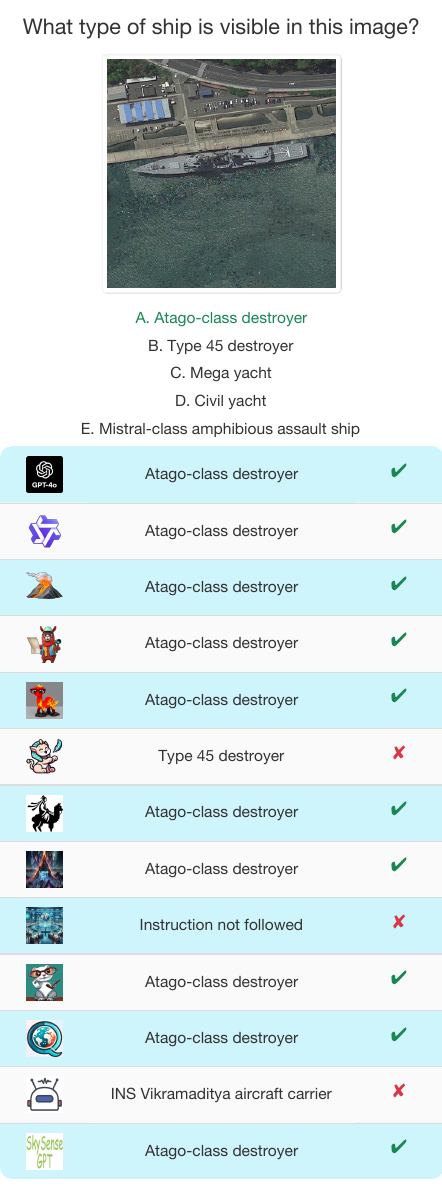}
    \includegraphics[width=0.22\textwidth]{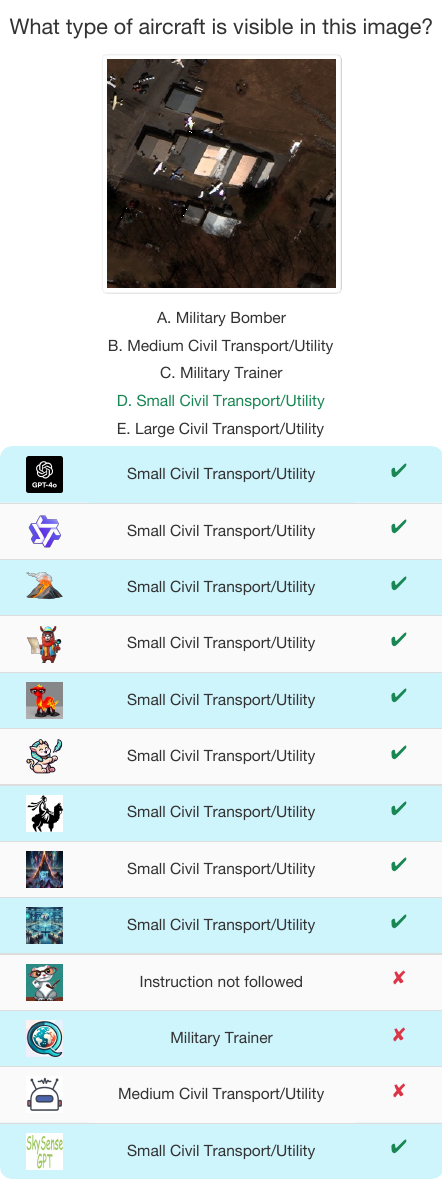}
    \includegraphics[width=0.22\textwidth]{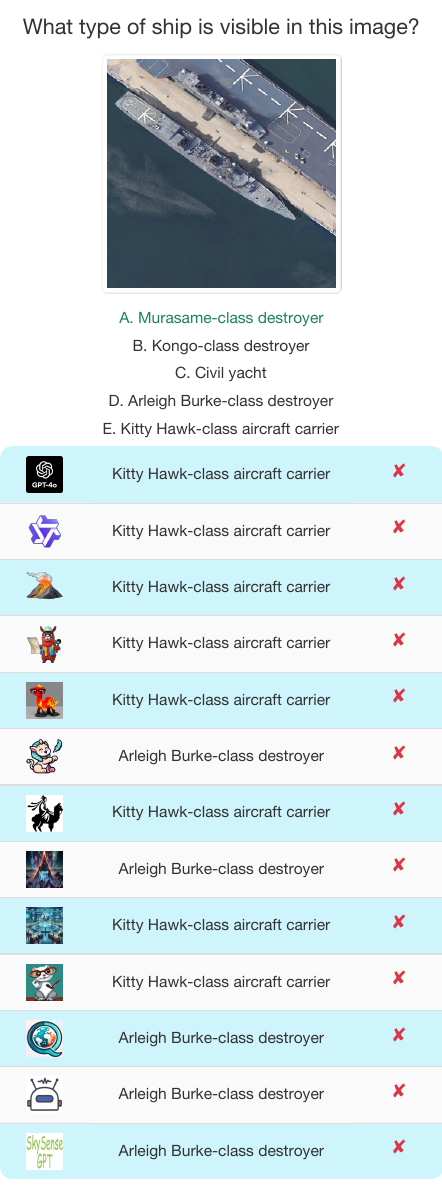}
    \includegraphics[width=0.22\textwidth]{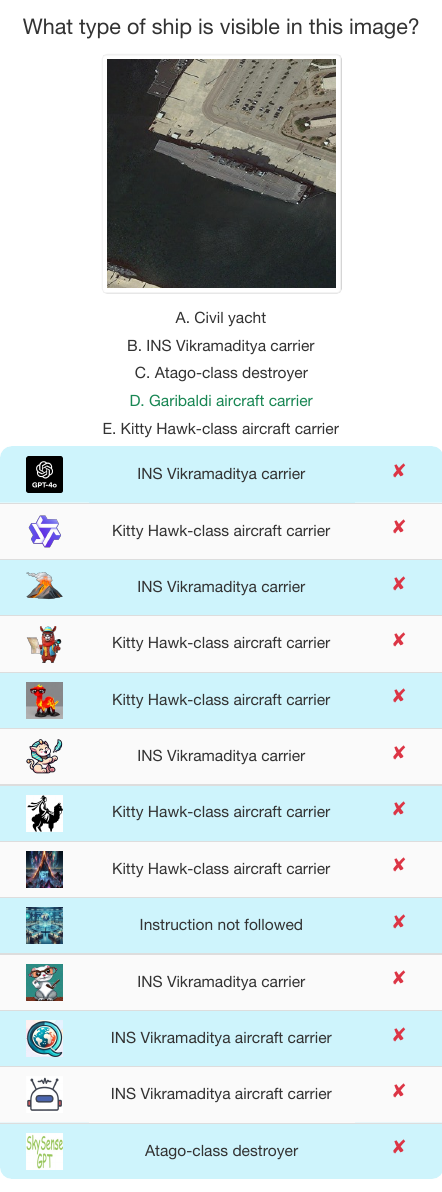}
    \caption{Object Classification: The figure highlights model performance on object classification, showing success with familiar objects like the ``atago-class destroyer" and ``small civil transport/utility" aircraft. However, models struggled with rarer objects like the ``murasame-class destroyer" and ``garibaldi aircraft carrier" indicating a need for improvement on less common classes and fine-grained recognition.}
    \label{fig:object_cls}
\end{figure*}

Fig.~\ref{fig:object_cls} shows that the models performed well in the first two images, probably due to familiar contextual clues. The ``atago-class destroyer" and ``small civil transport/utility" aircraft are common object types with distinct characteristics, making them easier for models to recognize. However, in the last two images, none of the models successfully identified the ``murasame-class destroyer" or ``garibaldi aircraft carrier" which are rarer categories. The failure is likely due to insufficient exposure to these specific classes in training datasets, coupled with the overlapping features of the objects that require advanced fine-grained recognition. 

\begin{figure*}[t]
    \centering
    \makebox[\textwidth][c]{
        \includegraphics[width=.85\textwidth]{images/supplementary/models_legend.pdf}
    }
    \includegraphics[width=0.22\textwidth]{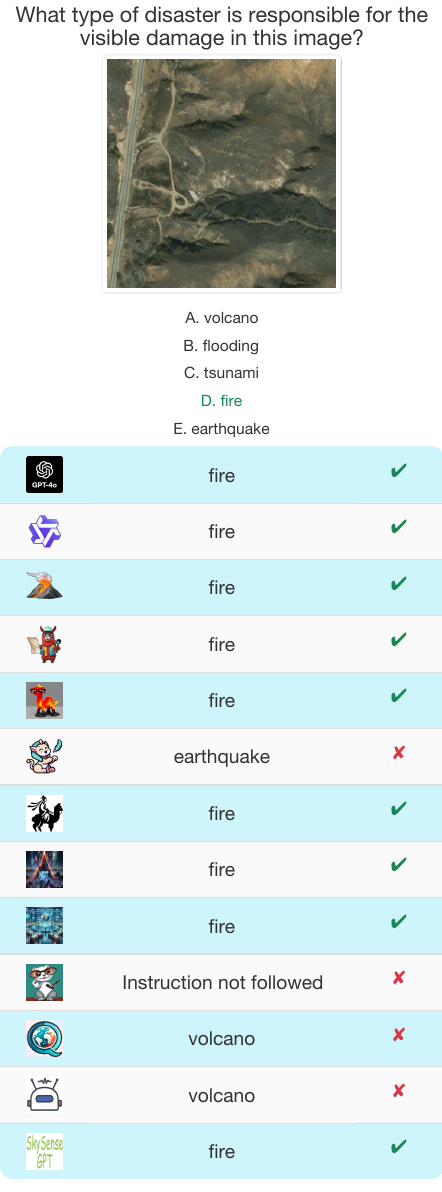}
    \includegraphics[width=0.22\textwidth]{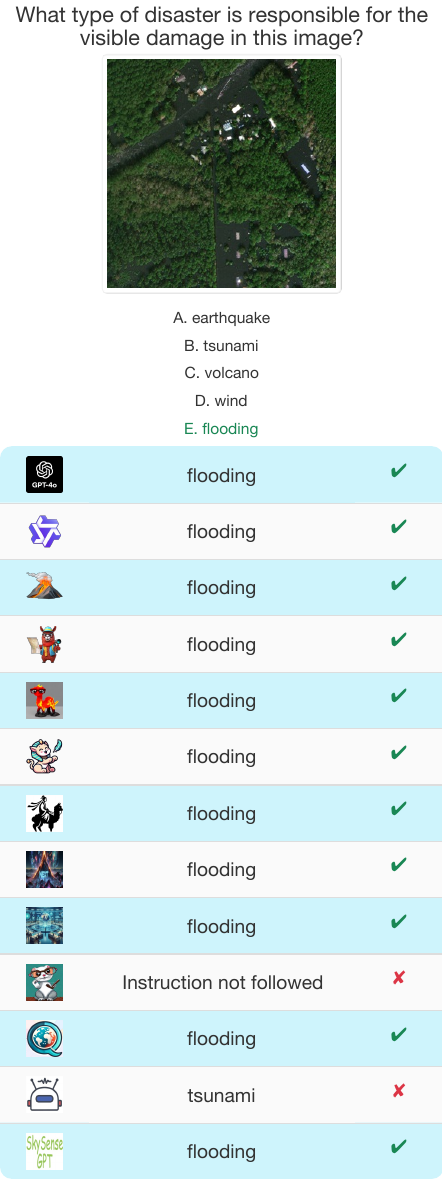}
    \includegraphics[width=0.22\textwidth]{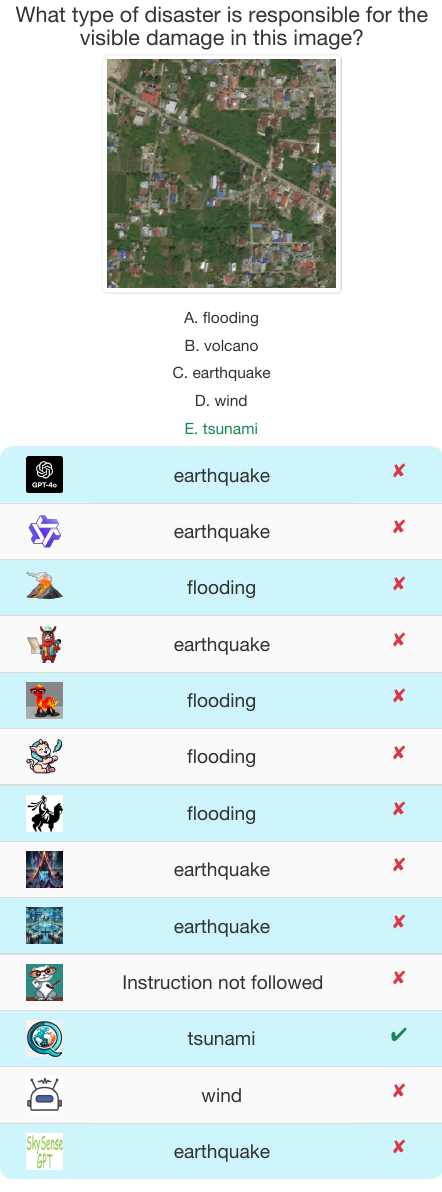}
    \includegraphics[width=0.22\textwidth]{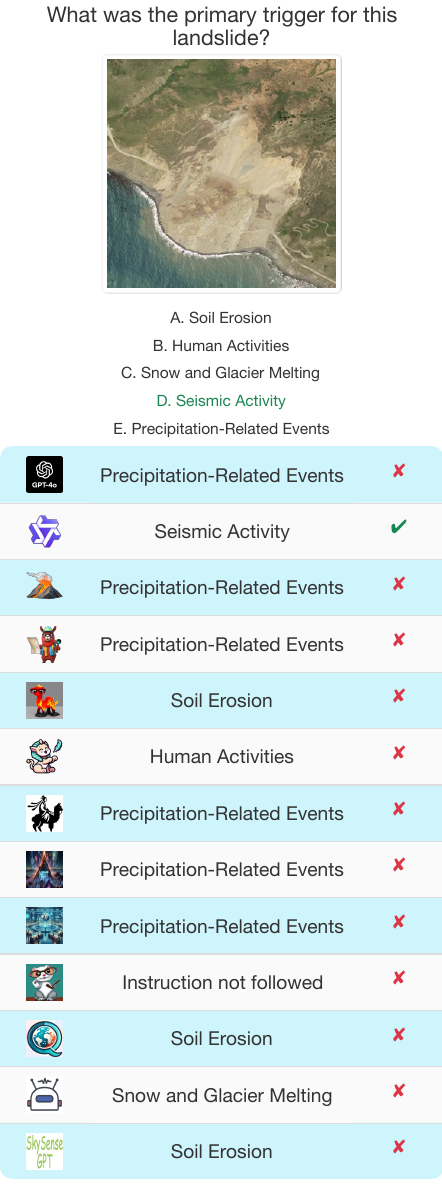}
   
    \caption{Event Detection: Model performance on disaster assessment tasks, with success in scenarios like 'fire' and 'flooding' but challenges in ambiguous cases like 'tsunami' and 'seismic activity'. Misclassifications highlight limitations in contextual reasoning and insufficient exposure on overlapping disaster features.}
    \label{fig:event_detection}
\end{figure*}

As shown in Fig.~\ref{fig:event_detection}, models performed well on disaster assessment with relatively clear indicators, such as ``fire" damage.
For the second image, depicting ``flooding", Ferret and LHRS-Bot-Nova struggled.
The third image depicts ``tsunami" damage, characterized by disrupted layouts, scattered debris, and damaged buildings, which are often visually similar to flooding. Models may misclassify this due to overlapping features, and insufficient tsunami-specific training data.
For the last image, only Qwen2VL identified the ``seismic activity", as others likely misclassified it due to overlapping features with ``precipitation-related events".

\begin{figure*}[t]
    \centering
    \makebox[\textwidth][c]{
        \includegraphics[width=.85\textwidth]{images/supplementary/models_legend.pdf}
    }

    \includegraphics[width=0.22\textwidth]{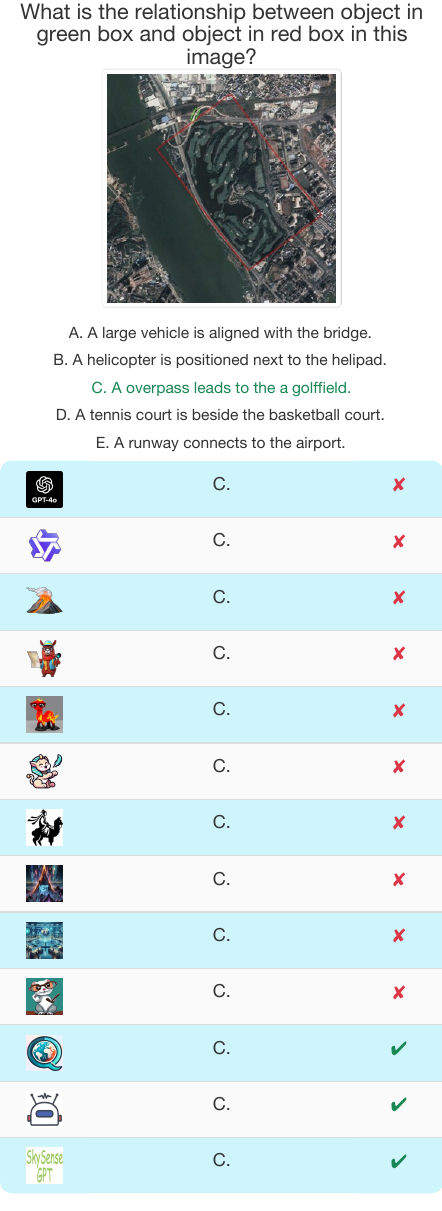}
    \includegraphics[width=0.22\textwidth]{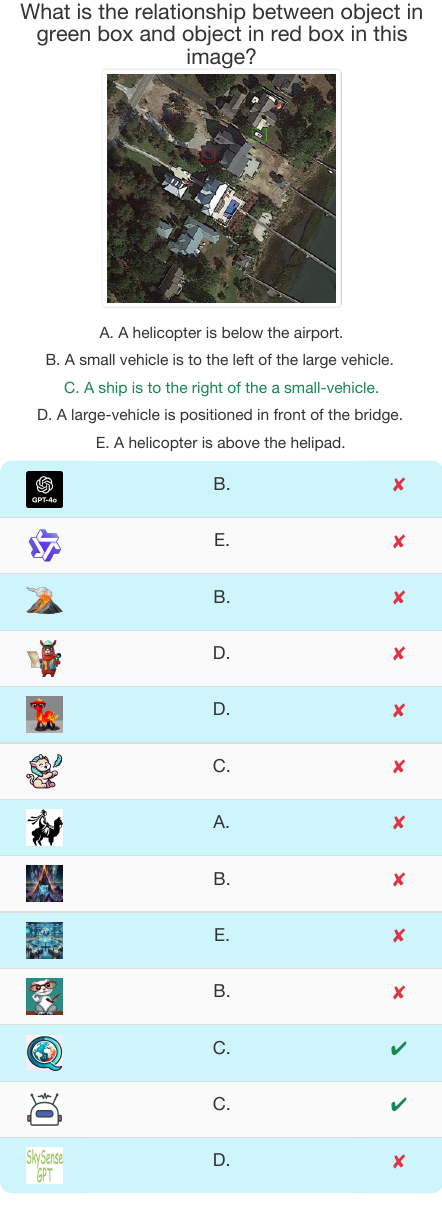}
    \includegraphics[width=0.22\textwidth]{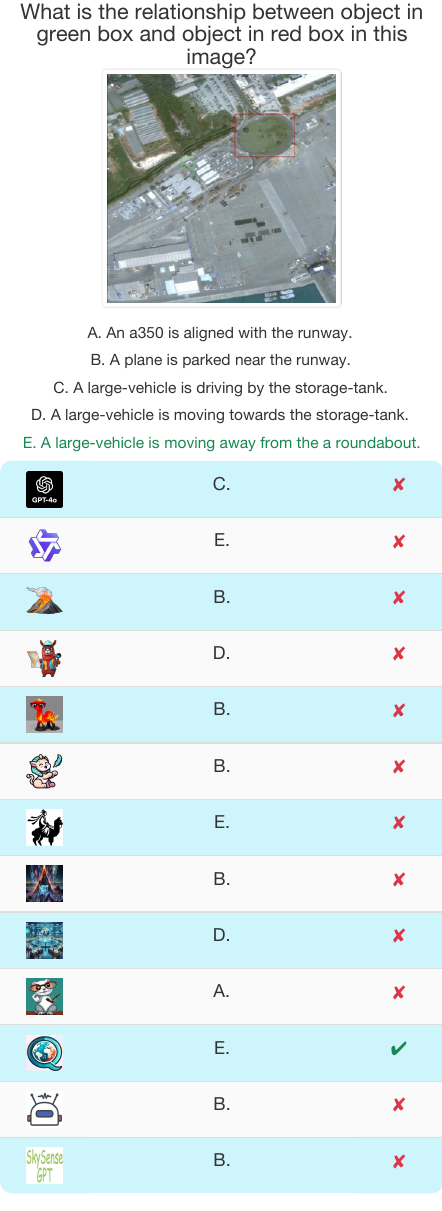}
    \includegraphics[width=0.22\textwidth]{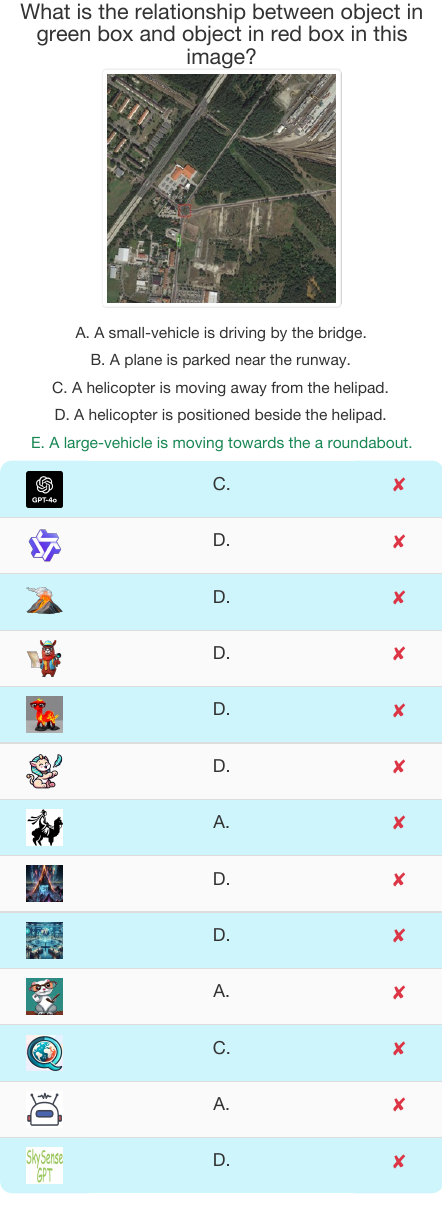}
   
    \caption{Spatial Relations: The figure demonstrates model performance on spatial relationship tasks, with success in close-object scenarios and struggles in cluttered environments with distant objects.}
    \label{fig:spatial_relations}
\end{figure*}

\begin{figure*}[t]
    \centering
    \makebox[\textwidth][c]{
        \includegraphics[width=.7\textwidth]{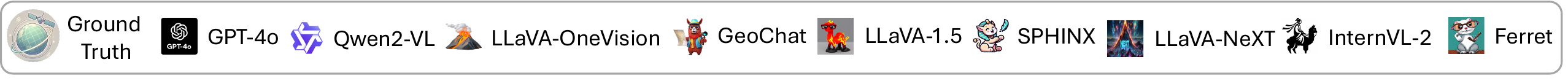}
    }
        \makebox[\textwidth][c]{
        \includegraphics[width=.5\textwidth]{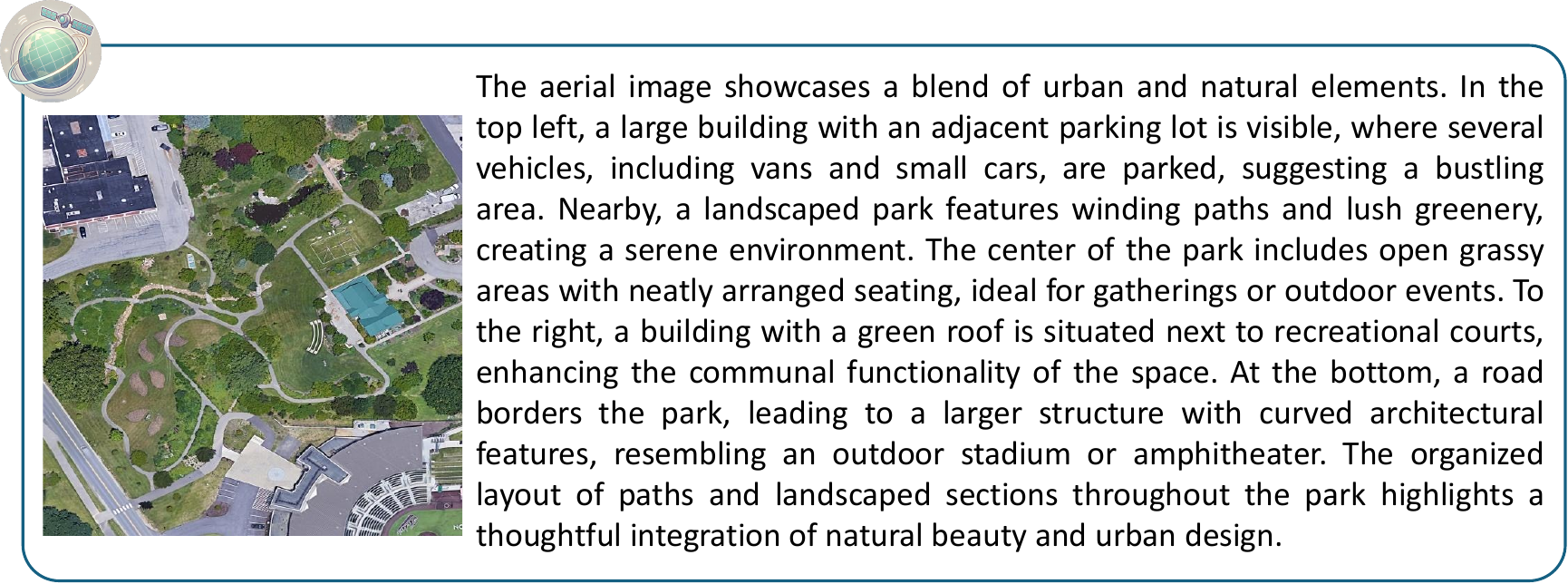}    
    }
    \includegraphics[width=0.32\textwidth]{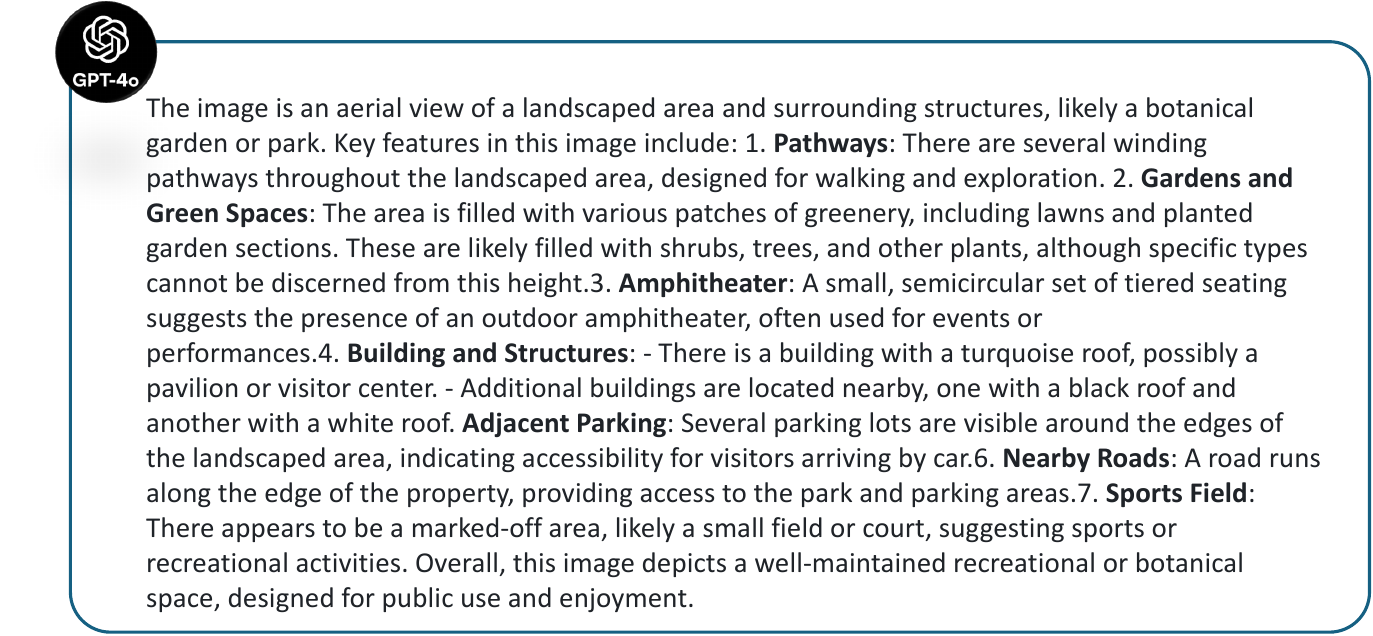}
    \includegraphics[width=0.32\textwidth]{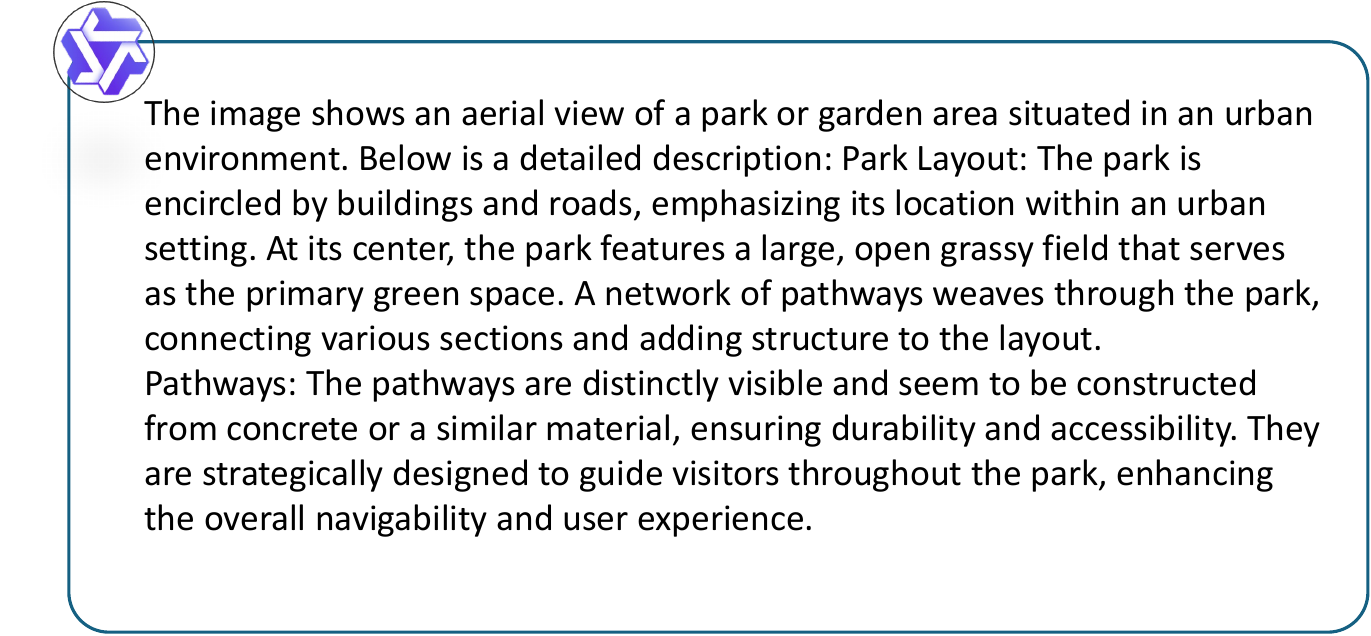}
    \includegraphics[width=0.32\textwidth]{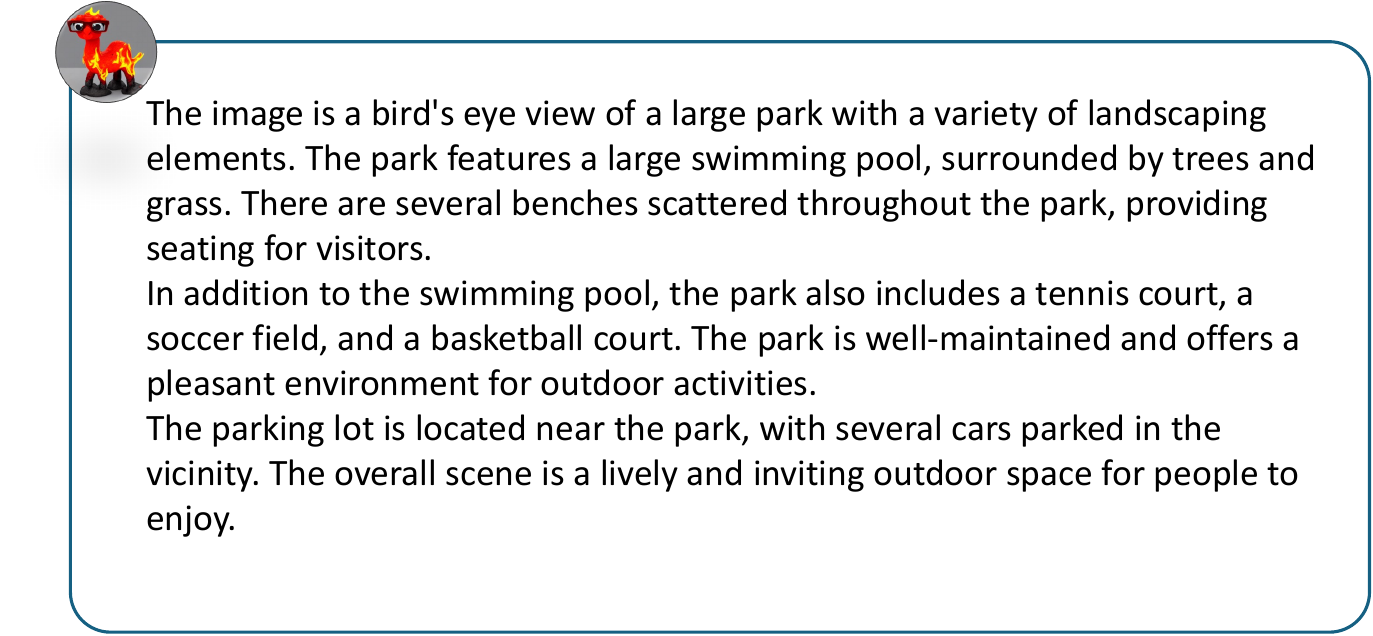}
    \vspace{5pt} 
    
    \includegraphics[width=0.32\textwidth]{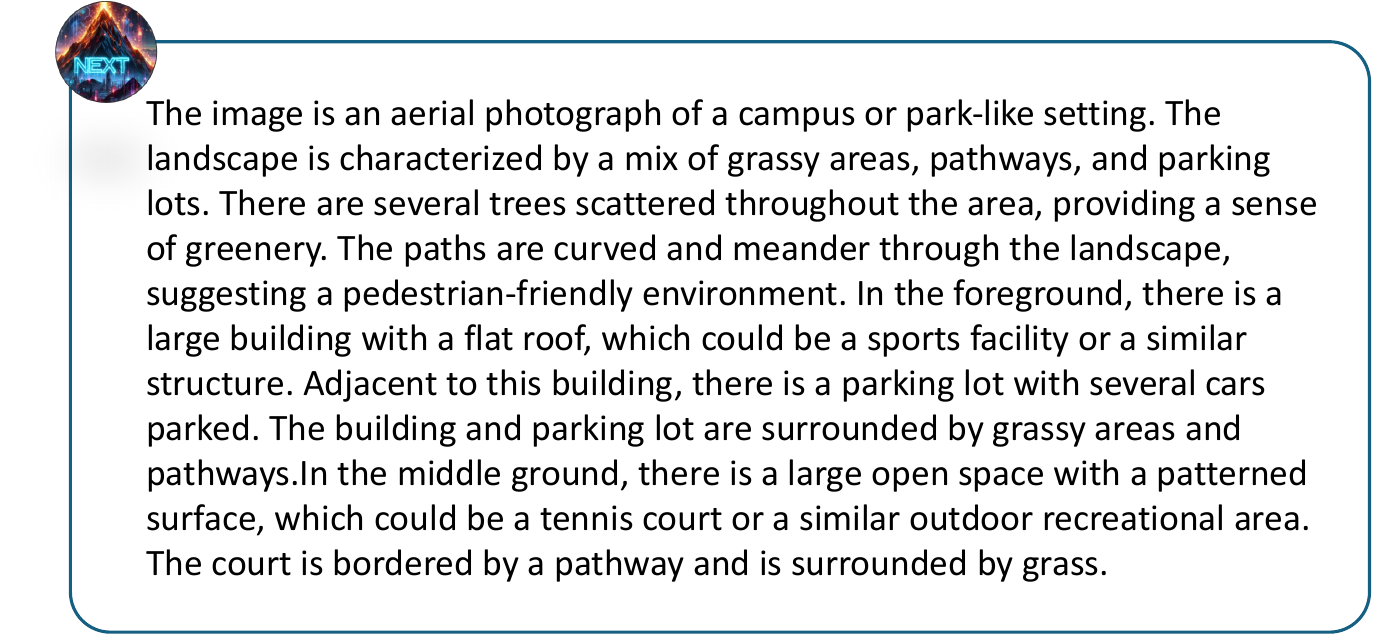}
    \includegraphics[width=0.32\textwidth]{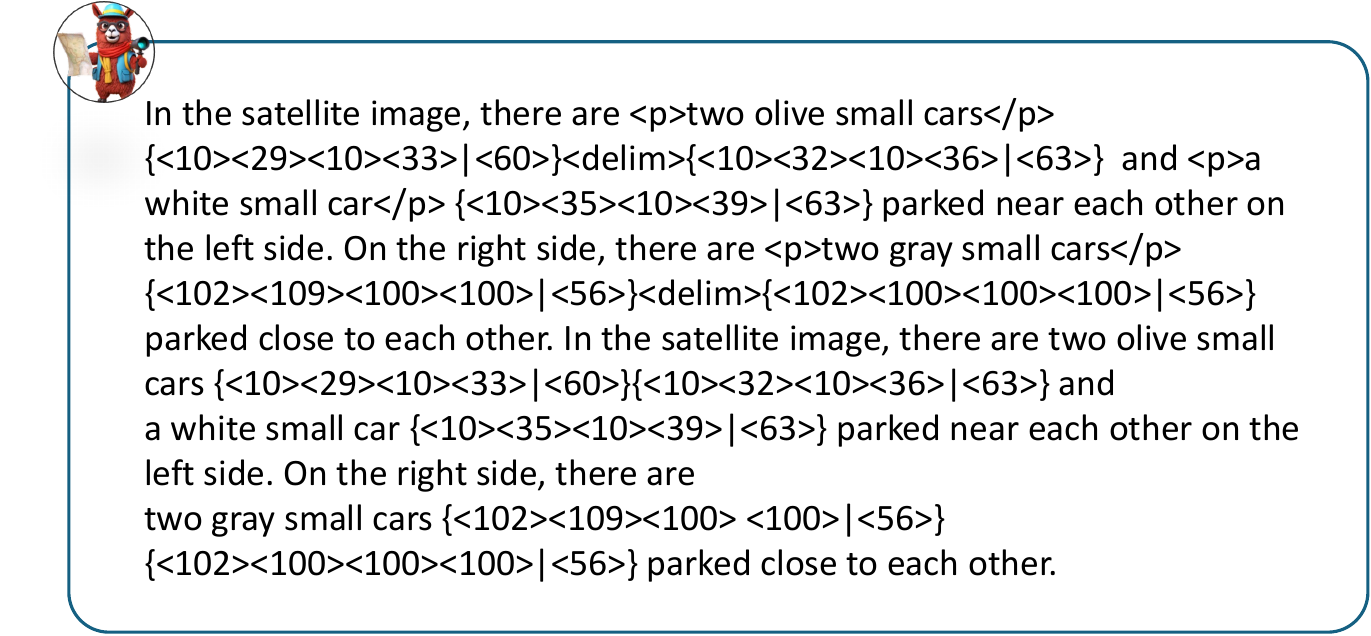}
    \includegraphics[width=0.32\textwidth]{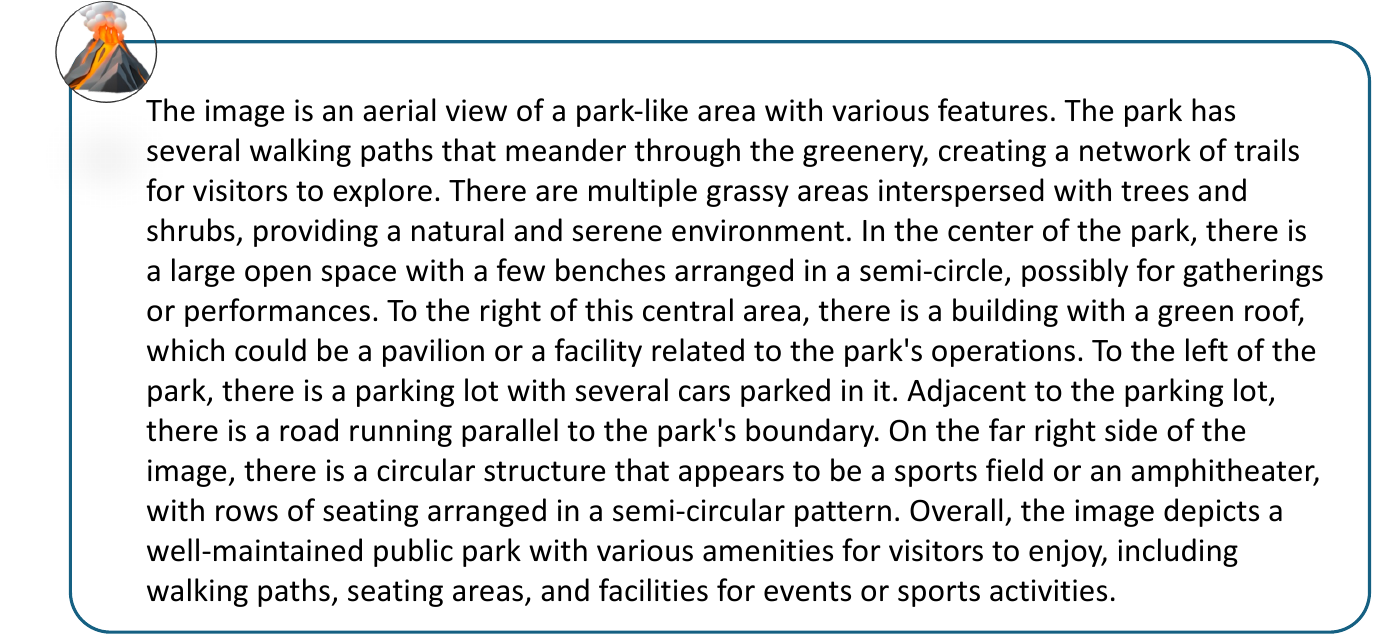}
    \vspace{5pt} 
    \includegraphics[width=0.64\textwidth]{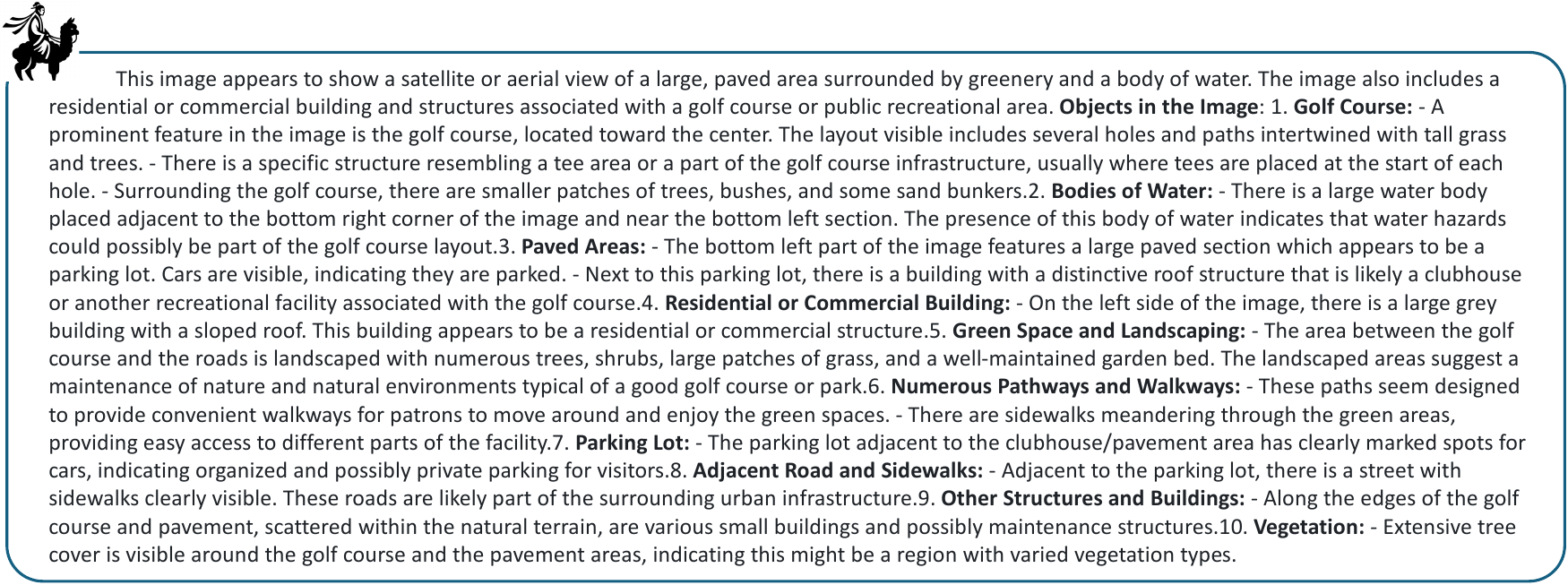}
    \includegraphics[width=0.32\textwidth]{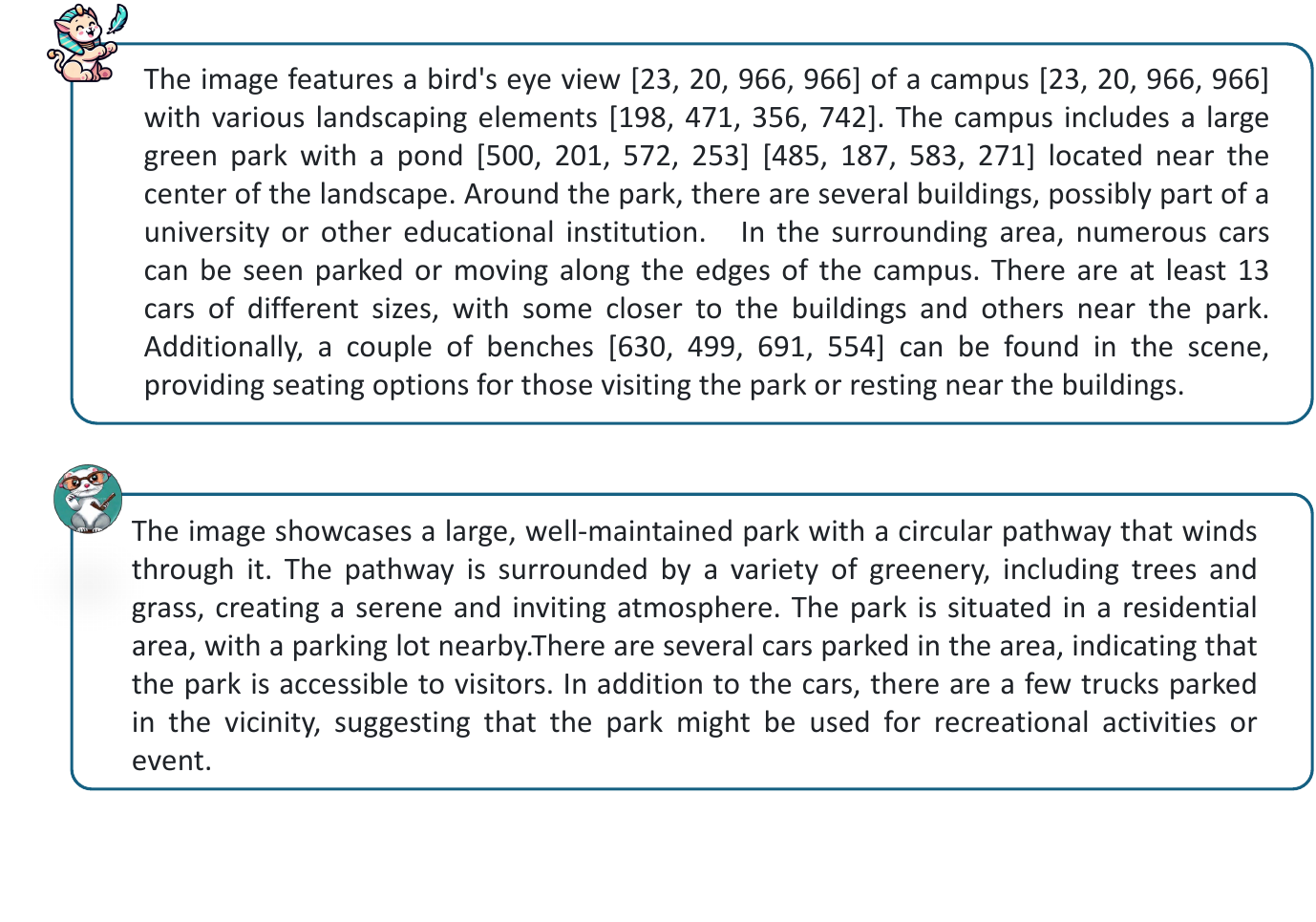}

        \vspace{3pt} 
    \includegraphics[width=0.64\textwidth]{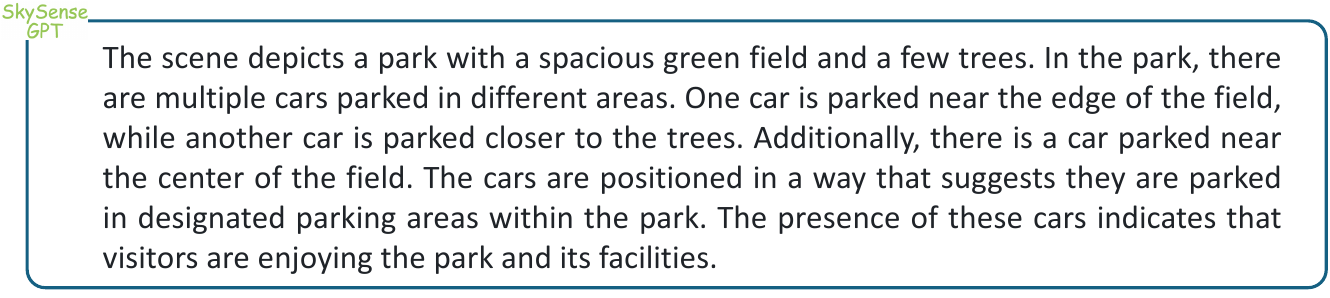}
    \includegraphics[width=0.32\textwidth]{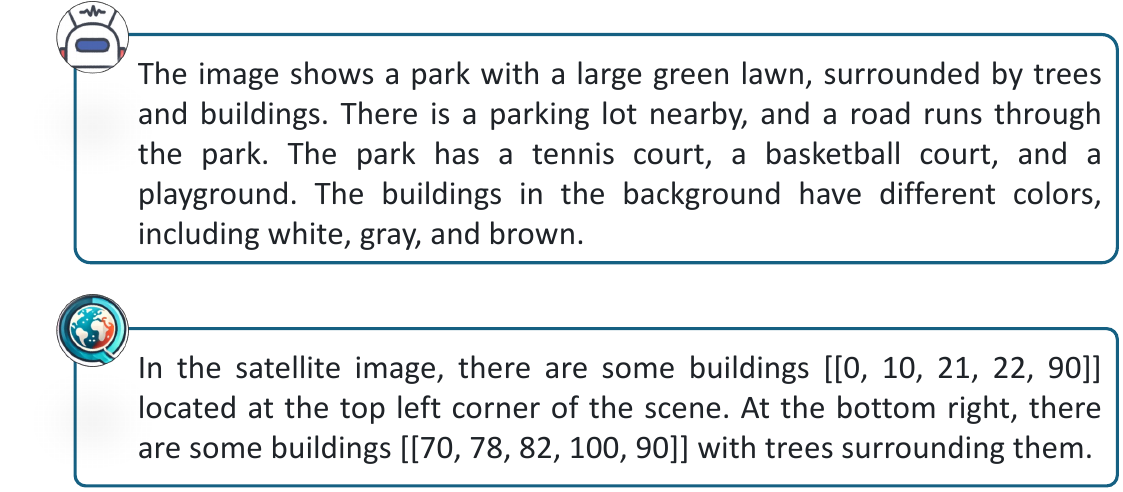}
    
    \caption{Image Captioning: Example response of different models.}
    \label{fig:captionss}
\end{figure*}

In Fig.~\ref{fig:spatial_relations}, for the first image, a few models performed well because the objects are close to each other, easy to identify, and have minimal visual complexity. In the remaining images, models struggled because the objects were farther apart, making it harder to identify their spatial relationships. The cluttered environments and larger spatial gaps made it difficult for the models to accurately understand the relationships between the objects. 
Fig.~\ref{fig:captionss} shows an aerial image alongside its ground truth caption and responses from various models. The ground truth provides a detailed and accurate description of the scene, while the model generated captions vary in capturing key elements such as urban \& natural features, pathways, and architectural structures. This comparison highlights differences in model responses for image captioning tasks.

\section{Multi-spectral}
\stepcounter{secnumber}
\label{sec:multispec}

\begin{figure}[b]
    \centering
    \includegraphics[width=0.9\linewidth]{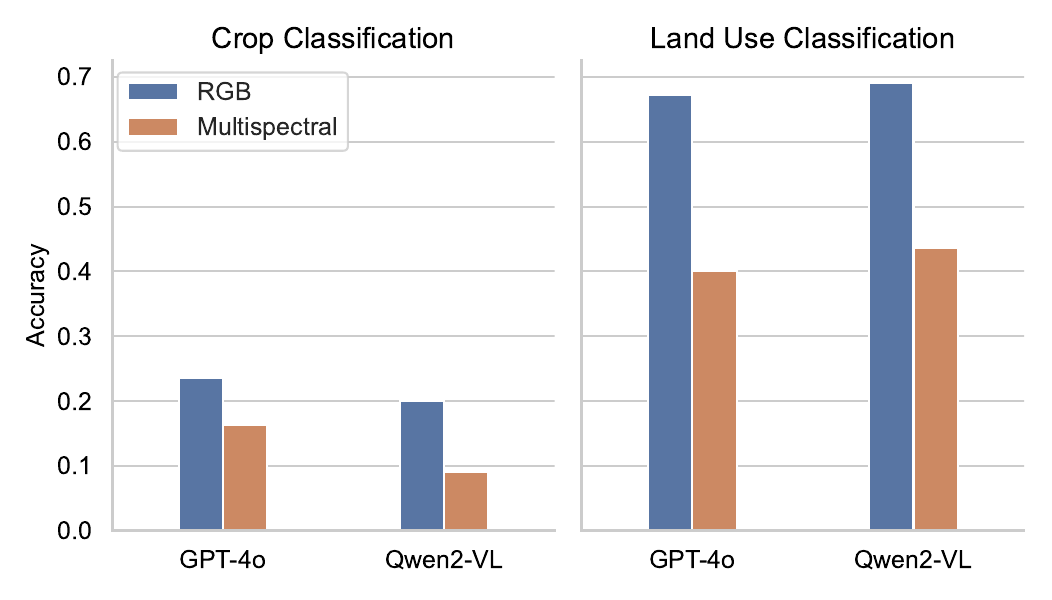}
    \captionsetup{justification=justified}
    \caption{It compares RGB and multispectral performance for Crop Type and Land Use Classification, showing a performance drop in multispectral accuracy.}
    \label{fig:rgb_vs_ms}
\end{figure}

In this section, we compare how GPT-4o and Qwen2-VL \cite{wang2024qwen2} perform on Crop Type Classification and Land Use Classification tasks using RGB and multispectral (MS) data (Fig.~\ref{fig:rgb_vs_ms}). The models perform much better with RGB inputs because they are designed and trained specifically for RGB images. The accuracy drops significantly for multispectral data, especially in crop-type classification. To use MS data with these models, Sentinel-2 bands were combined into three channels sequentially to mimic RGB inputs. For land use classification, which depends more on spatial patterns than detailed spectral information, the drop in performance is smaller. These results show the need for improved methods to adapt MS data for such tasks.

\begin{figure}[b]
    \centering
    \includegraphics[width=\linewidth]{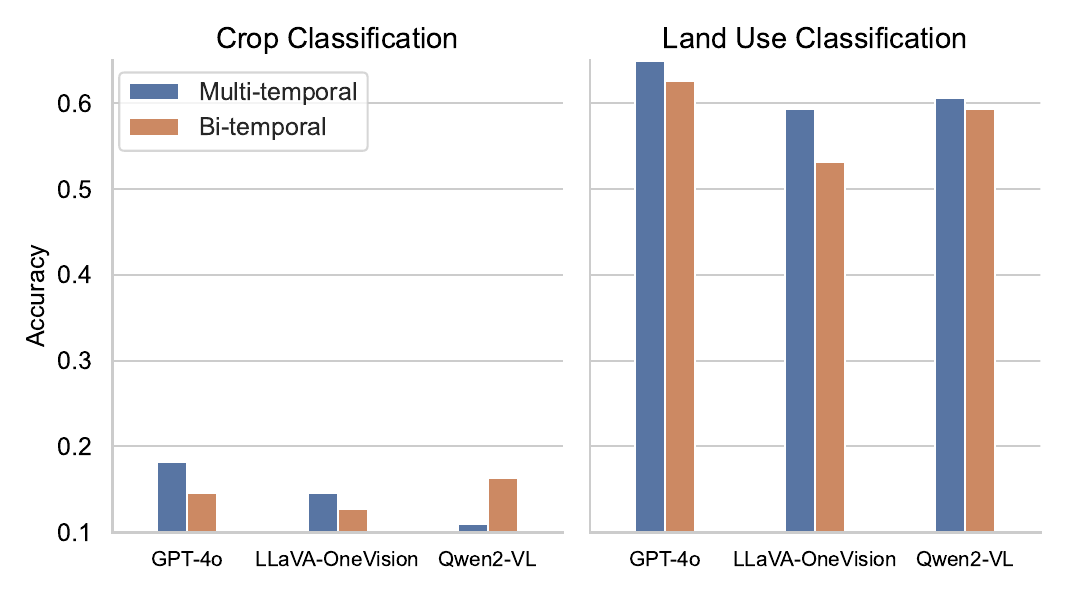}
    \captionsetup{justification=justified}
    \caption{Bi-temporal and Multi-temporal performance
    }
    \label{fig:bi_vs_mt}
\end{figure}
\section{Bi-temporal vs. Multi-temporal}
\stepcounter{secnumber}
\label{sec:bimulti}

We compare bi-temporal and multi-temporal image classification performance for Crop Type and Land Use classification tasks (Fig.~\ref{fig:bi_vs_mt}). Multi-temporal data outperforms bi-temporal data for land use classification, suggesting that more timestamps are sufficient to capture key temporal changes. For crop classification, multi-temporal inputs, reflect the improvement in GPT-4o and LLaVA-OneVision.

\section{Geographical Analysis}
\stepcounter{secnumber}
\label{sec:geoanalysis}

\begin{figure*}[t]
    \centering
    \includegraphics[width=\linewidth]{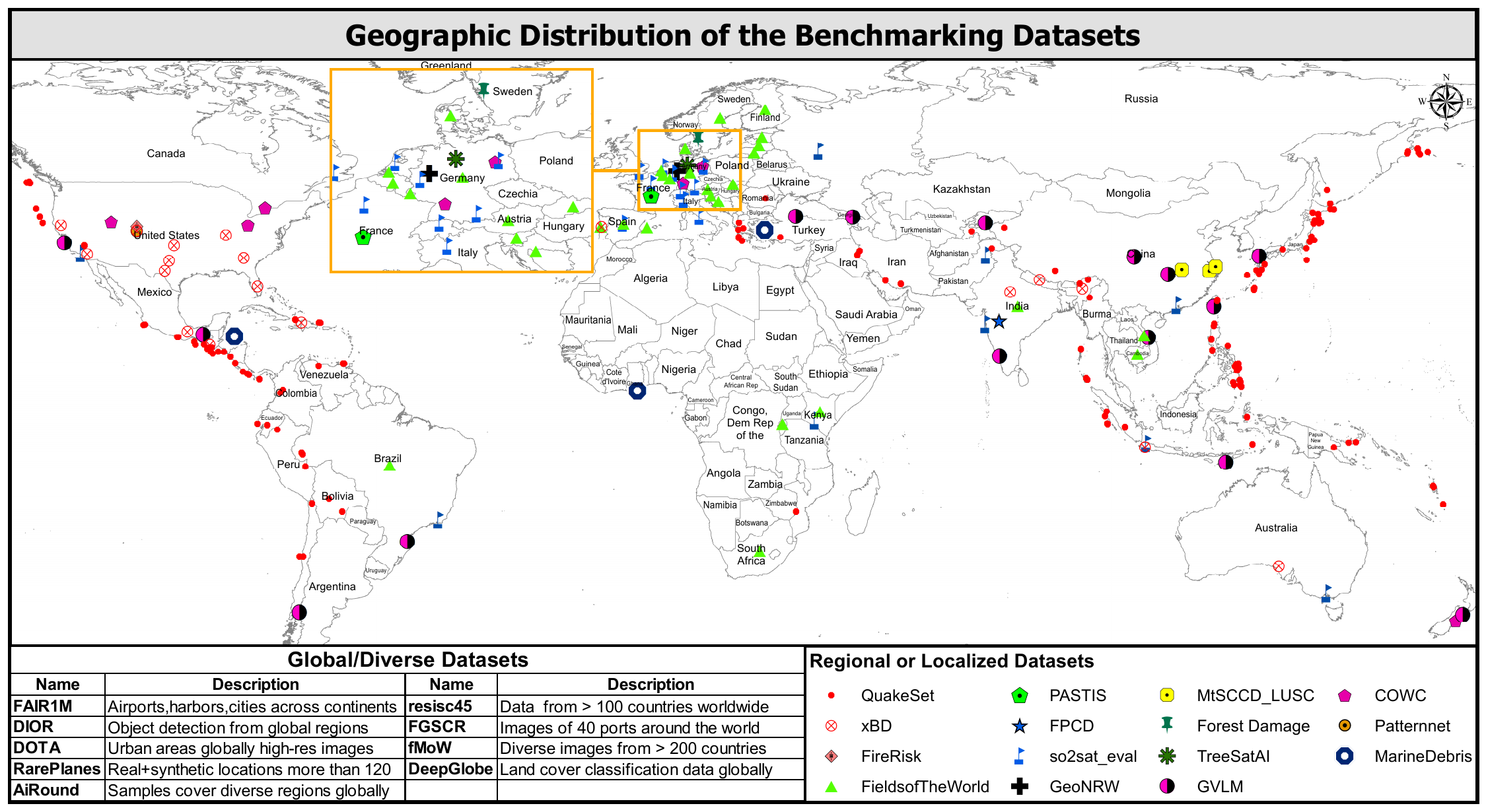}
    \captionsetup{justification=justified}
    \caption{Figure shows the geographic distribution of benchmarking datasets, highlighting global coverage and regional specialization.}
    \label{fig:geographyfig}
\end{figure*}

In this section, we detail the geographic distribution of benchmarking datasets used in the studied geospatial tasks. It categorizes datasets into global/diverse datasets and regional/localized datasets. Global datasets provide extensive coverage with samples from over 100 countries or diverse regions worldwide. On the other hand, regional and localized datasets, are tailored to specific tasks. The map in Fig.~\ref{fig:geographyfig} highlights that our benchmark is well represented across the globe. 

\begin{figure*}[t]
    \centering
    \begin{subfigure}[t]{0.49\linewidth}
        \centering
        \includegraphics[width=\linewidth]{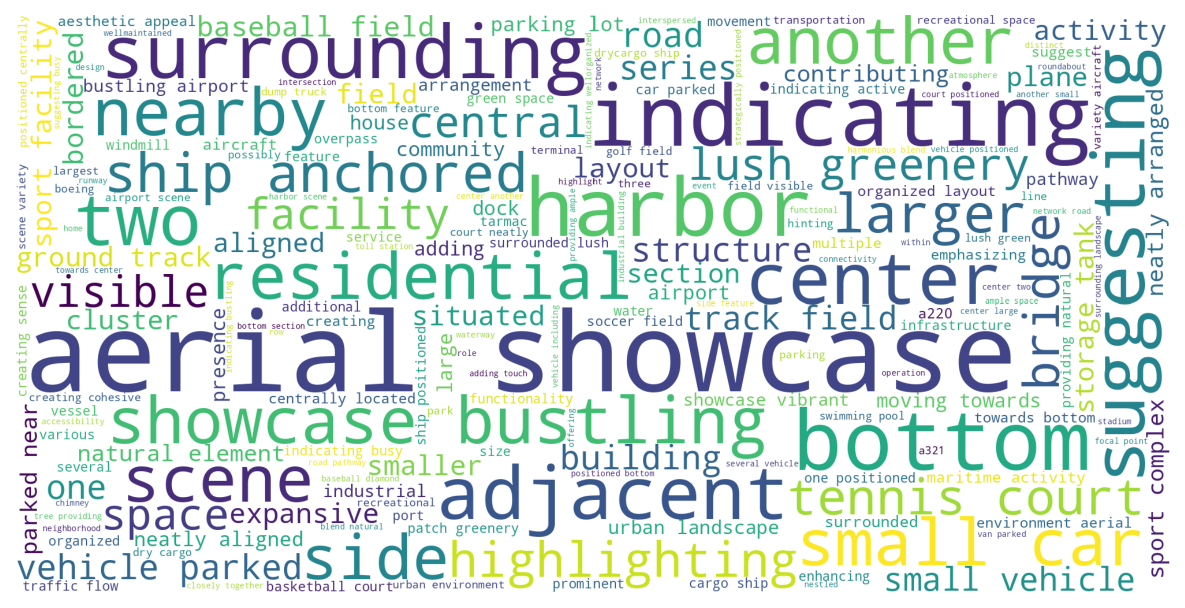}
        \caption{Word cloud from captions used in geospatial image description tasks.}
        \label{fig:combined_wordcloud1}
    \end{subfigure}
    \hfill
    \begin{subfigure}[t]{0.49\linewidth}
        \centering
        \includegraphics[width=\linewidth]{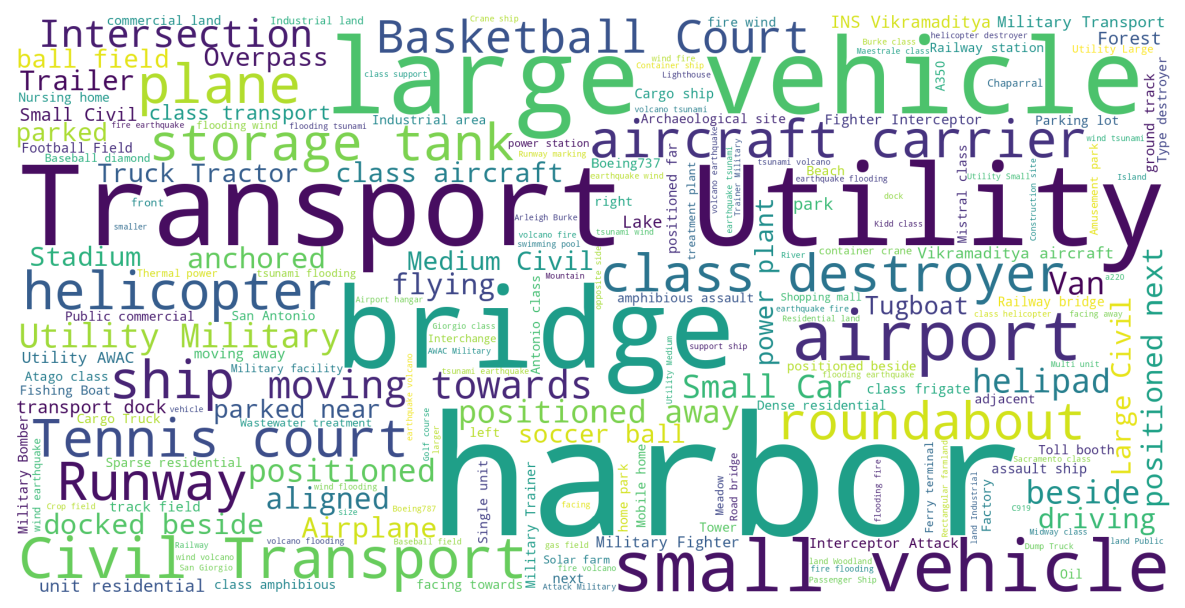}
        \caption{Word cloud from MCQs designed for geospatial task evaluation.}
        \label{fig:combined_wordcloud2}
    \end{subfigure}
    \caption{Word clouds showcasing terms used in evaluating VLMs on geospatial tasks, with the first focusing on image captions and the second on MCQs.}
    \label{fig:combined_wordcloud}
\end{figure*}

\section{Word Cloud}
\stepcounter{secnumber}
\label{sec:wordcloud}

The breakdown in Fig.~\ref{fig:combined_wordcloud1} leverages the word cloud as part of evaluating VLMs in geospatial tasks, with image captioning being one of the key areas of interest. 
The word cloud highlights terms commonly used in captions describing aerial or geospatial imagery. 
Words like ``aerial", ``surrounding" and ``residential" reflect spatial and contextual elements frequently addressed in such descriptions, while terms such as ``harbor", ``ship", ``tennis court" and ``greenery" represent specific features often observed in geospatial data. 
This provides a basis for understanding the capabilities and limitations of these models in capturing spatial relationships and identifying key features within geospatial tasks.

The word cloud in Fig.~\ref{fig:combined_wordcloud2} shows the terms used in MCQs. Keywords such as ``large vehicle", ``transport utility", ``harbor", ``bridge" and ``small vehicle" emphasize important categories and features frequently mentioned in the questions. Additional terms like ``aircraft carrier", ``runway", ``basketball court" and ``helicopter" represent a blend of transportation, infrastructure, and activity-based elements often linked to geospatial data. 
The use of varied and domain-specific vocabulary ensures the MCQs encompass a broad spectrum of scenarios for testing model capabilities.

%% file: main.bbl
\begin{thebibliography}{76}
\providecommand{\natexlab}[1]{#1}
\providecommand{\url}[1]{\texttt{#1}}
\expandafter\ifx\csname urlstyle\endcsname\relax
  \providecommand{\doi}[1]{doi: #1}\else
  \providecommand{\doi}{doi: \begingroup \urlstyle{rm}\Url}\fi

\bibitem[Achiam et~al.(2023)Achiam, Adler, Agarwal, Ahmad, Akkaya, Aleman, Almeida, Altenschmidt, Altman, Anadkat, et~al.]{achiam2023gpt}
Josh Achiam, Steven Adler, Sandhini Agarwal, Lama Ahmad, Ilge Akkaya, Florencia~Leoni Aleman, Diogo Almeida, Janko Altenschmidt, Sam Altman, Shyamal Anadkat, et~al.
\newblock Gpt-4 technical report.
\newblock \emph{arXiv preprint arXiv:2303.08774}, 2023.

\bibitem[Agency(2021)]{forest_damage}
Swedish~Forest Agency.
\newblock Forest damages – larch casebearer 1.0, 2021.
\newblock National Forest Data Lab. Dataset.

\bibitem[Bai et~al.(2023)Bai, Bai, Yang, Wang, Tan, Wang, Lin, Zhou, and Zhou]{bai2023qwen}
Jinze Bai, Shuai Bai, Shusheng Yang, Shijie Wang, Sinan Tan, Peng Wang, Junyang Lin, Chang Zhou, and Jingren Zhou.
\newblock Qwen-vl: A frontier large vision-language model with versatile abilities.
\newblock \emph{arXiv preprint arXiv:2308.12966}, 2023.

\bibitem[Baier et~al.(2022)Baier, Deschemps, Schmitt, and Yokoya]{GeoNRW}
Gerald Baier, Antonin Deschemps, Michael Schmitt, and Naoto Yokoya.
\newblock Synthesizing optical and sar imagery from land cover maps and auxiliary raster data.
\newblock \emph{IEEE Transactions on Geoscience and Remote Sensing}, 60:\penalty0 1--12, 2022.

\bibitem[Bazi et~al.(2024)Bazi, Bashmal, Al~Rahhal, Ricci, and Melgani]{bazi2024rs}
Yakoub Bazi, Laila Bashmal, Mohamad~Mahmoud Al~Rahhal, Riccardo Ricci, and Farid Melgani.
\newblock Rs-llava: A large vision-language model for joint captioning and question answering in remote sensing imagery.
\newblock \emph{Remote Sensing}, 16\penalty0 (9):\penalty0 1477, 2024.

\bibitem[Bubeck et~al.(2023)Bubeck, Chandrasekaran, Eldan, Gehrke, Horvitz, Kamar, Lee, Lee, Li, Lundberg, et~al.]{bubeck2023sparks}
S{\'e}bastien Bubeck, Varun Chandrasekaran, Ronen Eldan, Johannes Gehrke, Eric Horvitz, Ece Kamar, Peter Lee, Yin~Tat Lee, Yuanzhi Li, Scott Lundberg, et~al.
\newblock Sparks of artificial general intelligence: Early experiments with gpt-4. arxiv.
\newblock \emph{arXiv preprint arXiv:2303.12712}, 2023.

\bibitem[Chen et~al.(2024{\natexlab{a}})Chen, Xu, Kirmani, Ichter, Sadigh, Guibas, and Xia]{chen2024spatialvlm}
Boyuan Chen, Zhuo Xu, Sean Kirmani, Brain Ichter, Dorsa Sadigh, Leonidas Guibas, and Fei Xia.
\newblock Spatialvlm: Endowing vision-language models with spatial reasoning capabilities.
\newblock In \emph{Proceedings of the IEEE/CVF Conference on Computer Vision and Pattern Recognition}, pages 14455--14465, 2024{\natexlab{a}}.

\bibitem[Chen et~al.(2024{\natexlab{b}})Chen, Li, Dong, Zhang, Zang, Chen, Duan, Wang, Qiao, Lin, et~al.]{chen2024we}
Lin Chen, Jinsong Li, Xiaoyi Dong, Pan Zhang, Yuhang Zang, Zehui Chen, Haodong Duan, Jiaqi Wang, Yu Qiao, Dahua Lin, et~al.
\newblock Are we on the right way for evaluating large vision-language models?
\newblock \emph{arXiv preprint arXiv:2403.20330}, 2024{\natexlab{b}}.

\bibitem[Chen et~al.(2024{\natexlab{c}})Chen, Wu, Wang, Su, Chen, Xing, Zhong, Zhang, Zhu, Lu, et~al.]{chen2024internvl}
Zhe Chen, Jiannan Wu, Wenhai Wang, Weijie Su, Guo Chen, Sen Xing, Muyan Zhong, Qinglong Zhang, Xizhou Zhu, Lewei Lu, et~al.
\newblock Internvl: Scaling up vision foundation models and aligning for generic visual-linguistic tasks.
\newblock In \emph{Proceedings of the IEEE/CVF Conference on Computer Vision and Pattern Recognition}, pages 24185--24198, 2024{\natexlab{c}}.

\bibitem[Cheng et~al.(2017)Cheng, Han, and Lu]{resis}
Gong Cheng, Junwei Han, and Xiaoqiang Lu.
\newblock Remote sensing image scene classification: Benchmark and state of the art.
\newblock \emph{Proceedings of the IEEE}, 105\penalty0 (10):\penalty0 1865--1883, 2017.

\bibitem[Cheng et~al.(2021)Cheng, Wang, Li, Xie, Lang, Yao, and Han]{dior}
Gong Cheng, Jiabao Wang, Ke Li, Xingxing Xie, Chunbo Lang, Yanqing Yao, and Junwei Han.
\newblock Anchor-free oriented proposal generator for object detection.
\newblock \emph{CoRR}, abs/2110.01931, 2021.

\bibitem[CSE499DeforestationSatellite()]{deforestation-satellite-imagery-335n4_dataset}
CSE499DeforestationSatellite.
\newblock Deforestation-satellite-imagery dataset.
\newblock \url{ https://universe.roboflow.com/cse499deforestationsatellite/deforestation-satellite-imagery-335n4 }.

\bibitem[Demir et~al.(2018)Demir, Koperski, Lindenbaum, Pang, Huang, Basu, Hughes, Tuia, and Raskar]{DeepGlobe}
Ilke Demir, Krzysztof Koperski, David Lindenbaum, Guan Pang, Jing Huang, Saikat Basu, Forest Hughes, Devis Tuia, and Ramesh Raskar.
\newblock Deepglobe 2018: {A} challenge to parse the earth through satellite images.
\newblock \emph{CoRR}, abs/1805.06561, 2018.

\bibitem[Di et~al.(2021)Di, Jiang, and Zhang]{fgscr}
Yanghua Di, Zhiguo Jiang, and Haopeng Zhang.
\newblock A public dataset for fine-grained ship classification in optical remote sensing images.
\newblock \emph{Remote Sensing}, 13\penalty0 (4), 2021.

\bibitem[Gupta et~al.(2019)Gupta, Hosfelt, Sajeev, Patel, Goodman, Doshi, Heim, Choset, and Gaston]{xbd}
Ritwik Gupta, Richard Hosfelt, Sandra Sajeev, Nirav Patel, Bryce Goodman, Jigar Doshi, Eric~T. Heim, Howie Choset, and Matthew~E. Gaston.
\newblock xbd: {A} dataset for assessing building damage from satellite imagery.
\newblock \emph{CoRR}, abs/1911.09296, 2019.

\bibitem[Hu et~al.(2023)Hu, Yuan, Wen, Lu, and Li]{hu2023rsgpt}
Yuan Hu, Jianlong Yuan, Congcong Wen, Xiaonan Lu, and Xiang Li.
\newblock Rsgpt: A remote sensing vision language model and benchmark.
\newblock \emph{arXiv preprint arXiv:2307.15266}, 2023.

\bibitem[Jean et~al.(2016)Jean, Burke, Xie, Davis, Lobell, and Ermon]{jean2016combining}
Neal Jean, Marshall Burke, Michael Xie, W~Matthew Davis, David~B Lobell, and Stefano Ermon.
\newblock Combining satellite imagery and machine learning to predict poverty.
\newblock \emph{Science}, 353\penalty0 (6301):\penalty0 790--794, 2016.

\bibitem[Kerner et~al.(2024)Kerner, Chaudhari, Ghosh, Robinson, Ahmad, Choi, Jacobs, Holmes, Mohr, Dodhia, Ferres, and Marcus]{fmow}
Hannah Kerner, Snehal Chaudhari, Aninda Ghosh, Caleb Robinson, Adeel Ahmad, Eddie Choi, Nathan Jacobs, Chris Holmes, Matthias Mohr, Rahul Dodhia, Juan M.~Lavista Ferres, and Jennifer Marcus.
\newblock Fields of the world: A machine learning benchmark dataset for global agricultural field boundary segmentation, 2024.

\bibitem[Kuckreja et~al.(2024)Kuckreja, Danish, Naseer, Das, Khan, and Khan]{kuckreja2024geochat}
Kartik Kuckreja, Muhammad~Sohail Danish, Muzammal Naseer, Abhijit Das, Salman Khan, and Fahad~Shahbaz Khan.
\newblock Geochat: Grounded large vision-language model for remote sensing.
\newblock In \emph{Proceedings of the IEEE/CVF Conference on Computer Vision and Pattern Recognition}, pages 27831--27840, 2024.

\bibitem[Li et~al.(2023)Li, Wang, Wang, Ge, Ge, and Shan]{li2023seed}
Bohao Li, Rui Wang, Guangzhi Wang, Yuying Ge, Yixiao Ge, and Ying Shan.
\newblock Seed-bench: Benchmarking multimodal llms with generative comprehension.
\newblock \emph{arXiv preprint arXiv:2307.16125}, 2023.

\bibitem[Li et~al.(2024{\natexlab{a}})Li, Ge, Chen, Ge, Zhang, and Shan]{li2024seed2}
Bohao Li, Yuying Ge, Yi Chen, Yixiao Ge, Ruimao Zhang, and Ying Shan.
\newblock Seed-bench-2-plus: Benchmarking multimodal large language models with text-rich visual comprehension.
\newblock \emph{arXiv preprint arXiv:2404.16790}, 2024{\natexlab{a}}.

\bibitem[Li et~al.(2024{\natexlab{b}})Li, Ge, Ge, Wang, Wang, Zhang, and Shan]{li2024seed}
Bohao Li, Yuying Ge, Yixiao Ge, Guangzhi Wang, Rui Wang, Ruimao Zhang, and Ying Shan.
\newblock Seed-bench: Benchmarking multimodal large language models.
\newblock In \emph{Proceedings of the IEEE/CVF Conference on Computer Vision and Pattern Recognition}, pages 13299--13308, 2024{\natexlab{b}}.

\bibitem[Li et~al.(2024{\natexlab{c}})Li, Zhang, Guo, Zhang, Li, Zhang, Zhang, Li, Liu, and Li]{li2024llavaone}
Bo Li, Yuanhan Zhang, Dong Guo, Renrui Zhang, Feng Li, Hao Zhang, Kaichen Zhang, Yanwei Li, Ziwei Liu, and Chunyuan Li.
\newblock Llava-onevision: Easy visual task transfer.
\newblock \emph{arXiv preprint arXiv:2408.03326}, 2024{\natexlab{c}}.

\bibitem[Li et~al.(2020)Li, Wan, Cheng, Meng, and Han]{li2020object}
Ke Li, Gang Wan, Gong Cheng, Liqiu Meng, and Junwei Han.
\newblock Object detection in optical remote sensing images: A survey and a new benchmark.
\newblock \emph{ISPRS journal of photogrammetry and remote sensing}, 159:\penalty0 296--307, 2020.

\bibitem[Li et~al.(2024{\natexlab{d}})Li, Cao, and Meng]{li2024new}
Kaiyu Li, Xiangyong Cao, and Deyu Meng.
\newblock A new learning paradigm for foundation model-based remote-sensing change detection.
\newblock \emph{IEEE Transactions on Geoscience and Remote Sensing}, 62:\penalty0 1--12, 2024{\natexlab{d}}.

\bibitem[Li et~al.(2024{\natexlab{e}})Li, Ding, and Elhoseiny]{li2024vrsbench}
Xiang Li, Jian Ding, and Mohamed Elhoseiny.
\newblock Vrsbench: A versatile vision-language benchmark dataset for remote sensing image understanding.
\newblock \emph{arXiv preprint arXiv:2406.12384}, 2024{\natexlab{e}}.

\bibitem[Lin et~al.(2023)Lin, Liu, Zhang, Gao, Qiu, Xiao, Qiu, Lin, Shao, Chen, et~al.]{lin2023sphinx}
Ziyi Lin, Chris Liu, Renrui Zhang, Peng Gao, Longtian Qiu, Han Xiao, Han Qiu, Chen Lin, Wenqi Shao, Keqin Chen, et~al.
\newblock Sphinx: The joint mixing of weights, tasks, and visual embeddings for multi-modal large language models.
\newblock \emph{arXiv preprint arXiv:2311.07575}, 2023.

\bibitem[Liu et~al.(2024{\natexlab{a}})Liu, Chen, Guan, Zhou, Zhu, Ye, Fu, and Zhou]{liu2024remoteclip}
Fan Liu, Delong Chen, Zhangqingyun Guan, Xiaocong Zhou, Jiale Zhu, Qiaolin Ye, Liyong Fu, and Jun Zhou.
\newblock Remoteclip: A vision language foundation model for remote sensing.
\newblock \emph{IEEE Transactions on Geoscience and Remote Sensing}, 2024{\natexlab{a}}.

\bibitem[Liu et~al.(2024{\natexlab{b}})Liu, Li, Li, Li, Zhang, Shen, and Lee]{liu2024llavanext}
Haotian Liu, Chunyuan Li, Yuheng Li, Bo Li, Yuanhan Zhang, Sheng Shen, and Yong~Jae Lee.
\newblock Llava-next: Improved reasoning, ocr, and world knowledge, 2024{\natexlab{b}}.

\bibitem[Liu et~al.(2024{\natexlab{c}})Liu, Li, Wu, and Lee]{li2024llava}
Haotian Liu, Chunyuan Li, Qingyang Wu, and Yong~Jae Lee.
\newblock Visual instruction tuning.
\newblock \emph{Advances in neural information processing systems}, 36, 2024{\natexlab{c}}.

\bibitem[Liu et~al.(2024{\natexlab{d}})Liu, Zhou, Guan, and Zhao]{MtSCCD}
Jinglei Liu, Weixun Zhou, Haiyan Guan, and Wenzhi Zhao.
\newblock Similarity learning for land use scene-level change detection.
\newblock \emph{IEEE Journal of Selected Topics in Applied Earth Observations and Remote Sensing}, 17:\penalty0 6501--6513, 2024{\natexlab{d}}.

\bibitem[Liu et~al.(2024{\natexlab{e}})Liu, Ma, Zhang, Wang, Ji, Sun, and Ji]{liu2024rotated}
Sihan Liu, Yiwei Ma, Xiaoqing Zhang, Haowei Wang, Jiayi Ji, Xiaoshuai Sun, and Rongrong Ji.
\newblock Rotated multi-scale interaction network for referring remote sensing image segmentation.
\newblock In \emph{Proceedings of the IEEE/CVF Conference on Computer Vision and Pattern Recognition}, pages 26658--26668, 2024{\natexlab{e}}.

\bibitem[Liu et~al.(2025)Liu, Duan, Zhang, Li, Zhang, Zhao, Yuan, Wang, He, Liu, et~al.]{liu2025mmbench}
Yuan Liu, Haodong Duan, Yuanhan Zhang, Bo Li, Songyang Zhang, Wangbo Zhao, Yike Yuan, Jiaqi Wang, Conghui He, Ziwei Liu, et~al.
\newblock Mmbench: Is your multi-modal model an all-around player?
\newblock In \emph{European Conference on Computer Vision}, pages 216--233. Springer, 2025.

\bibitem[Luo et~al.(2024)Luo, Pang, Zhang, Wang, Wang, Dang, Lao, Wang, Chen, Tan, et~al.]{luo2024skysensegpt}
Junwei Luo, Zhen Pang, Yongjun Zhang, Tingzhu Wang, Linlin Wang, Bo Dang, Jiangwei Lao, Jian Wang, Jingdong Chen, Yihua Tan, et~al.
\newblock Skysensegpt: A fine-grained instruction tuning dataset and model for remote sensing vision-language understanding.
\newblock \emph{arXiv preprint arXiv:2406.10100}, 2024.

\bibitem[Machado et~al.(2020)Machado, Ferreira, Nogueira, Oliveira, Gama, and dos Santos]{airound}
Gabriel L.~S. Machado, Edemir Ferreira, Keiller Nogueira, Hugo~N. Oliveira, Pedro H.~T. Gama, and Jefersson~A. dos Santos.
\newblock Airound and cv-brct: Novel multi-view datasets for scene classification.
\newblock \emph{CoRR}, abs/2008.01133, 2020.

\bibitem[Muhtar et~al.(2024)Muhtar, Li, Gu, Zhang, and Xiao]{muhtar2024lhrs}
Dilxat Muhtar, Zhenshi Li, Feng Gu, Xueliang Zhang, and Pengfeng Xiao.
\newblock Lhrs-bot: Empowering remote sensing with vgi-enhanced large multimodal language model.
\newblock In \emph{European Conference on Computer Vision}, pages 440--457. Springer, 2024.

\bibitem[Mundhenk et~al.(2016)Mundhenk, Konjevod, Sakla, and Boakye]{cowc}
T.~Nathan Mundhenk, Goran Konjevod, Wesam~A. Sakla, and Kofi Boakye.
\newblock A large contextual dataset for classification, detection and counting of cars with deep learning.
\newblock In \emph{Computer Vision -- ECCV 2016}, pages 785--800, Cham, 2016. Springer International Publishing.

\bibitem[Nayak et~al.(2024)Nayak, Jain, Awal, Reddy, van Steenkiste, Hendricks, Sta{\'n}czak, and Agrawal]{nayak2024benchmarking}
Shravan Nayak, Kanishk Jain, Rabiul Awal, Siva Reddy, Sjoerd van Steenkiste, Lisa~Anne Hendricks, Karolina Sta{\'n}czak, and Aishwarya Agrawal.
\newblock Benchmarking vision language models for cultural understanding.
\newblock \emph{arXiv preprint arXiv:2407.10920}, 2024.

\bibitem[OpenAI(2024)]{openai2024gpt4o}
OpenAI.
\newblock Hello gpt-4o, 2024.

\bibitem[Rasheed et~al.(2024)Rasheed, Maaz, Shaji, Shaker, Khan, Cholakkal, Anwer, Xing, Yang, and Khan]{rasheed2024glamm}
Hanoona Rasheed, Muhammad Maaz, Sahal Shaji, Abdelrahman Shaker, Salman Khan, Hisham Cholakkal, Rao~M Anwer, Eric Xing, Ming-Hsuan Yang, and Fahad~S Khan.
\newblock Glamm: Pixel grounding large multimodal model.
\newblock In \emph{Proceedings of the IEEE/CVF Conference on Computer Vision and Pattern Recognition}, pages 13009--13018, 2024.

\bibitem[Rege~Cambrin and Garza(2024)]{QuakeSet}
Daniele Rege~Cambrin and Paolo Garza.
\newblock Quakeset: A dataset and low-resource models to monitor earthquakes through sentinel-1.
\newblock \emph{Proceedings of the International ISCRAM Conference}, 2024.

\bibitem[Ru{\ss}wurm et~al.(2020)Ru{\ss}wurm, Wang, Korner, and Lobell]{russwurm2020meta}
Marc Ru{\ss}wurm, Sherrie Wang, Marco Korner, and David Lobell.
\newblock Meta-learning for few-shot land cover classification.
\newblock In \emph{Proceedings of the ieee/cvf conference on computer vision and pattern recognition workshops}, pages 200--201, 2020.

\bibitem[Sainte Fare~Garnot and Landrieu(2021)]{pastis}
Vivien Sainte Fare~Garnot and Loic Landrieu.
\newblock Panoptic segmentation of satellite image time series with convolutional temporal attention networks.
\newblock \emph{ICCV}, 2021.

\bibitem[Shah et~al.(2021)Shah, Thomas, and Maskey]{nasa_marine_debris}
A. Shah, L. Thomas, and M. Maskey.
\newblock Marine debris dataset for object detection in planetscope imagery, 2021.

\bibitem[Shen et~al.(2023)Shen, Seneviratne, Wanyan, and Kirley]{FireRisk}
Shuchang Shen, Sachith Seneviratne, Xinye Wanyan, and Michael Kirley.
\newblock Firerisk: A remote sensing dataset for fire risk assessment with benchmarks using supervised and self-supervised learning, 2023.

\bibitem[Shermeyer et~al.(2020)Shermeyer, Hossler, Van~Etten, Hogan, Lewis, and Kim]{RarePlanes_Dataset}
Jacob Shermeyer, Thomas Hossler, Adam Van~Etten, Daniel Hogan, Ryan Lewis, and Daeil Kim.
\newblock Rareplanes dataset, 2020.

\bibitem[Soni et~al.(2024)Soni, Dudhane, Debary, Fiaz, Munir, Danish, Fraccaro, Watson, Klein, Khan, et~al.]{soni2024earthdial}
Sagar Soni, Akshay Dudhane, Hiyam Debary, Mustansar Fiaz, Muhammad~Akhtar Munir, Muhammad~Sohail Danish, Paolo Fraccaro, Campbell~D Watson, Levente~J Klein, Fahad~Shahbaz Khan, et~al.
\newblock Earthdial: Turning multi-sensory earth observations to interactive dialogues.
\newblock \emph{arXiv preprint arXiv:2412.15190}, 2024.

\bibitem[Sun et~al.(2021)Sun, Wang, Yan, Xu, Wang, Diao, Chen, Li, Feng, Xu, Weinmann, Hinz, Wang, and Fu]{fair1m}
Xian Sun, Peijin Wang, Zhiyuan Yan, F. Xu, Ruiping Wang, W. Diao, Jin Chen, Jihao Li, Yingchao Feng, Tao Xu, M. Weinmann, S. Hinz, Cheng Wang, and K. Fu.
\newblock Fair1m: A benchmark dataset for fine-grained object recognition in high-resolution remote sensing imagery.
\newblock \emph{Isprs Journal of Photogrammetry and Remote Sensing}, 2021.

\bibitem[Touvron et~al.(2023{\natexlab{a}})Touvron, Lavril, Izacard, Martinet, Lachaux, Lacroix, Rozi{\`e}re, Goyal, Hambro, Azhar, et~al.]{touvron2023llama1}
Hugo Touvron, Thibaut Lavril, Gautier Izacard, Xavier Martinet, Marie-Anne Lachaux, Timoth{\'e}e Lacroix, Baptiste Rozi{\`e}re, Naman Goyal, Eric Hambro, Faisal Azhar, et~al.
\newblock Llama: Open and efficient foundation language models.
\newblock \emph{arXiv preprint arXiv:2302.13971}, 2023{\natexlab{a}}.

\bibitem[Touvron et~al.(2023{\natexlab{b}})Touvron, Martin, Stone, Albert, Almahairi, Babaei, Bashlykov, Batra, Bhargava, Bhosale, et~al.]{touvron2023llama}
Hugo Touvron, Louis Martin, Kevin Stone, Peter Albert, Amjad Almahairi, Yasmine Babaei, Nikolay Bashlykov, Soumya Batra, Prajjwal Bhargava, Shruti Bhosale, et~al.
\newblock Llama 2: Open foundation and fine-tuned chat models.
\newblock \emph{arXiv preprint arXiv:2307.09288}, 2023{\natexlab{b}}.

\bibitem[Tundia et~al.(2023)Tundia, Kumar, Damani, and Sivakumar]{FPCD}
Chintan Tundia, Rajiv Kumar, Om Damani, and G. Sivakumar.
\newblock Fpcd: An open aerial vhr dataset for farm pond change detection.
\newblock In \emph{Proceedings of the 18th International Joint Conference on Computer Vision, Imaging and Computer Graphics Theory and Applications}, page 862–869. SCITEPRESS - Science and Technology Publications, 2023.

\bibitem[Wang et~al.(2024{\natexlab{a}})Wang, Xu, Xie, Wang, Li, Xie, Zhang, Xiong, and Chen]{wang2024m4u}
Hongyu Wang, Jiayu Xu, Senwei Xie, Ruiping Wang, Jialin Li, Zhaojie Xie, Bin Zhang, Chuyan Xiong, and Xilin Chen.
\newblock M4u: Evaluating multilingual understanding and reasoning for large multimodal models.
\newblock \emph{arXiv preprint arXiv:2405.15638}, 2024{\natexlab{a}}.

\bibitem[Wang et~al.(2024{\natexlab{b}})Wang, Zheng, Chen, Ma, and Zhong]{wang2024earthvqa}
Junjue Wang, Zhuo Zheng, Zihang Chen, Ailong Ma, and Yanfei Zhong.
\newblock Earthvqa: Towards queryable earth via relational reasoning-based remote sensing visual question answering.
\newblock In \emph{Proceedings of the AAAI Conference on Artificial Intelligence}, pages 5481--5489, 2024{\natexlab{b}}.

\bibitem[Wang et~al.(2024{\natexlab{c}})Wang, Bai, Tan, Wang, Fan, Bai, Chen, Liu, Wang, Ge, et~al.]{wang2024qwen2}
Peng Wang, Shuai Bai, Sinan Tan, Shijie Wang, Zhihao Fan, Jinze Bai, Keqin Chen, Xuejing Liu, Jialin Wang, Wenbin Ge, et~al.
\newblock Qwen2-vl: Enhancing vision-language model's perception of the world at any resolution.
\newblock \emph{arXiv preprint arXiv:2409.12191}, 2024{\natexlab{c}}.

\bibitem[Wenqi et~al.(2024)Wenqi, Gong, Meijun, Yanqing, Xingxing, Xiwen, and Junwei]{wenqi2024mar20}
YU Wenqi, CHENG Gong, WANG Meijun, YAO Yanqing, XIE Xingxing, YAO Xiwen, and HAN Junwei.
\newblock Mar20: A benchmark for military aircraft recognition in remote sensing images.
\newblock \emph{National Remote Sensing Bulletin}, 27\penalty0 (12):\penalty0 2688--2696, 2024.

\bibitem[Xia et~al.(2018)Xia, Bai, Ding, Zhu, Belongie, Luo, Datcu, Pelillo, and Zhang]{xia2018dota}
Gui-Song Xia, Xiang Bai, Jian Ding, Zhen Zhu, Serge Belongie, Jiebo Luo, Mihai Datcu, Marcello Pelillo, and Liangpei Zhang.
\newblock Dota: A large-scale dataset for object detection in aerial images.
\newblock In \emph{Proceedings of the IEEE conference on computer vision and pattern recognition}, pages 3974--3983, 2018.

\bibitem[Xiao et~al.(2024)Xiao, Wu, Xu, Dai, Hu, Lu, Zeng, Liu, and Yuan]{xiao2024florence}
Bin Xiao, Haiping Wu, Weijian Xu, Xiyang Dai, Houdong Hu, Yumao Lu, Michael Zeng, Ce Liu, and Lu Yuan.
\newblock Florence-2: Advancing a unified representation for a variety of vision tasks.
\newblock In \emph{Proceedings of the IEEE/CVF Conference on Computer Vision and Pattern Recognition}, pages 4818--4829, 2024.

\bibitem[Xie et~al.(2021)Xie, Wang, Yu, Anandkumar, Alvarez, and Luo]{xie2021segformer}
Enze Xie, Wenhai Wang, Zhiding Yu, Anima Anandkumar, Jose~M Alvarez, and Ping Luo.
\newblock Segformer: Simple and efficient design for semantic segmentation with transformers.
\newblock \emph{Advances in neural information processing systems}, 34:\penalty0 12077--12090, 2021.

\bibitem[Yang et~al.(2024)Yang, Wu, Yang, Chen, and Geng]{yang2024exploring}
Xu Yang, Yongliang Wu, Mingzhuo Yang, Haokun Chen, and Xin Geng.
\newblock Exploring diverse in-context configurations for image captioning.
\newblock \emph{Advances in Neural Information Processing Systems}, 36, 2024.

\bibitem[Yin et~al.(2021)Yin, Li, Hu, Peng, and Chang]{yin2021broaden}
Da Yin, Liunian~Harold Li, Ziniu Hu, Nanyun Peng, and Kai-Wei Chang.
\newblock Broaden the vision: Geo-diverse visual commonsense reasoning.
\newblock \emph{arXiv preprint arXiv:2109.06860}, 2021.

\bibitem[Yin et~al.(2024)Yin, Wang, Cao, Shi, Liu, Li, Huang, Wang, Sheng, Bai, et~al.]{yin2024lamm}
Zhenfei Yin, Jiong Wang, Jianjian Cao, Zhelun Shi, Dingning Liu, Mukai Li, Xiaoshui Huang, Zhiyong Wang, Lu Sheng, Lei Bai, et~al.
\newblock Lamm: Language-assisted multi-modal instruction-tuning dataset, framework, and benchmark.
\newblock \emph{Advances in Neural Information Processing Systems}, 36, 2024.

\bibitem[You et~al.(2023)You, Zhang, Gan, Du, Zhang, Wang, Cao, Chang, and Yang]{you2023ferret}
Haoxuan You, Haotian Zhang, Zhe Gan, Xianzhi Du, Bowen Zhang, Zirui Wang, Liangliang Cao, Shih-Fu Chang, and Yinfei Yang.
\newblock Ferret: Refer and ground anything anywhere at any granularity.
\newblock \emph{arXiv preprint arXiv:2310.07704}, 2023.

\bibitem[Yu et~al.(2023)Yu, Yang, Li, Wang, Lin, Liu, Wang, and Wang]{yu2023mm}
Weihao Yu, Zhengyuan Yang, Linjie Li, Jianfeng Wang, Kevin Lin, Zicheng Liu, Xinchao Wang, and Lijuan Wang.
\newblock Mm-vet: Evaluating large multimodal models for integrated capabilities.
\newblock \emph{arXiv preprint arXiv:2308.02490}, 2023.

\bibitem[Yuan et~al.(2024)Yuan, Mou, Hua, and Zhu]{yuan2024rrsis}
Zhenghang Yuan, Lichao Mou, Yuansheng Hua, and Xiao~Xiang Zhu.
\newblock Rrsis: Referring remote sensing image segmentation.
\newblock \emph{IEEE Transactions on Geoscience and Remote Sensing}, 2024.

\bibitem[Yue et~al.(2024)Yue, Ni, Zhang, Zheng, Liu, Zhang, Stevens, Jiang, Ren, Sun, et~al.]{yue2024mmmu}
Xiang Yue, Yuansheng Ni, Kai Zhang, Tianyu Zheng, Ruoqi Liu, Ge Zhang, Samuel Stevens, Dongfu Jiang, Weiming Ren, Yuxuan Sun, et~al.
\newblock Mmmu: A massive multi-discipline multimodal understanding and reasoning benchmark for expert agi.
\newblock In \emph{Proceedings of the IEEE/CVF Conference on Computer Vision and Pattern Recognition}, pages 9556--9567, 2024.

\bibitem[Zhan et~al.(2024)Zhan, Xiong, and Yuan]{zhan2024skyeyegpt}
Yang Zhan, Zhitong Xiong, and Yuan Yuan.
\newblock Skyeyegpt: Unifying remote sensing vision-language tasks via instruction tuning with large language model.
\newblock \emph{arXiv preprint arXiv:2401.09712}, 2024.

\bibitem[Zhang and Wang(2024)]{zhang2024good}
Chenhui Zhang and Sherrie Wang.
\newblock Good at captioning, bad at counting: Benchmarking gpt-4v on earth observation data.
\newblock \emph{arXiv preprint arXiv:2401.17600}, 2024.

\bibitem[Zhang et~al.(2023{\natexlab{a}})Zhang, Li, Liu, Zhang, Su, Zhu, Ni, and Shum]{zhangdino}
Hao Zhang, Feng Li, Shilong Liu, Lei Zhang, Hang Su, Jun Zhu, Lionel Ni, and Heung-Yeung Shum.
\newblock Dino: Detr with improved denoising anchor boxes for end-to-end object detection.
\newblock In \emph{The Eleventh International Conference on Learning Representations}, 2023{\natexlab{a}}.

\bibitem[Zhang* et~al.(2020)Zhang*, Kishore*, Wu*, Weinberger, and Artzi]{bert-score}
Tianyi Zhang*, Varsha Kishore*, Felix Wu*, Kilian~Q. Weinberger, and Yoav Artzi.
\newblock Bertscore: Evaluating text generation with bert.
\newblock In \emph{International Conference on Learning Representations}, 2020.

\bibitem[Zhang et~al.(2024{\natexlab{a}})Zhang, Cai, Zhang, Zhuang, and Mao]{zhang2024earthgpt}
Wei Zhang, Miaoxin Cai, Tong Zhang, Yin Zhuang, and Xuerui Mao.
\newblock Earthgpt: A universal multi-modal large language model for multi-sensor image comprehension in remote sensing domain.
\newblock \emph{IEEE Transactions on Geoscience and Remote Sensing}, 2024{\natexlab{a}}.

\bibitem[Zhang et~al.(2023{\natexlab{b}})Zhang, Yu, Pun, and Shi]{gvlm_cd}
Xiaokang Zhang, Weikang Yu, Man-On Pun, and Wenzhong Shi.
\newblock Cross-domain landslide mapping from large-scale remote sensing images using prototype-guided domain-aware progressive representation learning.
\newblock \emph{ISPRS Journal of Photogrammetry and Remote Sensing}, 197:\penalty0 1--17, 2023{\natexlab{b}}.

\bibitem[Zhang et~al.(2024{\natexlab{b}})Zhang, Zhang, Tian, Fu, Zhang, Wu, Li, Wang, Wen, Zhang, et~al.]{zhang2024mme}
Yi-Fan Zhang, Huanyu Zhang, Haochen Tian, Chaoyou Fu, Shuangqing Zhang, Junfei Wu, Feng Li, Kun Wang, Qingsong Wen, Zhang Zhang, et~al.
\newblock Mme-realworld: Could your multimodal llm challenge high-resolution real-world scenarios that are difficult for humans?
\newblock \emph{arXiv preprint arXiv:2408.13257}, 2024{\natexlab{b}}.

\bibitem[Zhao et~al.(2024)Zhao, Yuan, Chen, Li, Liu, Li, and Gao]{zhao2024panoptic}
Danpei Zhao, Bo Yuan, Ziqiang Chen, Tian Li, Zhuoran Liu, Wentao Li, and Yue Gao.
\newblock Panoptic perception: A novel task and fine-grained dataset for universal remote sensing image interpretation.
\newblock \emph{IEEE Transactions on Geoscience and Remote Sensing}, 2024.

\bibitem[Zhou et~al.(2017)Zhou, Newsam, Li, and Shao]{PatternNet}
Weixun Zhou, Shawn~D. Newsam, Congmin Li, and Zhenfeng Shao.
\newblock Patternnet: {A} benchmark dataset for performance evaluation of remote sensing image retrieval.
\newblock \emph{CoRR}, abs/1706.03424, 2017.

\bibitem[Zhu et~al.(2020{\natexlab{a}})Zhu, Su, Lu, Li, Wang, and Dai]{zhu2020deformable}
Xizhou Zhu, Weijie Su, Lewei Lu, Bin Li, Xiaogang Wang, and Jifeng Dai.
\newblock Deformable detr: Deformable transformers for end-to-end object detection.
\newblock \emph{arXiv preprint arXiv:2010.04159}, 2020{\natexlab{a}}.

\bibitem[Zhu et~al.(2020{\natexlab{b}})Zhu, Hu, Qiu, Shi, Kang, Mou, Bagheri, Haberle, Hua, Huang, Hughes, Li, Sun, Zhang, Han, Schmitt, and Wang]{so2Sat}
Xiao~Xiang Zhu, Jingliang Hu, Chunping Qiu, Yilei Shi, Jian Kang, Lichao Mou, Hossein Bagheri, Matthias Haberle, Yuansheng Hua, Rong Huang, Lloyd Hughes, Hao Li, Yao Sun, Guichen Zhang, Shiyao Han, Michael Schmitt, and Yuanyuan Wang.
\newblock So2sat lcz42: A benchmark data set for the classification of global local climate zones [software and data sets].
\newblock \emph{IEEE Geoscience and Remote Sensing Magazine}, 8\penalty0 (3):\penalty0 76--89, 2020{\natexlab{b}}.

\end{thebibliography}
